\documentclass[10pt,twocolumn,letterpaper]{article}

\usepackage{tikz}
\usetikzlibrary{shapes.callouts}

\usepackage{cvpr}

\usepackage{times}
\usepackage{relsize}
\usepackage[T1]{fontenc} 
\usepackage[latin1]{inputenc}
\usepackage[english]{babel}

\usepackage{epsfig}
\usepackage{graphicx}
\usepackage{wrapfig}
\usepackage[belowskip=0pt,aboveskip=3pt,font=small]{caption}
\usepackage[belowskip=0pt,aboveskip=0pt,font=small]{subcaption}
\usepackage{amsmath, amsthm, amssymb}
\usepackage{textcomp}
\usepackage{stmaryrd}
\usepackage{upgreek}
\usepackage{bm}
\usepackage{cases}
\usepackage{mathtools}
\usepackage{arydshln}
\usepackage{multirow}

\usepackage{cite}
\usepackage[pagebackref=true,breaklinks=true,letterpaper=true,colorlinks,bookmarks=false,citecolor=green]{hyperref}
\usepackage{bibspacing}
\usepackage{algorithm}
\usepackage{algpseudocode}

\usepackage{multirow}
\usepackage{rotating}
\usepackage{booktabs}

\usepackage{enumitem}
\usepackage[olditem,oldenum]{paralist}

\usepackage{alltt}
\usepackage{listings}

\belowdisplayskip \abovedisplayskip

\newlength{\sectionReduceTop}
\newlength{\sectionReduceBot}
\newlength{\subsectionReduceTop}
\newlength{\subsectionReduceBot}
\newlength{\abstractReduceTop}
\newlength{\abstractReduceBot}
\newlength{\captionReduceTop}
\newlength{\captionReduceBot}
\newlength{\subsubsectionReduceTop}
\newlength{\subsubsectionReduceBot}

\newlength{\eqnReduceTop}
\newlength{\eqnReduceBot}

\newlength{\horSkip}
\newlength{\verSkip}

\newlength{\figureHeight}
\usepackage{mysymbols}

\usepackage{url}
\usepackage{xspace}
\usepackage{comment}
\usepackage{color}
\usepackage{xcolor}
\usepackage{afterpage}
\usepackage{pdfpages}
\usepackage{framed}
\usepackage{fancybox}
\usepackage{cuted}
\usepackage{caption}

\setdefaultleftmargin{.2cm}{.2cm}{}{}{}{}
\setlist{leftmargin=.2cm}

\newcommand{\dialog}{dialog\xspace}
\newcommand{\Dialog}{Dialog\xspace}
\newcommand{\dialogs}{dialogs\xspace}

\newcommand{\vd}{VisDial\xspace}
\newcommand{\vdfull}{Visual \Dialog}
\newcommand{\vqa}{VQA\xspace}

\newcommand{\myquote}[1]{\emph{`#1'}}

\newcommand{\seqtoseq}{Seq2Seq\xspace}

\newcommand{\lffull}{Late Fusion\xspace}
\newcommand{\lf}{LF\xspace}
\newcommand{\hrefull}{Hierarchical Recurrent Encoder\xspace}
\newcommand{\hre}{HRE\xspace}
\newcommand{\mn}{MN\xspace}
\newcommand{\mnfull}{Memory Network\xspace}

\newcommand{\reffig}[1]{Fig.~\ref{#1}}
\newcommand{\refsec}[1]{Sec.~\ref{#1}}
\newcommand{\reftab}[1]{Tab.~\ref{#1}}

\newcommand{\train}{\texttt{train}\xspace}
\newcommand{\val}{\texttt{val}\xspace}
\newcommand{\test}{\texttt{test}\xspace}

\newcommand{\ad}[1]{\textcolor{teal}{#1}}
\newcommand{\sk}[1]{\textcolor{blue}{#1}}

\newcommand{\FIXME}[1]{\textbf{\textcolor{red}{#1}}}

\cvprfinalcopy 

\def\cvprPaperID{121} 

\usepackage{flushend}

\begin{document}

\title{Visual \Dialog}

\author{
    Abhishek Das$^1$, Satwik Kottur$^2$,
    Khushi Gupta$^2$\thanks{Work done while KG and AS were interns at Virginia Tech.},
    Avi Singh$^3$\footnotemark[1], Deshraj Yadav$^4$,
    Jos\'e M.F. Moura$^2$, \\
    Devi Parikh$^1$, Dhruv Batra$^1$ \\
    $^1$Georgia Institute of Technology, $^2$Carnegie Mellon University, $^3$UC Berkeley, $^4$Virginia Tech \\
    {\tt\small $^1$\{abhshkdz, parikh, dbatra\}@gatech.edu}
    \quad \tt\small $^2$\{skottur, khushig, moura\}@andrew.cmu.edu \\
    \quad \tt\small $^3$avisingh@cs.berkeley.edu
    \quad \tt\small $^4$deshraj@vt.edu \\
    \tt\normalsize \href{https://visualdialog.org}{visualdialog.org}
}

\maketitle

\vspace{\abstractReduceTop}
\begin{abstract}
\vspace{\abstractReduceBot}
\vspace{-2pt}

We introduce the task of \vdfull, which requires an AI agent to hold a meaningful \dialog with
humans in natural, conversational language about visual content.
Specifically, given an image, a \dialog history, and a question about the image,
the agent has to ground the question in image, infer context from history, and answer the question accurately.
\vdfull is disentangled enough from a specific downstream task so as to serve as a general test of machine
intelligence, while being grounded in vision enough to allow objective evaluation of
individual responses and benchmark progress.
We develop a novel two-person chat data-collection protocol to curate a large-scale \vdfull dataset (\vd).
VisDial v0.9 has been released and contains
1 \dialog with 10 question-answer pairs on $\sim$120k images from COCO, with a total
of $\sim$1.2M \dialog question-answer pairs.

We introduce a family of neural encoder-decoder models for \vdfull with 3 encoders --
\lffull, \hrefull and \mnfull
~-- and 2 decoders (generative and discriminative), which outperform a number of sophisticated baselines.
We propose a retrieval-based evaluation protocol for \vdfull where the AI agent is
asked to sort a set of candidate answers and evaluated on metrics such as mean-reciprocal-rank of human response.
We quantify gap between machine and human performance on the \vdfull task via human studies.
Putting it all together, we demonstrate the first `visual chatbot'!
Our dataset, code, trained models and visual chatbot are available on {\small\url{https://visualdialog.org}}.

\end{abstract}

\vspace{-10pt}
\section{Introduction}
\label{sec:intro}

We are witnessing unprecedented advances in
computer vision (CV) and artificial intelligence (AI) --
from `low-level' AI tasks such as image classification \cite{he_cvpr16},
scene recognition \cite{wang_arxiv16},
object detection \cite{liu_eccv16}
-- to `high-level' AI tasks such as
learning to play Atari video games~\cite{mnih_nature15} and Go \cite{silver_nature16},
answering reading comprehension questions by understanding short stories \cite{hermann_nips15,weston_iclr16},
and even answering questions about images \cite{antol_iccv15,ren_nips15,malinowski_iccv15,zitnick_ai16} and videos~\cite{tapaswi_cvpr16,tu_mm16}!

\begin{figure}[t]
  \centering
  \includegraphics[width=\linewidth]{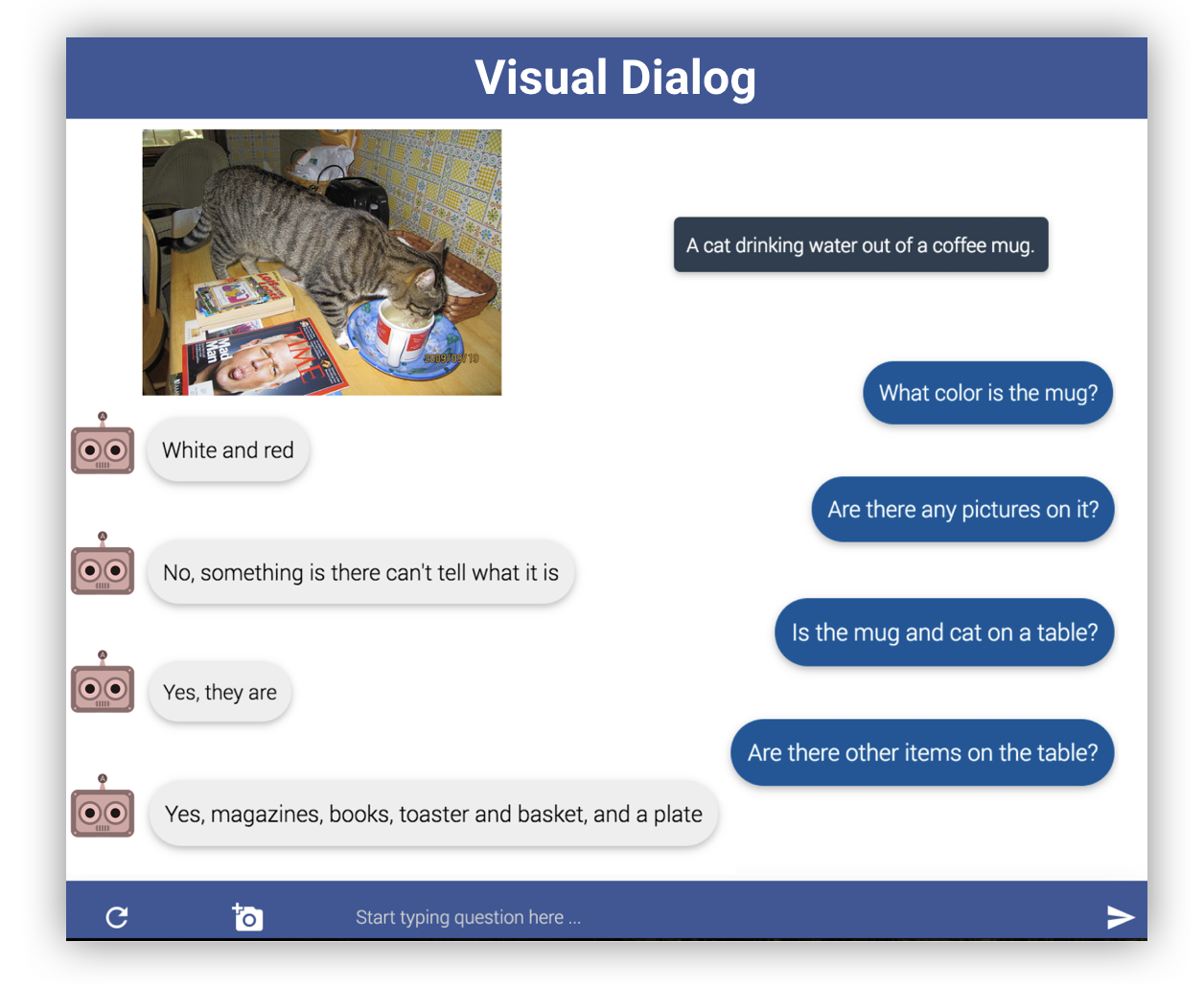}
  \caption{
  We introduce a new AI task -- Visual \Dialog, where an AI agent must
  hold a \dialog with a human about visual content.
  We introduce a large-scale dataset (\vd), an evaluation protocol, and novel encoder-decoder models for this task.
  }
  \label{fig:visualchatbot}
\vspace{-15pt}
\end{figure}

\textbf{What lies next for AI?}
We believe that the next generation of visual intelligence systems will need to posses
the ability to hold a meaningful \dialog with humans in natural language about visual content.
Applications include:
%
\begin{compactitem}

\item Aiding visually impaired users in understanding their surroundings~\cite{bigham_uist10}
or social media content~\cite{fb_blog16}
(AI: \myquote{John just uploaded a picture from his vacation in Hawaii},
Human: \myquote{Great, is he at the beach?},
AI: \myquote{No, on a mountain}).

\item Aiding analysts in making decisions based on large quantities of surveillance data
(Human: \myquote{Did anyone enter this room last week?},
AI: \myquote{Yes, 27 instances logged on camera},
Human: \myquote{Were any of them carrying a black bag?}),

\item Interacting with an AI assistant
(Human: \myquote{Alexa -- can you see the baby in the baby monitor?},
AI: \myquote{Yes, I can},
Human: \myquote{Is he sleeping or playing?}).

\item Robotics applications (\eg search and rescue missions)
where the operator may be `situationally blind' and operating via
language~\cite{mei_aaai16} (Human: \myquote{Is there smoke in any room around you?},
AI: \myquote{Yes, in one room},
Human: \myquote{Go there and look for people}).

\end{compactitem}

\begin{figure}
  \centering
  \includegraphics[width=0.95\columnwidth, clip, trim={ 2.5cm 2cm 2cm 2cm}]{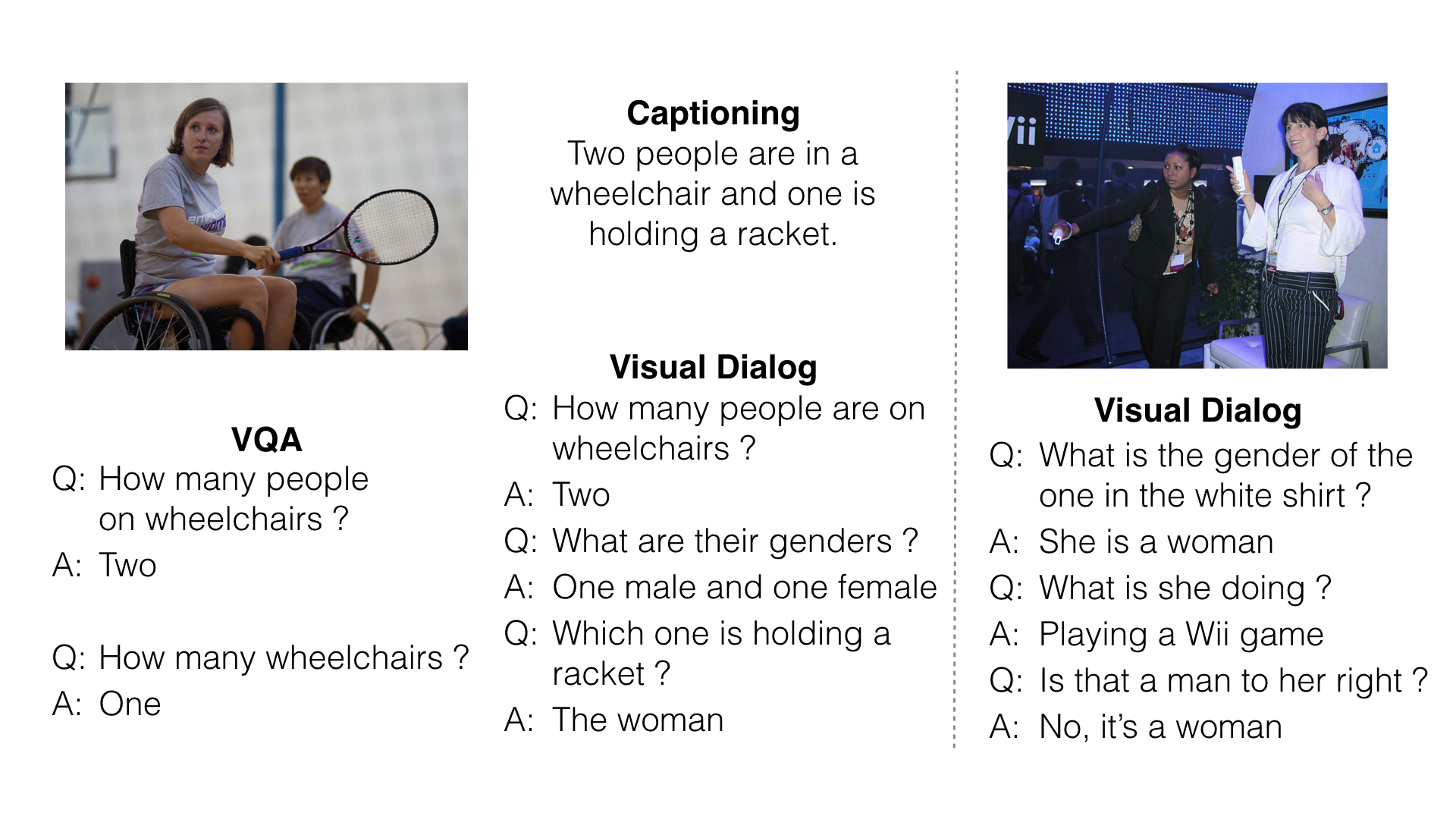}
  \caption{Differences between image captioning, Visual Question Answering (VQA) and \vdfull.
  Two (partial) \dialogs are shown from our \vd dataset, which is curated from a live chat between
  two Amazon Mechanical Turk workers (\refsec{sec:dataset}).}
  \label{fig:teaser}
\vspace{-15pt}
\end{figure}


Despite rapid progress at the intersection of vision and language -- in particular, in
image captioning and visual question answering (VQA) --
it is clear that we are far from this grand goal of an AI agent that can `see' and `communicate'.
%
In captioning, the human-machine interaction consists of the machine simply \emph{talking at} the human
(\myquote{Two people are in a wheelchair and one is holding a racket}),
with no \dialog or input from the human.
While \vqa takes a significant step towards human-machine interaction,
it still represents only \emph{a single round of a \dialog}  
-- unlike in human conversations, there is no scope for follow-up questions,
no memory in the system of previous questions asked by the user nor consistency with respect to previous answers provided by the system
(Q: \myquote{How many people on wheelchairs?},
A: \myquote{Two}; Q: \myquote{How many wheelchairs?}, A: \myquote{One}).

As a step towards conversational visual AI, we introduce a novel task
--  \textbf{\vdfull} -- along with a large-scale dataset, an evaluation protocol, and novel deep models.

\textbf{Task Definition.}
The concrete task in \vdfull is the following --
given an image $I$, a history of a \dialog consisting of a sequence of question-answer pairs
(Q1: \myquote{How many people are in wheelchairs?},
A1: \myquote{Two},
Q2: \myquote{What are their genders?},
A2: \myquote{One male and one female}),
and a natural language follow-up question (Q3: \myquote{Which one is holding a racket?}),
the task for the machine is to answer the question in free-form natural language (A3: \myquote{The woman}).
This task is the visual analogue of the Turing Test.


Consider the \vdfull examples in \reffig{fig:teaser}.
The question \myquote{What is the gender of the one in the white shirt?} requires the machine
to selectively focus and direct attention to a relevant region.
\myquote{What is she doing?} requires co-reference resolution (whom does the pronoun `she' refer to?),
\myquote{Is that a man to her right?} further requires the machine to have visual memory (which object in the image were we talking about?).
Such systems also need to be consistent with their outputs --
\myquote{How many people are in wheelchairs?}, \myquote{Two}, \myquote{What are their genders?},
\myquote{One male and one female} --
note that the number of genders being specified should add up to two.
Such difficulties make the problem a highly interesting and challenging one.


\textbf{Why do we talk to machines?}
Prior work in language-only (non-visual) \dialog can be arranged on a spectrum with the following two end-points:  \\
goal-driven \dialog (\eg booking a flight for a user) $\longleftrightarrow$
goal-free \dialog (or casual `chit-chat' with chatbots). \\
The two ends have vastly differing purposes and conflicting evaluation criteria.
Goal-driven \dialog is typically evaluated on task-completion rate (how frequently was the user able to
book their flight) or time to task completion \cite{paek_elds01,dodge_iclr16} --
clearly, the shorter the \dialog the better. In contrast, for chit-chat, the longer the user engagement and interaction,
the better. For instance, the goal of the 2017 \$2.5 Million Amazon Alexa Prize is to ``create a socialbot that
converses coherently and engagingly with humans on popular topics for 20 minutes.''

We believe our instantiation of \vdfull hits a sweet spot on this spectrum.
It is \emph{disentangled enough} from a specific downstream task so as to serve as a general test of machine
intelligence, while being \emph{grounded enough} in vision to allow objective evaluation of
individual responses and benchmark progress.
The former discourages task-engineered bots for `slot filling' \cite{lemon_eacl06} and the latter
discourages bots that put on a personality to avoid answering questions while keeping the user engaged \cite{eliza}.

\textbf{Contributions.}
We make the following contributions:
\begin{compactitem}

\item We propose a new AI task: \vdfull, where a machine must
hold \dialog with a human about visual content.

\item
We develop a novel two-person chat data-collection protocol to curate a large-scale \vdfull dataset (\vd).
  Upon completion\footnote{VisDial data on COCO-\train ($\sim$83k images) and COCO-\val($\sim$40k images) is already available
  for download at {\scriptsize \url{https://visualdialog.org}}. Since dialog history contains the ground-truth caption,
  we will not be collecting dialog data on COCO-\test. Instead, we will collect dialog data on 20k extra images from COCO distribution
  (which will be provided to us by the COCO team) for our \test set.},
  \vd will contain 1 \dialog each (with 10 question-answer pairs) on $\sim$140k images from the COCO dataset~\cite{mscoco},
  for a total of $\sim$1.4M \dialog question-answer pairs.
When compared to VQA~\cite{antol_iccv15}, \vd studies a significantly richer task (dialog),
overcomes a `visual priming bias' in VQA (in \vd, the questioner does not see the image),
contains free-form longer answers, and is \emph{an order of magnitude} larger.

\item
We introduce a family of neural encoder-decoder models for \vdfull with 3 novel encoders
\begin{compactitem}
\item \lffull: that embeds the image, history, and question into vector spaces separately and performs a `late fusion' of these
into a joint embedding.

\item \hrefull: that
contains a \dialog-level Recurrent Neural Network (RNN) sitting on top of a question-answer ($QA$)-level
recurrent block. In each $QA$-level recurrent block, we also include an attention-over-history mechanism to choose and attend to
the round of the history relevant to the current question.

\item \mnfull: that
treats each previous $QA$ pair as a `fact' in its memory bank and learns to `poll' the stored facts and the image to develop a
context vector.

\end{compactitem}

We train all these encoders with 2 decoders (generative and discriminative) -- all settings outperform a number of sophisticated baselines,
including our adaption of state-of-the-art VQA models to \vd.

\item
We propose a retrieval-based evaluation protocol for \vdfull where the AI agent is
asked to sort a list of candidate answers and evaluated on metrics such as mean-reciprocal-rank of the human response.

\item We conduct studies to quantify human performance.

\item
Putting it all together, on the project page we demonstrate the first visual chatbot!

\end{compactitem}

\section{Related Work}
\label{sec:related}

\textbf{Vision and Language.}
A number of problems at the intersection of vision and language have recently gained prominence --
image captioning~\cite{fang_cvpr15,vinyals_cvpr15,karpathy_cvpr15,donahue_cvpr15},
video/movie description~\cite{venugopalan_iccv15,venugopalan_naacl15, rohrbach_cvpr15},
text-to-image coreference/grounding~\cite{kong_cvpr14,ramanathan_eccv14,plummer_iccv15,hu_eccv16,rohrbach_eccv16,christie_emnlp16},
visual storytelling~\cite{huang_naacl16, agrawal_emnlp16},
and of course, visual question answering (VQA)
\cite{malinowski_nips14,antol_iccv15,ren_nips15,gao_nips15,malinowski_iccv15,lu_nips16,zhang_cvpr16,
aagrawal_emnlp16,das_emnlp16,goyal_cvpr17}.
However, all of these involve (at most) a single-shot natural language interaction -- there is no \dialog.
    Concurrent with our work, two recent works~\cite{vries_cvpr17, mostafazadeh_arxiv17} have also
    begun studying visually-grounded dialog.



\textbf{Visual Turing Test.}
Closely related to our work is that of Geman~\etal~\cite{geman_pnas14},
who proposed a fairly restrictive `Visual Turing Test' -- a system that asks templated, binary questions.
    In comparison, 1) our dataset has \emph{free-form, open-ended} natural language questions collected via two subjects chatting on
    Amazon Mechanical Turk (AMT), resulting in a more realistic and diverse dataset (see \reffig{fig:quesCircles_main}).
    2) The dataset in \cite{geman_pnas14} only contains street scenes, while our dataset has
    considerably more variety since it uses images from COCO~\cite{mscoco}.
    Moreover, our dataset is \emph{two orders of magnitude larger}  --
    2,591 images in \cite{geman_pnas14} vs $\sim$140k images, 10 question-answer pairs per image, total of
    $\sim$1.4M QA pairs.

\textbf{Text-based Question Answering.}
Our work is related to text-based question answering or `reading comprehension' tasks studied in the NLP community.
Some recent large-scale datasets in this domain include the
30M Factoid Question-Answer corpus~\cite{serban_acl16},
100K SimpleQuestions dataset \cite{bordes_arxiv15},
DeepMind Q\&A dataset~\cite{hermann_nips15},
the 20 artificial tasks in the bAbI dataset~\cite{weston_iclr16}, and
the SQuAD dataset for reading comprehension~\cite{rajpurkar_emnlp16}.
\vd can be viewed as a \emph{fusion} of reading comprehension and VQA. In \vd,
the machine must comprehend the history of the past \dialog and then understand the image to answer the question.
By design, the answer to any question in \vd is not present in the past \dialog~-- if it were, the question would not be
asked. The history of the \dialog \emph{contextualizes} the question~-- the question \myquote{what else is she holding?}
requires a machine to comprehend the history to realize who the question is talking about and what has been
excluded, and then understand the image to answer the question.



\textbf{Conversational Modeling and Chatbots.}
\vdfull is the visual analogue of text-based \dialog and conversation modeling.
While some of the earliest developed chatbots were rule-based~\cite{eliza}, end-to-end learning based
approaches are now being actively explored~\cite{serban_aaai16,kannan_kdd16,vinyals_arxiv15,dodge_iclr16,bordes_arxiv16,serban_arxiv16,li_emnlp16}.
A recent large-scale conversation dataset is the Ubuntu Dialogue Corpus~\cite{lowe_sigdial15},
which contains about 500K \dialogs extracted from the Ubuntu channel on Internet Relay Chat (IRC).
Liu \etal~\cite{liu_emnlp16} perform a study of problems in existing evaluation protocols for free-form \dialog.
One important difference between free-form textual \dialog and \vd is that in \vd, the two participants are not symmetric --
one person (the `questioner') asks questions about an image \emph{that they do not see};
the other person (the `answerer') sees the image and only answers the questions
(in otherwise unconstrained text, but no counter-questions allowed).
This role assignment gives a sense of purpose to the interaction
(why are we talking? To help the questioner build a mental model of the image),
and allows objective evaluation of individual responses.

\begin{figure*}[ht]
\centering
\begin{subfigure}[b]{0.32\textwidth}
	\centering
	\includegraphics[trim={2cm 2cm 2cm 2cm},clip, height=0.7\textwidth]{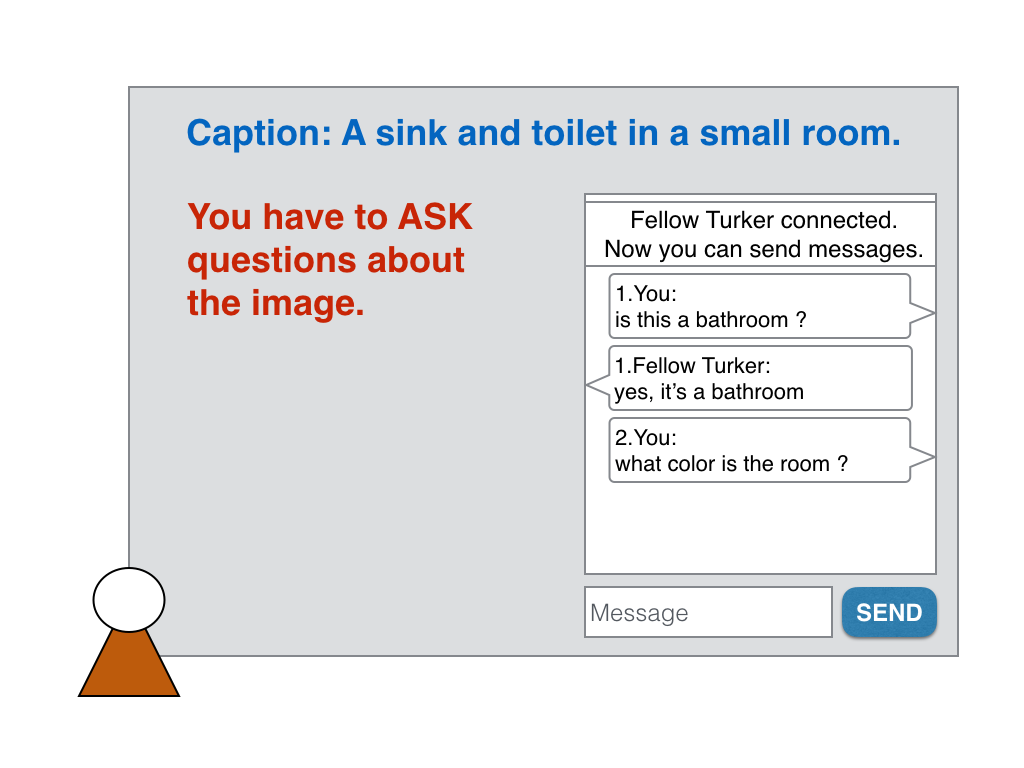}
	\caption{What the `questioner' sees.}
	\label{fig:q_interface}
\end{subfigure}
\begin{subfigure}[b]{0.32\textwidth}
	\centering
	\includegraphics[trim={2cm 2cm 2cm 2cm},clip, height=0.7\textwidth]{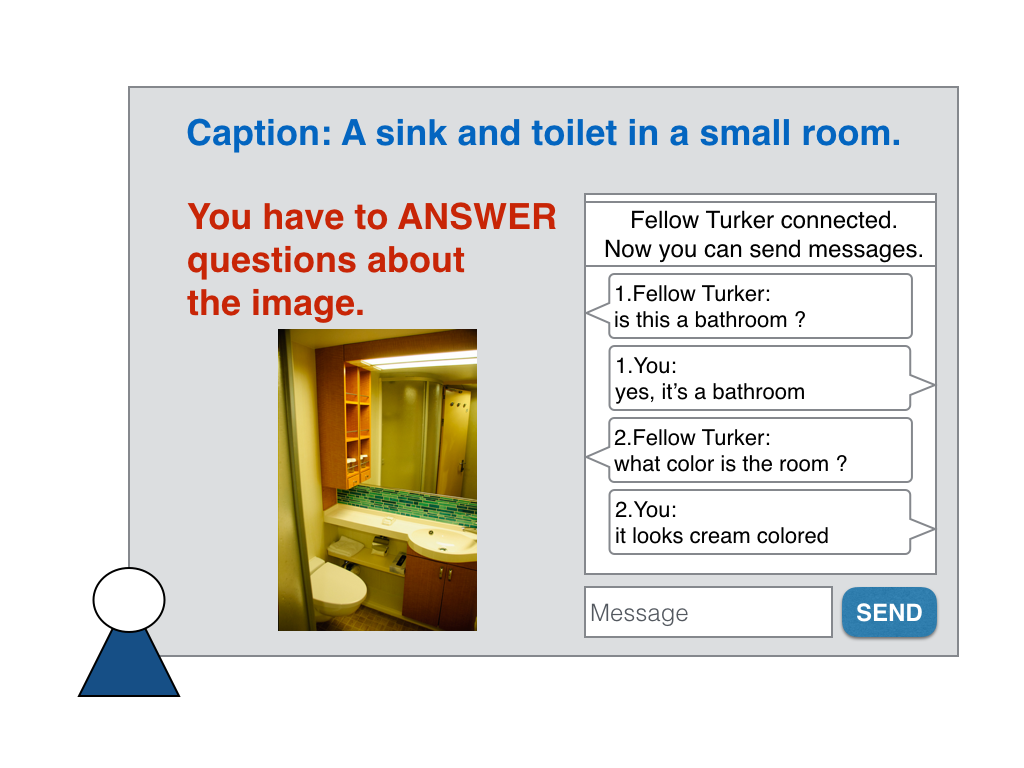}
	\caption{What the `answerer' sees.}
	\label{fig:a_interface}
\end{subfigure}
\begin{subfigure}[b]{0.32\textwidth}
	\centering
	\includegraphics[trim={2cm 1.5cm 2cm 2cm},clip, height=0.7\textwidth]{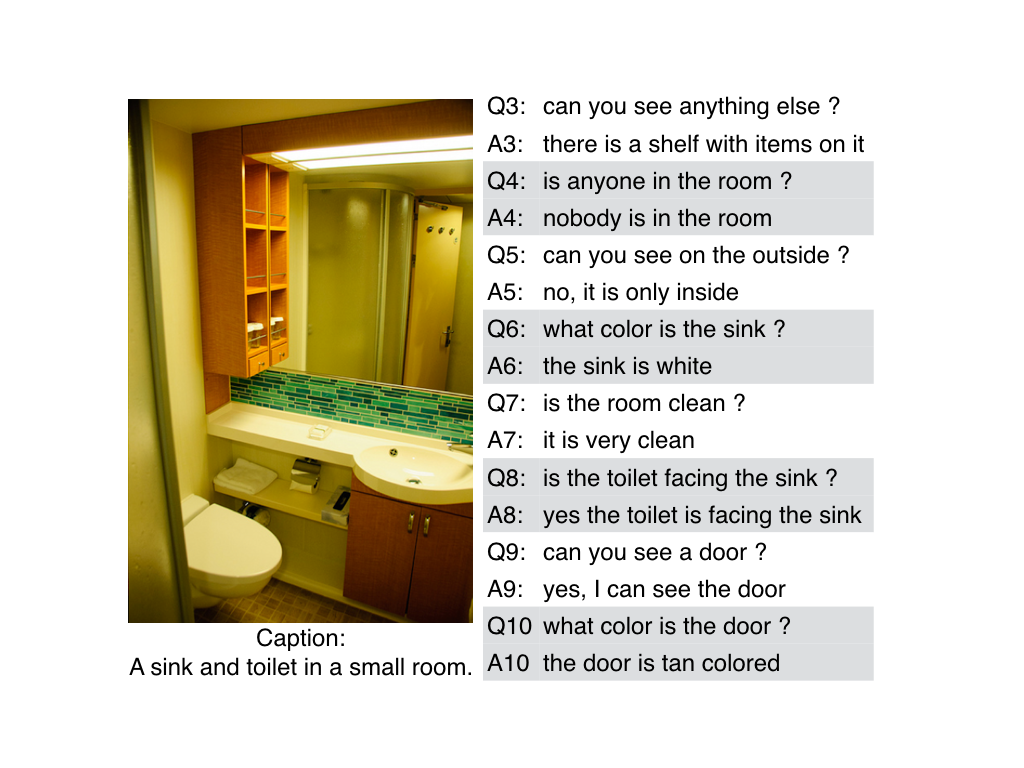}
	\caption{Example \dialog from our \vd dataset.}
	\label{fig:vd_eg}
\end{subfigure}
\caption{Collecting visually-grounded dialog data on Amazon Mechanical Turk via a live chat interface
where one person is assigned the role of `questioner' and the second person is the `answerer'.
We show the first two questions being collected via the interface as Turkers interact with each other in \figref{fig:q_interface} and \figref{fig:a_interface}. Remaining questions are shown in \figref{fig:vd_eg}.}
\vspace{-10pt}
\label{fig:vd_amt}
\end{figure*}

\vspace{\sectionReduceBot}
\section{The \vdfull Dataset (\vd)}
\label{sec:dataset}


We now describe our \vd dataset.
We begin by describing the chat interface and data-collection process on AMT,
analyze the dataset, then discuss the evaluation protocol.

Consistent with previous data collection efforts, we collect visual \dialog data on
images from the Common Objects in Context (COCO)~\cite{mscoco} dataset, which contains
multiple objects in everyday scenes.
The visual complexity of these images allows for engaging and diverse conversations.

\textbf{Live Chat Interface.}
Good data for this task should include \dialogs that have
(1) temporal continuity, (2) grounding in the image, and (3) mimic natural `conversational' exchanges.
To elicit such responses, we paired 2 workers on AMT to chat with each other in real-time (\figref{fig:vd_amt}).
Each worker was assigned a specific role. One worker (the `questioner')
sees only a single line of text describing an image (caption from COCO); the image remains hidden to the questioner.
Their task is to ask questions about this hidden image to `imagine the scene better'.
The second worker (the `answerer') sees the image and caption. Their task is to answer
questions asked by their chat partner. Unlike VQA \cite{antol_iccv15}, answers are not restricted to
be short or concise, instead workers are encouraged to reply as naturally and `conversationally' as possible.
\figref{fig:vd_eg} shows an example \dialog.

This process is an unconstrained `live' chat, with the only exception
that the questioner must wait to receive an answer before posting the next question.
The workers are allowed to end the conversation after 20 messages are exchanged
(10 pairs of questions and answers).
Further details about our final interface can be found in the supplement.

We also piloted a different setup where the questioner saw a
highly blurred version of the image, instead of the caption.
The conversations seeded with blurred images resulted in questions that were essentially `blob recognition'
-- \myquote{What is the pink patch at the bottom right?}.
For our full-scale data-collection, we decided to seed with just the captions since it
resulted in more `natural' questions and more closely modeled the real-world
applications discussed in \secref{sec:intro} where no visual signal is available to the human.\\

\vspace{2pt}
\textbf{Building a 2-person chat on AMT.}
Despite the popularity of AMT as a data collection platform in computer vision, our setup had to
design for and overcome some unique challenges -- the key issue being that AMT is simply not designed
for multi-user Human Intelligence Tasks (HITs).
Hosting a live two-person chat on AMT meant that
none of the Amazon tools could be used and we developed our own backend messaging and data-storage
infrastructure based on Redis messaging queues and Node.js.
To support data quality, we ensured that a worker could not chat with themselves (using
say, two different browser tabs) by maintaining a pool of worker IDs paired.
To minimize wait time for one worker while the second was being searched for,
we ensured that there
was always a significant pool of available HITs.
If one of the workers abandoned a HIT (or was disconnected) midway,
automatic conditions in the code kicked in asking the remaining worker to either continue
asking questions or providing facts (captions) about the image (depending on their role) till 10 messages
were sent by them. Workers who completed the task in this way were fully compensated, but our backend discarded this data and automatically launched a new HIT on this image so a real two-person conversation could be recorded.
Our entire data-collection infrastructure (front-end UI, chat interface, backend storage and messaging system,
error handling protocols) is publicly available\footnote{\url{https://github.com/batra-mlp-lab/visdial-amt-chat}}.




\section{\vd Dataset Analysis}
\label{sec:analysis}

	We now analyze the v0.9 subset of our \vd dataset~-- it contains 1 \dialog
	(10 QA pairs) on $\sim$123k images from COCO-\train/\val,
	a total of 1,232,870 QA pairs.


\subsection{Analyzing \vd Questions}
\label{sec:analysis_questions}



\textbf{Visual Priming Bias.}
One key difference between \vd and previous image question-answering datasets (VQA~\cite{antol_iccv15},
Visual 7W~\cite{zhu_cvpr16}, Baidu mQA~\cite{gao_nips15}) is the lack of a `visual priming bias' in \vd.
Specifically, in all previous datasets, subjects saw an image while asking questions about it.
As analyzed in \cite{zhang_cvpr16, aagrawal_emnlp16, goyal_cvpr17}, this leads to a particular bias in the questions --
people only ask \myquote{Is there a clocktower in the picture?} on pictures actually containing clock towers.
This allows language-only models to perform remarkably well on VQA and results in an inflated sense
of progress~\cite{zhang_cvpr16, goyal_cvpr17}.
As one particularly perverse example -- for questions in the VQA dataset starting with 
\myquote{Do you see a \ldots}, blindly answering \myquote{yes} without reading the rest of the question
or looking at the associated image results in an average VQA accuracy of $87\%$!
In \vd, questioners \emph{do not} see the image. As a result, this bias is reduced.

\begin{figure}[t]
	\centering
	\begin{subfigure}[b]{0.48\columnwidth}
		\centering
		\includegraphics[width=\textwidth]{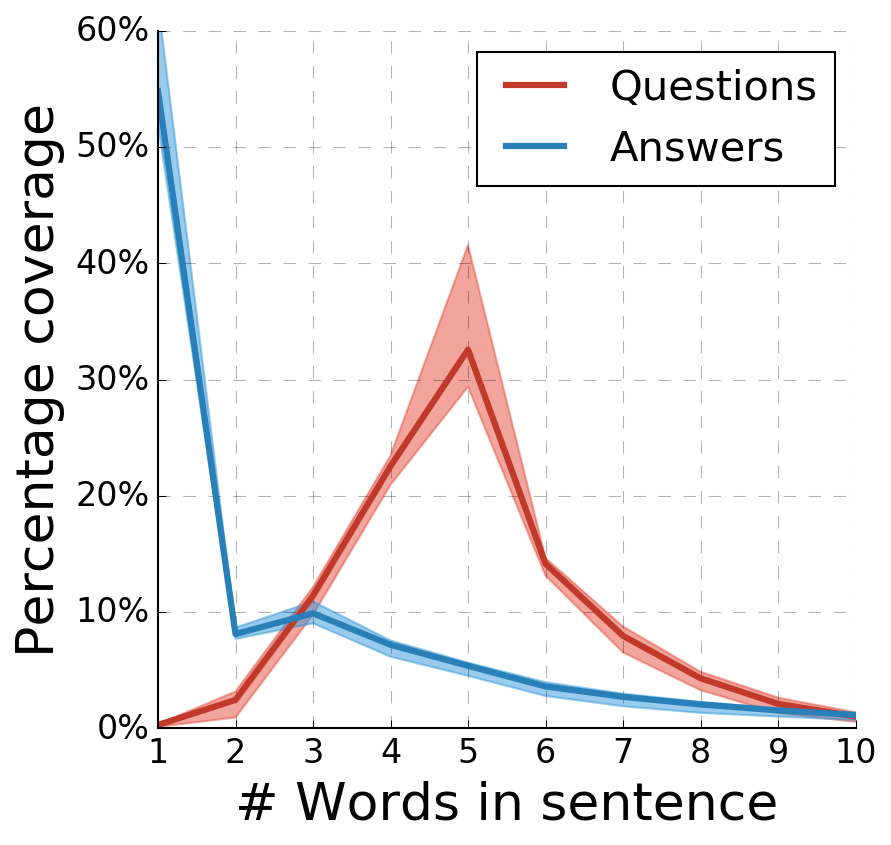}
		\caption{}
		\label{fig:quesAnsLengths_main}
	\end{subfigure}
	\begin{subfigure}[b]{0.48\columnwidth}
		\centering
		\includegraphics[width=\textwidth]{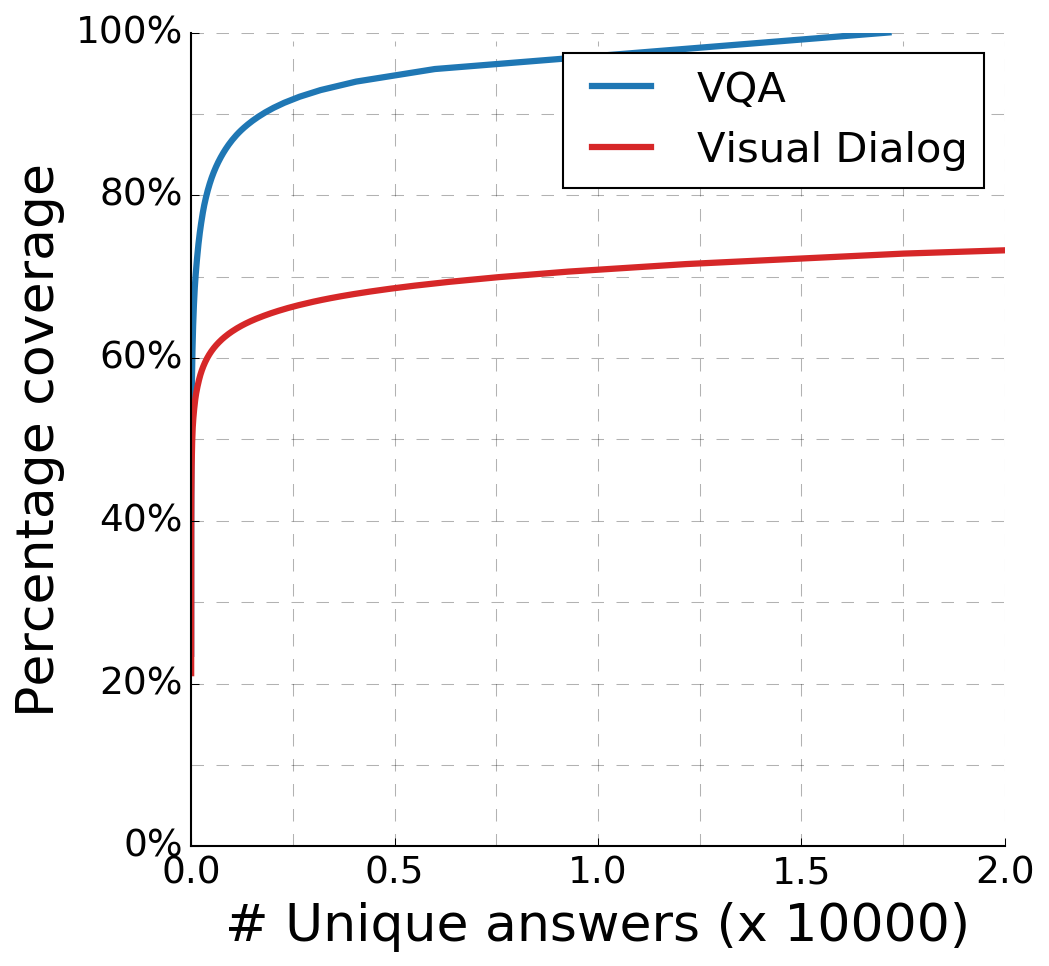}
		\caption{}
		\label{fig:coverage_main}
	\end{subfigure}
	\quad
	\caption{Distribution of lengths for questions and answers (left); and percent coverage of unique answers over all answers from the train dataset (right), compared to VQA. For a given coverage, \vd has more unique answers indicating greater answer diversity.}
	\vspace{-10pt}
\end{figure}

\begin{figure*}[t]
\centering
	\vspace{-15pt}
	\begin{subfigure}[t]{0.32\textwidth}
		\includegraphics[trim={0 0 0 0},clip,height=\textwidth]{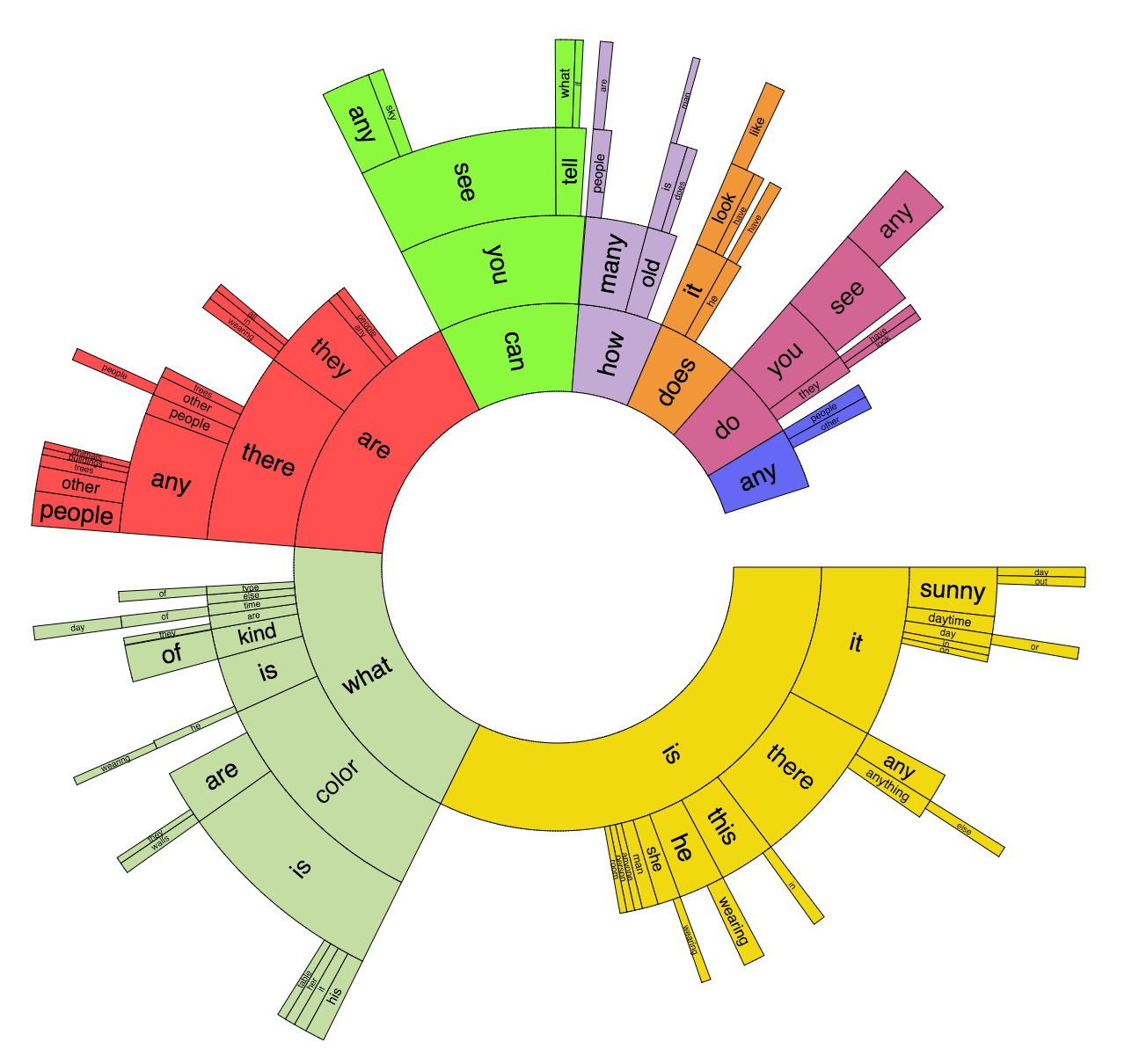}
		\caption{\vd Questions}
		\label{fig:vd_qlen}
	\end{subfigure}
	\begin{subfigure}[t]{0.32\textwidth}
		\includegraphics[trim={0 0 0 0},clip,height=\textwidth]{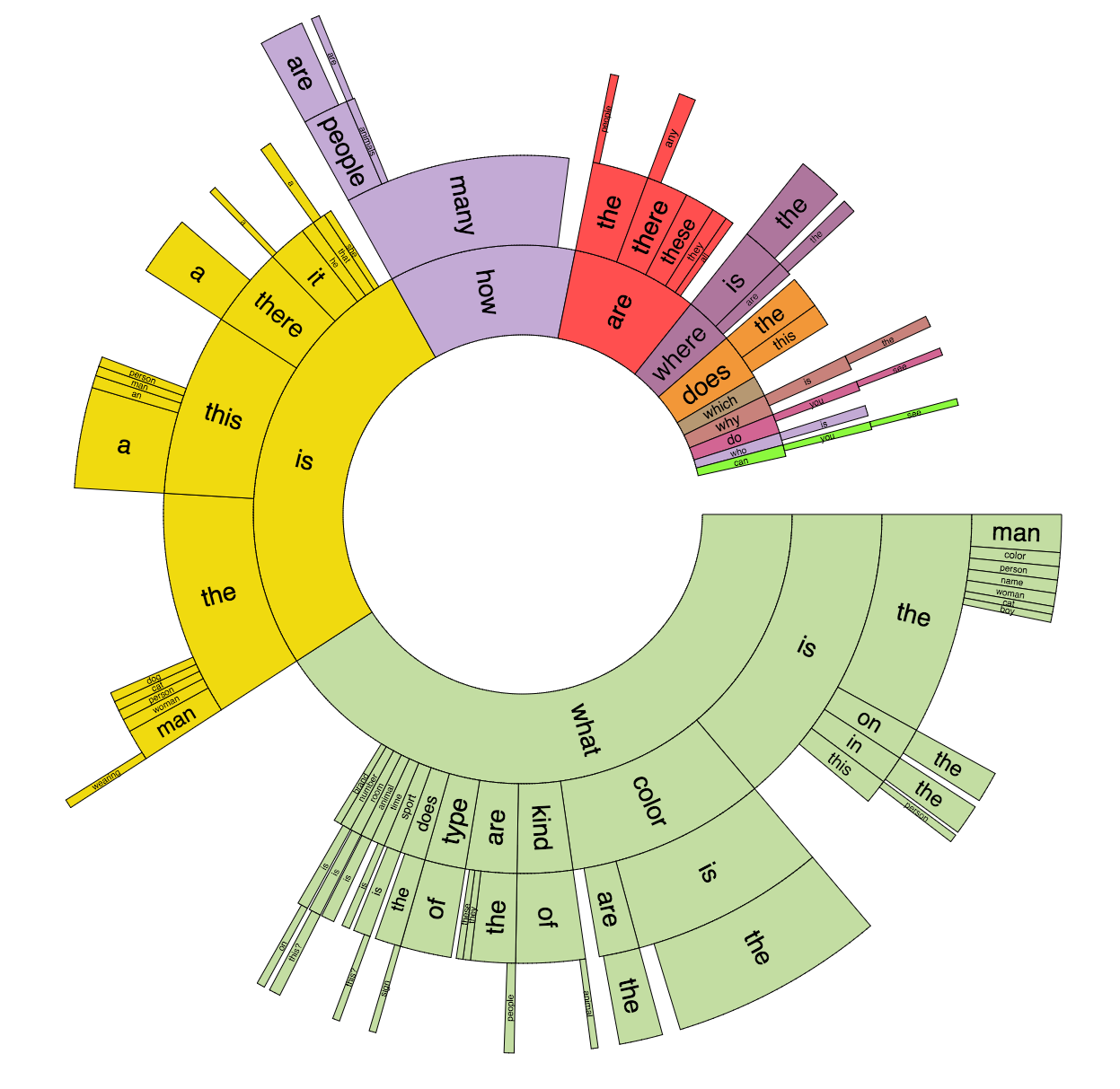}
		\caption{\vqa Questions}
		\label{fig:vqa_qlen}
	\end{subfigure}
	\begin{subfigure}[t]{0.32\textwidth}
		\centering
		\includegraphics[trim={0 0 0 0},clip,height=\textwidth]{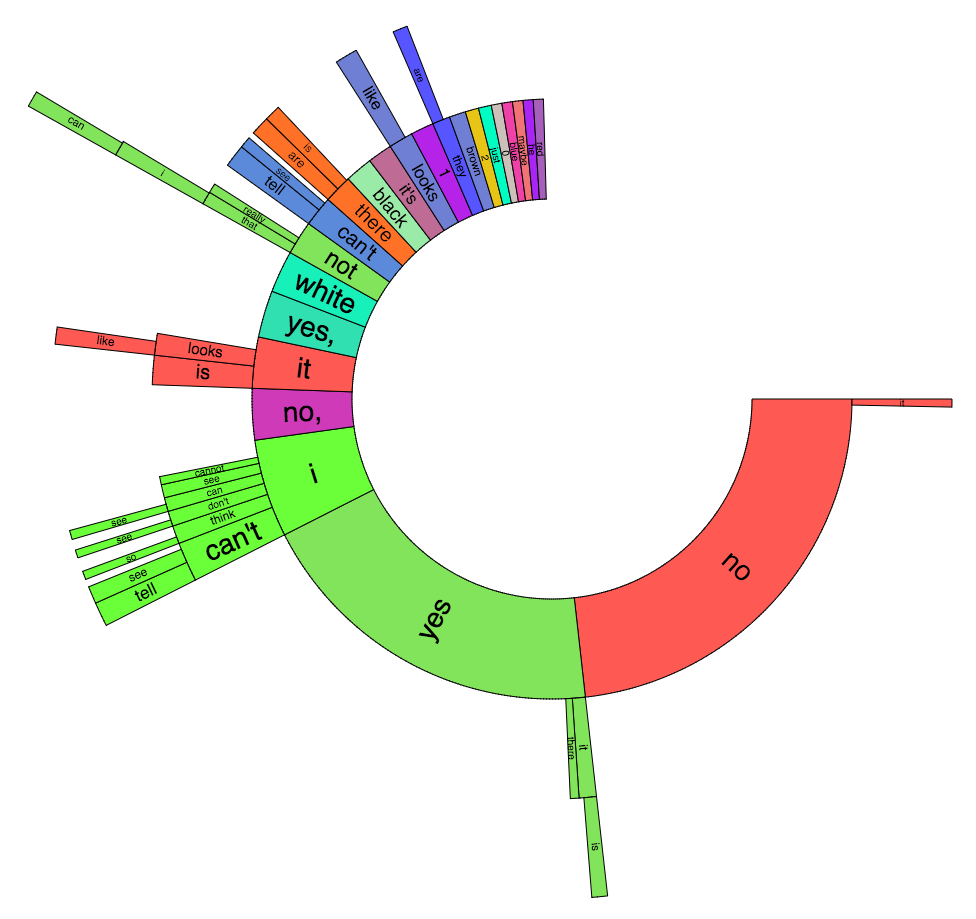}
		\caption{\vd Answers}
		\label{fig:ansCircle_main}
	\end{subfigure}
	\caption{Distribution of first n-grams for (left to right) \vd questions, \vqa questions and \vd answers.
	Word ordering starts towards the center and radiates outwards, and arc length is proportional to number of questions containing the word.}
	\label{fig:quesCircles_main}
    \vspace{-10pt}
\end{figure*}

\vspace{10pt}
\textbf{Distributions.}
\figref{fig:quesAnsLengths_main} shows the distribution of question lengths in \vd~--
we see that most questions range from four to ten words.
\figref{fig:quesCircles_main} shows `sunbursts' visualizing the distribution of questions (based on the first four words)
in \vd \vs \vqa. While there are a lot of similarities, some differences immediately jump out.
There are more binary questions\footnote{
Questions starting in `Do', `Did', `Have', `Has', `Is', `Are', `Was', `Were', `Can', `Could'.}
in \vd as compared to \vqa~-- the most frequent first question-word in \vd is `is' \vs `what' in \vqa.
A detailed comparison of the statistics of \vd \vs other datasets
is available in Table 1 in the supplement.

Finally, there is a stylistic difference in the questions that is difficult to capture with the simple statistics above.
In \vqa, subjects saw the image and were asked to stump a smart robot. Thus, most queries
involve specific details, often about the background (\myquote{What program is being utilized in the background on the computer?}).
In \vd, questioners did not see the original image and were asking questions to build a mental model of the scene.
Thus, the questions tend to be open-ended, and often follow a pattern:
\begin{compactitem}

\item Generally starting with the entities in the caption:

\begin{flushright}
\myquote{An elephant walking away from a pool in an exhibit}, \\
\myquote{Is there only 1 elephant?},
\end{flushright}

\item digging deeper into their parts or attributes:

\begin{flushright}
\myquote{Is it full grown?}, 
\myquote{Is it facing the camera?},
\end{flushright}

\item asking about the scene category or the picture setting:

\begin{flushright}
\myquote{Is this indoors or outdoors?}, \myquote{Is this a zoo?},
\end{flushright}

\item the weather:

\begin{flushright}
\myquote{Is it snowing?}, \myquote{Is it sunny?},
\end{flushright}


\item simply exploring the scene:

\begin{flushright}
\myquote{Are there people?},
\myquote{Is there shelter for elephant?},
\end{flushright}

\item and asking follow-up questions about the new visual entities discovered
from these explorations:

\begin{flushright}
\myquote{There's a blue fence in background, like an enclosure}, \\
\myquote{Is the enclosure inside or outside?}.
\end{flushright}

\end{compactitem}

\vspace{\subsectionReduceTop}
\subsection{Analyzing \vd Answers}
\vspace{\subsectionReduceBot}

%

\textbf{Answer Lengths.}
\reffig{fig:quesAnsLengths_main} shows the distribution of answer lengths.
Unlike previous datasets, answers in \vd are longer and more descriptive --
mean-length 2.9 words (\vd) vs 1.1 (\vqa), 2.0 (Visual 7W), 2.8 (Visual Madlibs).



\figref{fig:coverage_main} shows the cumulative coverage of all answers (y-axis) by the most frequent answers (x-axis).
The difference between \vd and \vqa is stark -- the top-1000 answers in VQA cover $\sim$83\% of all answers,
while in \vd that figure is only $\sim$63\%. There is a significant heavy tail in \vd~-- most long strings are unique,
and thus the coverage curve in \figref{fig:coverage_main} becomes a straight line with slope 1.
In total, there are 337,527 unique answers in \vd v0.9.

\textbf{Answer Types.}
Since the answers in \vd are longer strings, we can visualize their distribution based
on the starting few words (\reffig{fig:ansCircle_main}). An interesting category of answers emerges
-- \myquote{I think so}, \myquote{I can't tell}, or \myquote{I can't see}  -- expressing doubt, uncertainty, or lack of information.
This is a consequence of the questioner not being able to see the image -- they are asking contextually relevant questions,
but not all questions may be answerable with certainty from that image. We believe this
is rich data for building more human-like AI that refuses to answer questions it doesn't have enough information to answer. 
See~\cite{ray_emnlp16} for a related, but complementary effort on question relevance in VQA.



\textbf{Binary Questions vs Binary Answers.}
In VQA, binary questions are simply those with `yes', `no', `maybe' as answers~\cite{antol_iccv15}.
In \vd, we must distinguish between binary questions and binary answers.
Binary questions are those starting in `Do', `Did', `Have', `Has', `Is', `Are', `Was', `Were', `Can', `Could'.
Answers to such questions can
(1) contain only `yes' or `no',
(2) begin with `yes', `no', and contain additional information or clarification,
(3) involve ambiguity (
\myquote{It's hard to see}, \myquote{Maybe}),
or
(4) answer the question without explicitly saying `yes' or `no'
(Q: \myquote{Is there any type of design or pattern on the cloth?},
A: \myquote{There are circles and lines on the cloth}).
We call answers that contain `yes' or `no' as binary answers~-- 149,367 and 76,346 answers in subsets (1) and (2) from above respectively.
%
Binary answers in \vqa are biased towards `yes'~\cite{antol_iccv15, zhang_cvpr16} --
61.40\% of yes/no answers are `yes'.
In \vd, the trend is reversed. Only 46.96\% are `yes' for all yes/no responses.
This is understandable since workers did not see the image, and were more likely to end up with negative responses.


\vspace{\subsectionReduceTop}
\subsection{Analyzing \vd \Dialog}
\vspace{\subsectionReduceBot}

In \secref{sec:analysis_questions}, we discussed a typical flow of \dialog in \vd. We analyze two quantitative
statistics here.

\textbf{Coreference in \dialog.}
Since language in \vd is the result of a sequential conversation, it naturally contains pronouns --
`he', `she', `his', `her', `it', `their', `they', `this', `that', `those',  \etc.
In total, 38\% of questions, 19\% of answers, and \emph{nearly all} (98\%) \dialogs contain at least one pronoun,
thus confirming that a machine will need to overcome coreference ambiguities
to be successful on this task.
We find that pronoun usage is low in the first round (as expected) and then picks up in frequency.
A fine-grained per-round analysis is available in the supplement.
%


\textbf{Temporal Continuity in \Dialog Topics.}
It is natural for conversational \dialog data to have continuity in the `topics' being discussed.
We have already discussed qualitative differences in \vd questions \vs VQA.
In order to quantify the differences, we performed a human study where we manually annotated
question `topics' for $40$ images (a total of $400$ questions), chosen randomly from the \val set.
The topic annotations were based on human judgement with a consensus of 4 annotators, with topics
such as: asking about a particular object (\myquote{What is the man doing?}) ,
scene (\myquote{Is it outdoors or indoors?}), weather (\myquote{`Is the weather sunny?}),
the image (\myquote{Is it a color image?}), and exploration (\myquote{Is there anything else?'}).
We performed similar topic annotation for questions from VQA for the same set of $40$ images,
and compared topic continuity in questions. 
Across $10$ rounds, \vd question have $4.55 \pm 0.17$ topics on average, confirming that these are not independent questions.
Recall that \vd has $10$ questions per image as opposed to $3$ for VQA.
Therefore, for a fair comparison, we compute average number of topics in \vd over all subsets of $3$ successive questions.
For $500$ bootstrap samples of batch size $40$, \vd has $2.14 \pm 0.05$ topics while VQA has $2.53 \pm 0.09$.
Lower mean suggests there is more continuity in \vd because questions do not change topics as often.
\subsection{\vd Evaluation Protocol}

One fundamental challenge in \dialog systems is evaluation.
    Similar to the state of affairs in captioning and machine translation,
    it is an open problem to automatically evaluate the quality of free-form answers.
    Existing metrics such as BLEU, METEOR, ROUGE are known to correlate poorly with human
    judgement in evaluating \dialog responses~\cite{liu_emnlp16}.

    Instead of evaluating on a downstream task~\cite{bordes_arxiv16}
    or holistically evaluating the entire conversation (as in goal-free chit-chat~\cite{alexa}),
    we evaluate \emph{individual responses} 
    at each round ($t=1,2,\ldots,10$) in a retrieval or multiple-choice setup.

Specifically, at test time, a \vd system
is given an image $I$, the `ground-truth' \dialog history (including the image caption) $C, (Q_1, A_1), \ldots, (Q_{t-1}, A_{t-1})$,
the question $Q_t$, \emph{and} a list of $N=100$ candidate answers, and asked to return a sorting of the
candidate answers. The model is evaluated on retrieval metrics --
(1) rank of human response (lower is better),
(2) recall@$k$, \ie existence of the human response in top-$k$ ranked responses, and
(3) mean reciprocal rank (MRR) of the human response (higher is better).



The evaluation protocol is compatible with both discriminative models (that simply score the input candidates,
\eg via a softmax over the options, and cannot generate new answers), and
generative models (that generate an answer string, \eg via Recurrent Neural Networks)
by ranking the candidates by the model's log-likelihood scores.



\textbf{Candidate Answers.}
We generate a candidate set of correct and incorrect answers from four sets: \\
\textbf{Correct}: The ground-truth human response to the question. \\
\textbf{Plausible}: Answers to 50 most similar questions. 
Similar questions are those that start with similar tri-grams and mention similar semantic concepts in the rest of the question. To capture this, all questions are embedded into a vector space by concatenating the GloVe embeddings of the first three words with the averaged GloVe embeddings of the remaining words in the questions.  Euclidean distances are used to compute
neighbors.
Since these neighboring questions were asked on different images,
their answers serve as `hard negatives'. \\
\textbf{Popular}: The 30 most popular answers from the dataset -- \eg
`yes', `no', `2', `1', `white', `3', `grey', `gray', `4', `yes it is'.
The inclusion of popular answers forces the machine to pick between likely \emph{a priori} responses
and plausible responses for the question, thus increasing the task difficulty. \\
%
\textbf{Random}: The remaining are answers to random questions in the dataset.
To generate 100 candidates, we first find the union of the correct, plausible, and popular answers, and
include random answers until a unique set of 100 is found.

\section{Neural \vdfull Models}
\label{sec:models}

    In this section, we develop a number of neural \vdfull answerer models.
Recall that the model is given as input -- an image $I$,
the `ground-truth' \dialog history (including the image caption)
$H = (\underbrace{\vphantom{(Q_1, A_1)}C}_{H_0}, \underbrace{(Q_1, A_1)}_{H_1}, \ldots, \underbrace{(Q_{t-1}, A_{t-1})}_{H_{t-1}})$,
the question $Q_t$, and a list of 100 candidate answers $\calA_t = \{A^{(1)}_t, \ldots,A^{(100)}_t\}$ --
and asked to return a sorting of $\calA_t$. 

At a high level, all our models follow the encoder-decoder framework, \ie factorize into two parts --
(1) an \emph{encoder} that converts the input $(I, H, Q_t)$ into a vector space, and
(2) a \emph{decoder} that converts the embedded vector into an output.
We describe choices for each component next and present experiments with all encoder-decoder combinations.

\textbf{Decoders:}
We use two types of decoders:

\begin{compactitem}

\item \textbf{Generative} (LSTM) decoder:
where the encoded vector is set as the initial state of the Long Short-Term Memory (LSTM) RNN language model.
During training, we maximize the log-likelihood of the ground truth answer sequence given its
corresponding encoded representation (trained end-to-end).
To evaluate, we use the model's log-likelihood scores and rank candidate answers.

Note that this decoder does not need to score options during training. As a result, such models
do not exploit the biases in option creation and typically underperform models that do~\cite{jabri_eccv16},
but it is debatable whether exploiting such biases is really indicative of progress.
Moreover, generative decoders are more practical in that they can actually be deployed in realistic applications.

\item \textbf{Discriminative} (softmax) decoder: computes dot product similarity between
input encoding and an LSTM encoding of each of the answer options.
These dot products are fed into a softmax to compute the posterior probability over options.
During training, we maximize the log-likelihood of the correct option.
During evaluation, options are simply ranked based on their posterior probabilities.

\end{compactitem}


\textbf{Encoders:}
We develop 3 different encoders (listed below) that
convert inputs $(I, H, Q_t)$ into a joint representation.
In all cases, we 
represent $I$ via the $\ell2$-normalized activations from the penultimate layer of VGG-16~\cite{simonyan_iclr15}.
For each encoder $E$, we experiment with all possible ablated versions: $E(Q_t), E(Q_t,I), E(Q_t, H), E(Q_t, I, H)$
(for some encoders, not all combinations are `valid'; details below).

%
\begin{compactitem}

\item \textbf{\lffull (\lf) Encoder}:
In this encoder, we treat $H$ as a long string with the entire history $(H_0, \ldots, H_{t-1})$ concatenated.
$Q_t$ and $H$ are separately encoded with 2 different LSTMs, and
individual representations of participating inputs $(I, H, Q_t)$ are concatenated and
linearly transformed to a desired size of joint representation.

\item \textbf{\hrefull (\hre)}:
In this encoder, we capture the intuition that there is a hierarchical nature to our problem -- each question $Q_t$ is
a sequence of words that need to be embedded, and the \dialog as a whole is a sequence of question-answer pairs $(Q_t, A_t)$.
Thus, similar to \cite{serban_arxiv16}, as shown in \figref{fig:hre}, we propose an \hre model that contains a \dialog-RNN sitting on top of
a recurrent block ($R_t$). The recurrent block $R_t$ embeds the question and image jointly via an LSTM (early fusion),
embeds each round of the history $H_t$,
and passes a concatenation of these to the \dialog-RNN above it. The \dialog-RNN produces
both an encoding for this round ($E_t$ in \figref{fig:hre}) and a \dialog context to pass onto the next round.
We also add an attention-over-history (`Attention' in \figref{fig:hre}) mechanism allowing
the recurrent block $R_t$ to choose and attend to the round of the history relevant to the current question.
This attention mechanism consists of a softmax over previous rounds ($0,1,\ldots, t-1$)
computed from the history and question+image encoding.

\begin{figure}[h]
    \centering
    \includegraphics[width=\columnwidth]{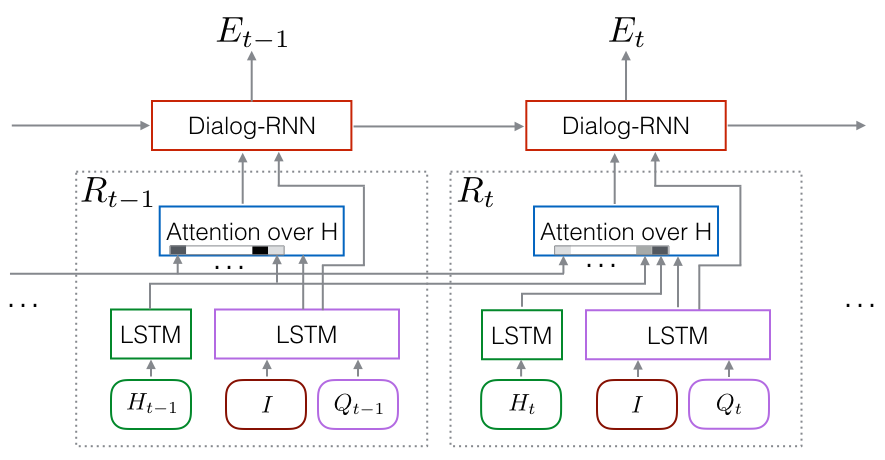}
    \caption{Architecture of HRE encoder with attention. At the current round $R_t$, the model has the capability to choose and attend to relevant history from previous rounds, based on the current question. This attention-over-history feeds into a dialog-RNN along with question to generate joint representation $E_t$ for the decoder.}
    \label{fig:hre}
\end{figure}


\item \textbf{\mnfull (\mn) Encoder}:
%
We develop a \mn encoder that maintains each previous question and answer as
a `fact' in its memory bank and learns to refer to the stored facts and image to answer the question.
Specifically, we encode $Q_t$ with an LSTM to get a $512$-d vector,
encode each previous round of history $(H_0, \ldots, H_{t-1})$ with another LSTM to get a $t\times512$ matrix.
We compute inner product of question vector with each history vector to get scores over previous rounds, which are fed to
a softmax to get attention-over-history probabilities.
Convex combination of history vectors using these attention probabilities gives us the `context vector', which
is passed through an fc-layer and added to the question vector
to construct the \mn encoding. In the language of \mnfull~\cite{bordes_arxiv16}, this is a `1-hop' encoding.

\end{compactitem}

We use a `[encoder]-[input]-[decoder]' convention to refer to model-input combinations.
For example, `\lf-QI-D' has a \lffull encoder with question+image inputs (no history), and a discriminative decoder.
Implementation details about the models can be found in the supplement.

\vspace{-8pt}
\section{Experiments}
\label{sec:exp}

\textbf{Splits.}
    \vd v0.9 contains 83k \dialogs on COCO-\train and 40k on COCO-\val images.
    We split the 83k into 80k for training, 3k for validation, and use the 40k as test.

Data preprocessing, hyperparameters and training details
are included in the supplement.

\textbf{Baselines}
We compare to a number of baselines:
\textbf{Answer Prior:}
Answer options to a test question are encoded with an LSTM and scored by a linear classifier. This captures
ranking by frequency of answers in our training set without resolving to exact string matching.
\textbf{NN-Q:} Given a test question, we find $k$ nearest neighbor questions (in GloVe space) from train,
and score answer options by their mean-similarity with these $k$ answers.
\textbf{NN-QI:} First, we find $K$ nearest neighbor questions for a test question.
Then, we find a subset of size $k$ based on image feature similarity.
Finally, we rank options by their mean-similarity to answers to these $k$ questions.
We use $k=20, K=100$.

Finally, we adapt several (near) state-of-art VQA models
(SAN~\cite{yang_cvpr16}, HieCoAtt~\cite{lu_nips16})
to \vdfull.
Since VQA is posed as classification, we `chop' the final VQA-answer softmax from these models,
feed these activations to our discriminative decoder (\secref{sec:models}), and train end-to-end on \vd.
Note that our \lf-QI-D model is similar to that in \cite{lu_github15}.
Altogether, these form fairly sophisticated baselines.

\begin{table}[t]
{
    \small
    \setlength\tabcolsep{3.8pt}
    \centering
    \begin{tabular}{ccccccc}
    \toprule
    & \textbf{Model} & \textbf{MRR} & \textbf{R@1} & \textbf{R@5} & \textbf{R@10} & \textbf{Mean} \\
    \midrule
    \multirow{3}{*}{\rotatebox[origin=c]{90}{Baseline} $\begin{dcases} \\ \\ \\ \end{dcases}$}
    &Answer prior & 0.3735 & 23.55 & 48.52 & 53.23 & 26.50 \\[1.5pt]
    &NN-Q & 0.4570 & 35.93 & 54.07 & 60.26 & 18.93 \\[1.5pt]
    &NN-QI & 0.4274 & 33.13 & 50.83 & 58.69 & 19.62 \\[1.5pt]
    \midrule
    \multirow{9}{*}{\rotatebox[origin=c]{90}{Generative} $\begin{dcases}\\  \\ \\ \\  \\ \\ \\ \end{dcases}$}
    &\lf-Q-G & 0.5048 & 39.78 & 60.58 & 66.33 & 17.89 \\
    &\lf-QH-G & 0.5055 & 39.73 & 60.86 & 66.68 & 17.78 \\
    &\lf-QI-G & 0.5204 & 42.04 & 61.65 & 67.66 & 16.84 \\
    &\lf-QIH-G & 0.5199 & 41.83 & 61.78 & 67.59 & 17.07 \\
    \cdashline{2-7}
    & \hre-QH-G & 0.5102 & 40.15 & 61.59 & 67.36 & 17.47 \\
    &\hre-QIH-G & 0.5237 & \textbf{42.29} & 62.18 & 67.92 & 17.07 \\
    &\hre{}A-QIH-G & 0.5242 & 42.28 & 62.33 & 68.17 & \textbf{16.79} \\
    \cdashline{2-7}
    &\mn-QH-G & 0.5115 & 40.42 & 61.57 & 67.44 & 17.74 \\
    &\mn-QIH-G & \textbf{0.5259} & \textbf{42.29} & \textbf{62.85} & \textbf{68.88} & 17.06 \\
    \midrule
    \multirow{9}{*}{\rotatebox[origin=c]{90}{Discriminative} $\begin{dcases} \\ \\ \\  \\ \\ \\  \\ \end{dcases}$}
    &\lf-Q-D & 0.5508 & 41.24 & 70.45 & 79.83 & 7.08 \\
    &\lf-QH-D & 0.5578 & 41.75 & 71.45 & 80.94 & 6.74 \\
    &\lf-QI-D & 0.5759 & 43.33 & 74.27 & 83.68 & 5.87  \\
    &\lf-QIH-D & 0.5807 & 43.82 & 74.68 & 84.07 & 5.78 \\
    \cdashline{2-7}
    &\hre-QH-D & 0.5695 & 42.70 & 73.25 & 82.97 & 6.11 \\
    &\hre-QIH-D & 0.5846 & 44.67 & 74.50 & 84.22 & 5.72 \\
    &\hre{}A-QIH-D & 0.5868 & 44.82 & 74.81 & 84.36 & 5.66 \\
    \cdashline{2-7}
    &\mn-QH-D & 0.5849 & 44.03 & 75.26 & 84.49 & 5.68 \\
    &\mn-QIH-D & \textbf{0.5965} & \textbf{45.55} & \textbf{76.22} & \textbf{85.37} & \textbf{5.46} \\
    \midrule
    \multirow{2}{*}{\rotatebox[origin=c]{90}{VQA} $\begin{dcases} \\ \end{dcases}$} &
    SAN1-QI-D & 0.5764 & 43.44 & 74.26 & 83.72 & 5.88 \\
    &HieCoAtt-QI-D & 0.5788 & 43.51 & 74.49 & 83.96 & 5.84 \\
    \bottomrule
    \end{tabular}
    \caption{Performance of methods on \vd v0.9, measured by mean reciprocal rank (MRR), recall@$k$ and mean rank. Higher is better for MRR and recall@k, while lower is better for mean rank.
    Performance on \vd v0.5 is included in the supplement.}
    \vspace{-15pt}
    \label{table:model_results}
}
\end{table}

\textbf{Results.}
    \reftab{table:model_results} shows results for our models and baselines on \vd v0.9 (evaluated on 40k from COCO-\val).

A few key takeaways --
    1) As expected, all learning based models significantly outperform non-learning baselines.
    2) All discriminative models significantly outperform generative models,
    which as we discussed is expected since discriminative models can tune to the biases in the answer options.
    3) Our best generative and discriminative models are
    \mn-QIH-G with 0.526 MRR,
    and \mn-QIH-D with 0.597 MRR.
    4) We observe that naively incorporating history doesn't help much (\lf-Q \vs \lf-QH and \lf-QI \vs \lf-QIH)
    or can even hurt a little (\lf-QI-G \vs \lf-QIH-G).
    However, models that better encode history (\mn/\hre) perform better than corresponding \lf models with/without history
    (\eg \lf-Q-D vs. \mn-QH-D).
    5) Models looking at $I$ (\{\lf,\mn,\hre\}-QIH) outperform corresponding blind models (without $I$).
\textbf{Human Studies.}
    We conduct studies on AMT to quantitatively evaluate human performance on this task for all combinations
    of \{with image, without image\}$\times$\{with history, without history\}.
    We find that without image, humans perform better when they have access to dialog history. As expected, this gap narrows down when they have access to the image.
    Complete details can be found in supplement.

\vspace{\sectionReduceBot}
\section{Conclusions}
\label{sec:conclusions}


    To summarize, we introduce a new AI task -- \vdfull, where an AI agent must hold a \dialog with a human about visual content.
    We develop a novel two-person chat data-collection protocol to curate a large-scale dataset (\vd), propose
    retrieval-based evaluation protocol, and develop a family of encoder-decoder models for \vdfull.
    We quantify human performance on this task via human studies.
    Our results indicate that there is significant scope for improvement, and we
    believe this task can serve as a testbed for measuring progress towards visual intelligence.
\section{Acknowledgements}
\label{sec:ack}

    We thank Harsh Agrawal, Jiasen Lu for help with AMT data collection;
    Xiao Lin, Latha Pemula for model discussions;
    Marco Baroni, Antoine Bordes, Mike Lewis, Marc'Aurelio Ranzato for helpful discussions.
    We are grateful to the developers of Torch~\cite{torch} for building an excellent framework.
    This work was funded in part by
    NSF CAREER awards to DB and DP,
    ONR YIP awards to DP and DB, ONR Grant N00014-14-1-0679 to DB,
    a Sloan Fellowship to DP,
    ARO YIP awards to DB and DP,
    an Allen Distinguished Investigator award to DP from the Paul G. Allen Family Foundation,
    ICTAS Junior Faculty awards to DB and DP,
    Google Faculty Research Awards to DP and DB,
    Amazon Academic Research Awards to DP and DB,
    AWS in Education Research grant to DB, and NVIDIA GPU donations to DB.
    SK was supported by ONR Grant N00014-12-1-0903.
    The views and conclusions contained herein are those of the authors and should not be interpreted as necessarily representing the
    official policies or endorsements, either expressed or implied, of the U.S. Government, or any sponsor.

\clearpage

\appendix
\section*{Appendix Overview}
This supplementary document is organized as follows:

\begin{itemize}
\item \refsec{sec:vd_as_qaplusplus} studies how and why \vd is more than just a collection of independent Q\&As.

\item \refsec{sec:qual} shows qualitative examples from our dataset.

\item \refsec{sec:human_supp} presents detailed human studies along with comparisons to machine accuracy.
The interface for human studies is demonstrated in a video\footnote{\url{ https://goo.gl/yjlHxY}}.

\item \refsec{sec:interface} shows snapshots of our two-person chat data-collection interface on Amazon Mechanical Turk.
The interface is also demonstrated in the video\footnotemark[3].

\item \refsec{sec:analysis} presents further analysis of \vd, such as question types, question and answer lengths per question type.
A video with an interactive sunburst visualization of the dataset is included\footnotemark[3].

\item \refsec{sec:vdot5-results} presents performance of our models on \vd v0.5 \test.

\item \refsec{sec:exp-details} presents implementation-level training details including data preprocessing, and model architectures.

\item Putting it all together, we compile a video demonstrating
our visual chatbot\footnotemark[3] that answers a sequence of questions from a user about an image.
This demo uses one of our best generative models from the main paper, MN-QIH-G, and uses
sampling (without any beam-search) for inference in the LSTM decoder. Note that these
videos demonstrate an `unscripted' \dialog ~-- in the sense that the particular QA sequence is not present in
\vd and the model is not provided with any list of answer options.

\end{itemize}


\section{In what ways are \dialogs in \vd more than just 10 visual Q\&As?}
\label{sec:vd_as_qaplusplus}

In this section, we lay out an exhaustive list of differences between \vd and image question-answering datasets,
with the VQA dataset~\cite{antol_iccv15} serving as the representative.

In essence, we characterize what makes an instance
in \vd more than a collection of 10 independent question-answer pairs about an image -- \emph{what makes it a \dialog}.

In order to be self-contained and an exhaustive list, some parts of this section repeat content from the main document.

\subsection{\vd has longer free-form answers}

\begin{figure}[h]
	\centering
	\begin{subfigure}[b]{0.48\columnwidth}
		\centering
		\includegraphics[width=\textwidth]{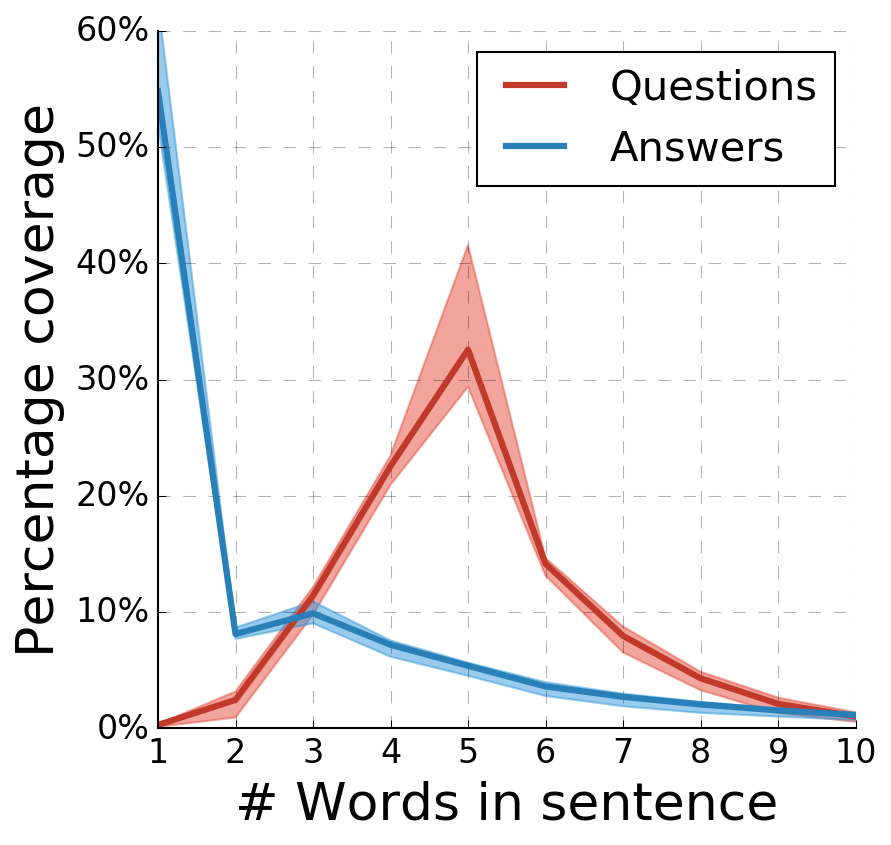}
		\caption{}
		\label{fig:quesAnsLengths}
	\end{subfigure}
	\begin{subfigure}[b]{0.48\columnwidth}
		\centering
		\includegraphics[width=\textwidth]{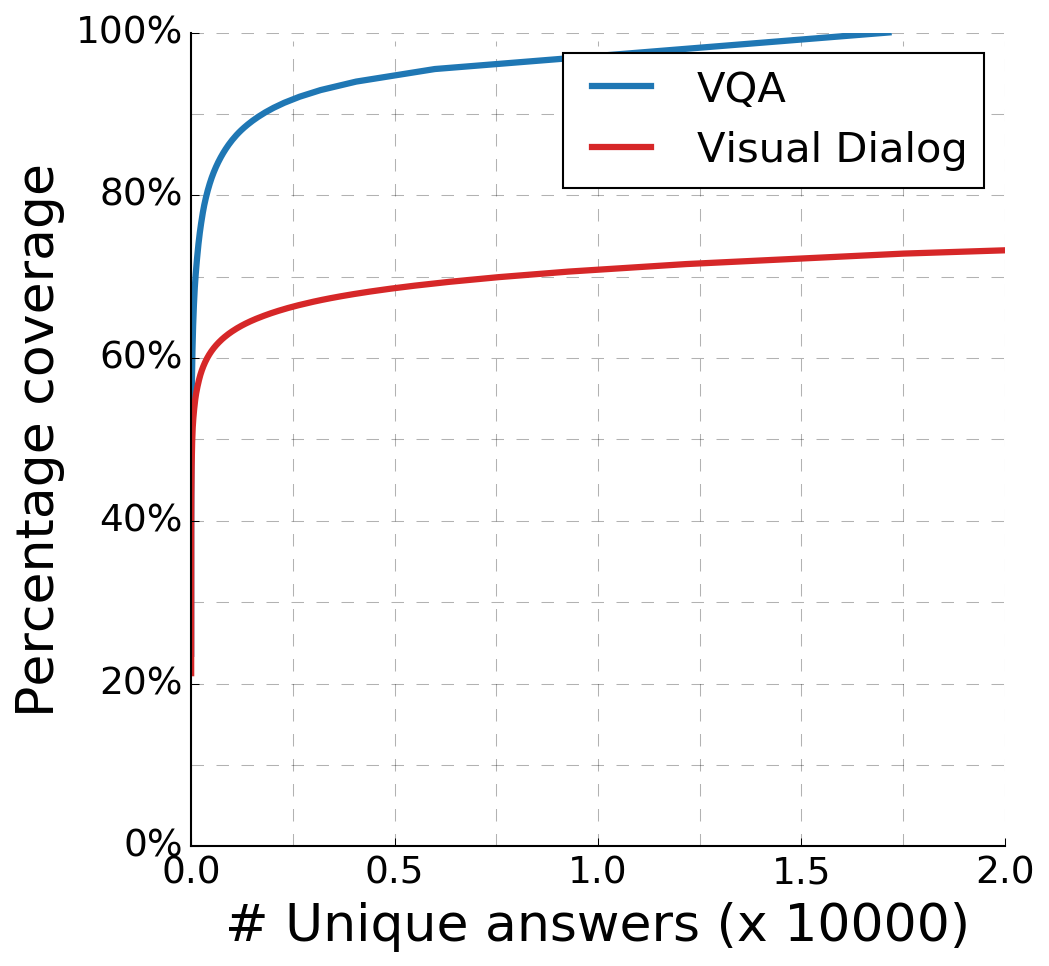}
		\caption{}
		\label{fig:coverage}
	\end{subfigure}
	\quad
	\caption{Distribution of lengths for questions and answers (left); and percent coverage of unique answers over all answers from the train dataset (right), compared to VQA. For a given coverage, \vd has more unique answers indicating greater answer diversity.}
	\label{fig:svqa_amt}
\end{figure}

\reffig{fig:quesAnsLengths} shows the distribution of answer lengths in \vd.
and \reftab{table:comparisons_between_datasets} compares statistics of \vd with existing image question answering datasets.
Unlike previous datasets, answers in \vd are longer, conversational, and more descriptive --
mean-length 2.9 words (\vd) vs 1.1 (\vqa), 2.0 (Visual 7W), 2.8 (Visual Madlibs).
Moreover, $37.1\%$ of answers in \vd are longer than 2 words while the \vqa dataset has only $3.8\%$ answers longer than 2 words.




\begin{table*}[t]
\small
\centering
\begin{tabular}{lrrccccccccc}
\hline
& \textbf{\# QA} & \textbf{\# Images} & \textbf{Q Length} & \textbf{A Length} & \textbf{A Length > 2} & \textbf{Top-1000 A} & \textbf{Human Accuracy} \\
\hline
DAQUAR~\cite{malinowski_nips14} & 12,468 & 1,447 & 11.5$\,\pm\,$2.4 & 1.2$\,\pm\,$0.5 & 3.4\% & 96.4\% & - \\
Visual Madlibs~\cite{yu_iccv15} & 56,468 & 9,688 & 4.9$\,\pm\,$2.4 & 2.8$\,\pm\,$2.0 & 47.4\% & 57.9\% & - \\
COCO-QA~\cite{ren_nips15} & 117,684 & 69,172 & 8.7$\,\pm\,$2.7 & 1.0$\,\pm\,$0 & 0.0\% & 100\% & - \\
Baidu~\cite{gao_nips15} & 316,193 & 316,193 & - & - & - & - & - \\
VQA~\cite{antol_iccv15} & 614,163 & 204,721 & 6.2$\,\pm\,$2.0 & 1.1$\,\pm\,$0.4 & 3.8\% & 82.7\% & $\checkmark$ \\
Visual7W~\cite{zhu_cvpr16} & 327,939 & 47,300 & 6.9$\,\pm\,$2.4 & 2.0$\,\pm\,$1.4 & 27.6\% & 63.5\% & $\checkmark$ \\
\hline
\vd (Ours) & 1,232,870 & 123,287 & 5.1$\,\pm\,$0.0 & 2.9$\,\pm\,$0.0 & 37.1\% & 63.2\% & $\checkmark$ \\
\hline
\end{tabular}
\caption{Comparison of existing image question answering datasets with \vd}
\label{table:comparisons_between_datasets}
\vspace{-5pt}
\end{table*}

\figref{fig:coverage} shows the cumulative coverage of all answers (y-axis) by the most frequent answers (x-axis).
The difference between \vd and \vqa is stark -- the top-1000 answers in VQA cover $\sim$83\% of all answers,
while in \vd that figure is only $\sim$63\%. There is a significant heavy tail of answers in \vd~-- most long strings are unique,
and thus the coverage curve in \figref{fig:coverage} becomes a straight line with slope 1.
In total, there are 337,527 unique answers in \vd (out of the 1,232,870 answers currently in the dataset).

\subsection{\vd has co-references in \dialogs}

People conversing with each other tend to use pronouns to refer to already mentioned entities.
Since language in \vd is the result of a sequential conversation, it naturally contains pronouns --
`he', `she', `his', `her', `it', `their', `they', `this', `that', `those',  \etc.
In total, $38\%$ of questions, $19\%$ of answers, and \emph{nearly all} ($98\%$) \dialogs contain at least one pronoun,
thus confirming that a machine will need to overcome coreference ambiguities
to be successful on this task.
As a comparison, only $9\%$ of questions and $0.25\%$ of answers in \vqa contain at least one pronoun.

%

\begin{figure}[h]
\includegraphics[width=\linewidth]{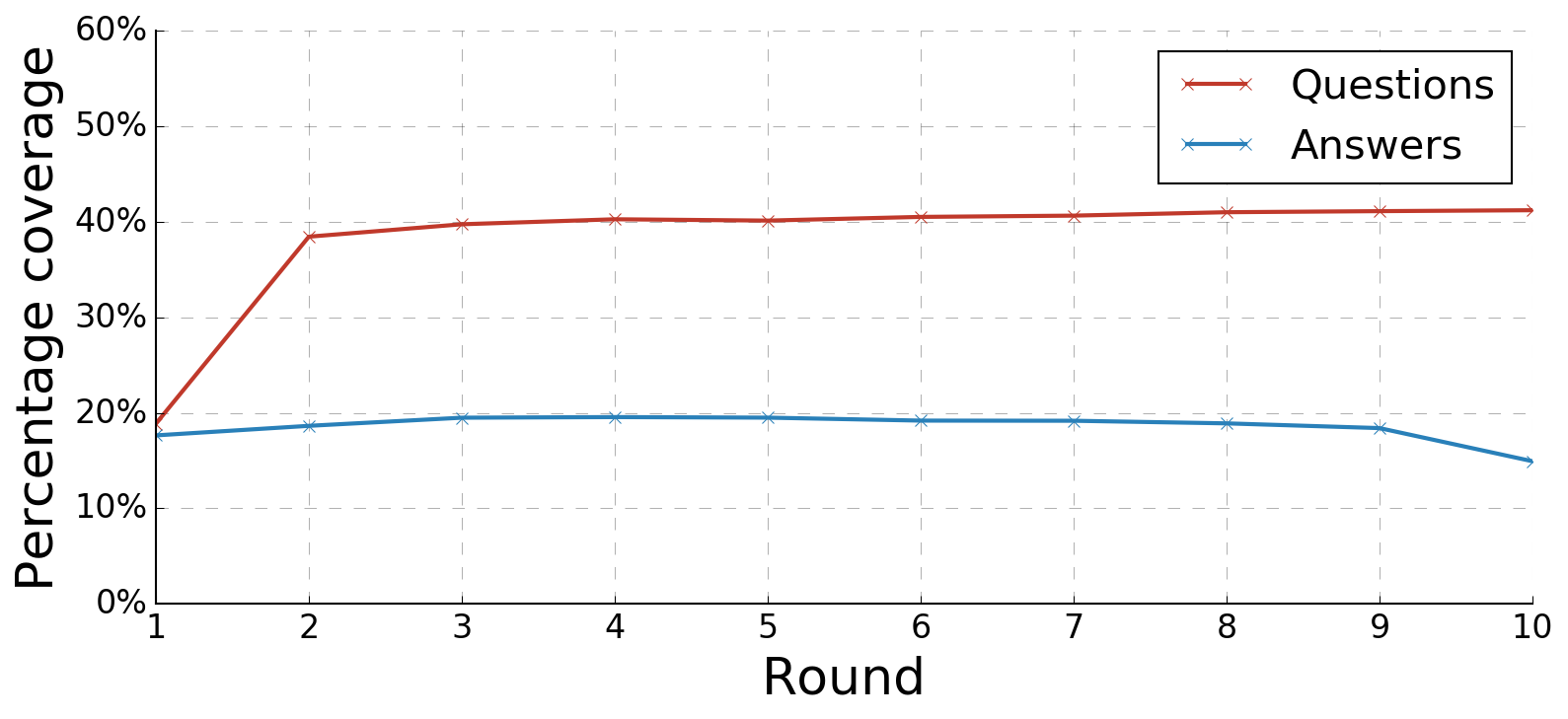}
\caption{Percentage of QAs with pronouns for different rounds.
In round 1, pronoun usage in questions is low (in fact, almost equal to usage in answers).
From rounds 2 through 10, pronoun usage is higher in questions and fairly consistent across rounds.}
\label{fig:quesPronouns}
\vspace{-10pt}
\end{figure}

In \figref{fig:quesPronouns}, we see that pronoun usage is lower in the first round compared to other rounds,
which is expected since there are fewer entities
to refer to in the earlier rounds. The pronoun usage is also generally lower in answers than questions, which is also understandable
since the answers are generally shorter than questions and thus less likely to contain pronouns.
In general, the pronoun usage is
fairly consistent across rounds (starting from round 2) for both questions and answers.

\subsection{\vd has smoothness/continuity in `topics'}

\paragraph{Qualitative Example of Topics.}
There is a stylistic difference in the questions asked in \vd
(compared to the questions in \vqa) due to the nature of the task assigned
to the subjects asking the questions.
In \vqa, subjects saw the image and were asked to ``stump a smart robot''. Thus, most queries
involve specific details, often about the background
(Q: \myquote{What program is being utilized in the background on the computer?}).
In \vd, questioners did not see the original image and were asking questions to build a mental model of the scene.
Thus, the questions tend to be open-ended, and often follow a pattern:

\begin{compactitem}

\item Generally starting with the \textbf{entities in the caption}:

\begin{flushright}
\myquote{An elephant walking away from a pool in an exhibit}, \\
\myquote{Is there only 1 elephant?},
\end{flushright}

\item digging deeper into their \textbf{parts, attributes, or properties}:

\begin{flushright}
\myquote{Is it full grown?}, 
\myquote{Is it facing the camera?},
\end{flushright}

\item asking about the \textbf{scene category or the picture setting}:

\begin{flushright}
\myquote{Is this indoors or outdoors?}, \myquote{Is this a zoo?},
\end{flushright}

\item \textbf{the weather}:

\begin{flushright}
\myquote{Is it snowing?}, \myquote{Is it sunny?},
\end{flushright}


\item simply \textbf{exploring the scene}:

\begin{flushright}
\myquote{Are there people?},
\myquote{Is there shelter for elephant?},
\end{flushright}

\item and asking \textbf{follow-up questions} about the new visual entities discovered
from these explorations:

\begin{flushright}
\myquote{There's a blue fence in background, like an enclosure}, \\
\myquote{Is the enclosure inside or outside?}.
\vspace{-5pt}
\end{flushright}

\end{compactitem}

Such a line of questioning does not exist in the \vqa dataset, where the subjects were shown the questions already asked about an image,
and explicitly instructed to ask about \emph{different entities}~\cite{antol_iccv15}.

\paragraph{Counting the Number of Topics.}
In order to quantify these qualitative differences, we performed a human study where we manually annotated
question `topics' for $40$ images (a total of $400$ questions), chosen randomly from the \val set.
The topic annotations were based on human judgement with a consensus of 4 annotators, with topics
such as: asking about a particular object (\myquote{What is the man doing?}),
the scene (\myquote{Is it outdoors or indoors?}), the weather (\myquote{`Is the weather sunny?}),
the image (\myquote{Is it a color image?}), and exploration (\myquote{Is there anything else?'}).
We performed similar topic annotation for questions from VQA for the same set of $40$ images,
and compared topic continuity in questions. 

Across $10$ rounds, \vd questions have $4.55 \pm 0.17$ topics on average,
confirming that these are not $10$ independent questions.
Recall that \vd has $10$ questions per image as opposed to $3$ for VQA.
Therefore, for a fair comparison, we compute average number of topics in \vd over all `sliding windows'
of $3$ successive questions.
For $500$ bootstrap samples of batch size $40$, \vd has $2.14 \pm 0.05$ topics while VQA has $2.53 \pm 0.09$.
Lower mean number of topics suggests there is more continuity in \vd because questions do not change topics as often.

\paragraph{Transition Probabilities over Topics.}
We can take this analysis a step further by computing topic transition probabilities over topics as follows.
For a given sequential \dialog exchange, we now count the number of topic transitions between
consecutive QA pairs, normalized by the total number of possible transitions between rounds (9 for \vd and 2 for \vqa).
We compute this `topic transition probability' (how likely are two successive QA pairs to be about two different topics)
for \vd and \vqa in two different settings -- (1) in-order and (2) with a permuted sequence of QAs.
Note that if \vd were simply a collection of 10 independent QAs as opposed to a \dialog, we would expect the topic
transition probabilities to be similar for in-order and permuted variants.
However, we find that for 1000 permutations of 40 topic-annotated image-\dialogs,
in-order-\vd has an average topic transition probability of $0.61$, while permuted-\vd has $0.76 \pm 0.02$.
In contrast, \vqa has a topic transition probability of $0.80$ for in-order \vs $0.83 \pm 0.02$ for permuted QAs.

There are two key observations:
(1) In-order transition probability is lower for \vd than \vqa (\ie topic transition is less likely in \vd), and
(2) Permuting the order of questions results in a larger increase for \vd, around $0.15$, compared to a mere $0.03$ in case of \vqa
(\ie in-order-\vqa and permuted-\vqa behave significantly more similarly than in-order-\vd and permuted-\vd).

Both these observations establish that there is smoothness in the temporal order of topics in \vd, which is
indicative of the narrative structure of a \dialog, rather than independent question-answers.

\subsection{\vd has the statistics of an NLP \dialog dataset}

In this analysis, our goal is to measure whether \vd \emph{behaves like a dialog dataset}.

In particular, we compare \vd, \vqa, and Cornell Movie-Dialogs Corpus~\cite{mizil_cmcl11}.
The Cornell Movie-Dialogs corpus is a text-only dataset extracted from pairwise interactions between characters
from approximately $617$ movies, and is widely used as a standard dialog corpus in the natural language
processing (NLP) and \dialog communities.

One popular evaluation criteria used in the \dialog-systems research community is the \emph{perplexity} of language
models trained on \dialog datasets -- the lower the perplexity of a model, the better it has learned the structure in the \dialog
dataset.

For the purpose of our analysis, we pick the popular sequence-to-sequence (\seqtoseq) language model~\cite{sutskever_nips14}
and use the perplexity of this model trained on different datasets as a measure
of temporal structure in a dataset.

As is standard in the \dialog literature, we train the \seqtoseq model to predict the probability of utterance $U_t$
given the previous utterance $U_{t-1}$, \ie $\textbf{P}(U_t \mid U_{t-1})$ on the Cornell corpus.
For \vd and \vqa, we train the \seqtoseq model to predict the probability of a question $Q_t$
given the previous question-answer pair, \ie $\textbf{P}(Q_t \mid (Q_{t-1}, A_{t-1}))$.

For each dataset, we used its \texttt{train} and \texttt{val} splits for training and hyperparameter tuning respectively, and report results on \texttt{test}.
At test time, we only use conversations of length $10$ from Cornell corpus for a fair comparison to \vd (which has 10 rounds of QA).

For all three datasets, we created $100$ permuted versions of \texttt{test},
where either QA pairs or utterances are randomly shuffled to disturb their natural order. This allows us to compare
datasets in their natural ordering \wrt permuted orderings. Our hypothesis is that since \dialog datasets have linguistic structure in
the sequence of QAs or utterances they contain, this structure will be significantly affected by permuting the sequence.
In contrast, a collection of independent question-answers (as in VQA) will not be significantly affected by a permutation.

\reftab{tab:ques-generation} compares the original, unshuffled \texttt{test} with the shuffled testsets on two metrics:

\begin{table}[t]
\centering
\begin{tabular}{cccc}
\hline
\multirow{2}{*}{\textbf{Dataset}} & \multicolumn{2}{c}{\textbf{Perplexity Per Token}} & \multirow{2}{*}{\textbf{Classification}} \\
& Orig & Shuffled &  \\
\hline
\vqa & 7.83 & 8.16 $\pm$ 0.02 & 52.8 $\pm$ 0.9\\
Cornell (10) & 82.31 & 85.31 $\pm$ 1.51 & 61.0 $\pm$ 0.6\\
\vd (Ours) & 6.61 & 7.28 $\pm$ 0.01 & 73.3 $\pm$ 0.4\\
\hline
\end{tabular}
\caption{Comparison of sequences in \vd, \vqa, and Cornell Movie-Dialogs corpus in their original ordering \vs permuted `shuffled'
ordering. Lower is better for perplexity while higher is better for classification accuracy.
Left: the absolute increase in perplexity from natural to permuted ordering is highest in
the Cornell corpus ($3.0$) followed by \vd with $~0.7$, and \vqa at $~0.35$,
which is indicative of the degree of linguistic structure in the sequences in these datasets.
Right: The accuracy of a simple threshold-based classifier trained to differentiate between the original sequences and their permuted or shuffled versions.
A higher classification rate indicates the existence of a strong temporal continuity in the conversation,
thus making the ordering important.
We can see that the classifier on \vd achieves the highest accuracy ($73.3\%$), followed by Cornell ($61.0\%$).
Note that this is a binary classification task with the prior probability of each class by design being equal, thus
chance performance is $50\%$. The classifier on \vqa performs close to chance.
}
\label{tab:ques-generation}
\vspace{-10pt}
\end{table}

\paragraph{Perplexity:}
We compute the standard metric of \emph{perplexity per token},
\ie exponent of the normalized negative-log-probability of a sequence (where normalized is by the length of the sequence).
\reftab{tab:ques-generation} shows these perplexities for the original unshuffled \test and permuted \test sequences.

We notice a few trends.

First, we note that the absolute perplexity values are higher for the Cornell corpus than QA datasets.
We hypothesize that this is due to the broad, unrestrictive dialog generation task in Cornell corpus, which is a more difficult task
than question prediction about images, which is in comparison a more restricted task.

Second, in all three datasets, the shuffled \test has statistically significant higher perplexity than the original \test,
which indicates that shuffling does indeed break the linguistic structure in the sequences.

Third, the absolute increase in perplexity from natural to permuted ordering is highest in
the Cornell corpus ($3.0$) followed by our \vd with $~0.7$, and \vqa at $~0.35$,
which is indicative of the degree of linguistic structure in the sequences in these datasets.
Finally, the relative increases in perplexity are $3.64\%$ in Cornell, $10.13\%$ in \vd, and $4.21\%$ in \vqa~--
\vd suffers the highest relative increase in perplexity due to shuffling,
indicating the existence of temporal continuity that gets disrupted.

\paragraph{Classification:}

As our second metric to compare datasets in their natural \vs permuted order, we test whether we can reliably
classify a given sequence as natural or permuted.

Our classifier is a simple threshold on perplexity of a sequence. Specifically, given a pair of sequences,
we compute the perplexity of both from our \seqtoseq model, and predict that the one with higher perplexity is the
sequence in permuted ordering, and the sequence with lower perplexity is the one in natural ordering.
The accuracy of this simple classifier indicates how easy or difficult it is to tell the difference between
natural and permuted sequences.
A higher classification rate indicates existence of temporal continuity in the conversation,
thus making the ordering important.

\reftab{tab:ques-generation} shows the classification accuracies achieved on all datasets.
We can see that the classifier on \vd achieves the highest accuracy ($73.3\%$), followed by Cornell ($61.0\%$).
Note that this is a binary classification task with the prior probability of each class by design being equal, thus
chance performance is $50\%$. The classifiers on \vd and Cornell both significantly outperforming chance.
On the other hand, the classifier on \vqa is near chance ($52.8\%$), indicating a lack of general temporal continuity.

To summarize this analysis, our experiments show that \vd is significantly more dialog-like than \vqa,
and \emph{behaves} more like a
standard dialog dataset, the Cornell Movie-Dialogs corpus.

\subsection{\vd eliminates visual priming bias in \vqa}

One key difference between \vd and previous image question answering datasets (VQA~\cite{antol_iccv15},
Visual 7W~\cite{zhu_cvpr16}, Baidu mQA~\cite{gao_nips15}) is the lack of a `visual priming bias' in \vd.
Specifically, in all previous datasets, subjects saw an image while asking questions about it.
As described in \cite{zhang_cvpr16}, this leads to a particular bias in the questions --
people only ask \myquote{Is there a clocktower in the picture?} on pictures actually containing clock towers.
This allows language-only models to perform remarkably well on VQA and results in an inflated sense
of progress~\cite{zhang_cvpr16}.
As one particularly perverse example -- for questions in the VQA dataset starting with 
\myquote{Do you see a \ldots}, blindly answering \myquote{yes} without reading the rest of the question
or looking at the associated image results in an average VQA accuracy of $87\%$!
In \vd, questioners \emph{do not} see the image. As a result, this bias is reduced.

This lack of visual priming bias (\ie not being able to see the image while asking questions) and
holding a \dialog with another person while asking questions results in the following two unique features in \vd.

\begin{figure}[h]
\centering
\includegraphics[width=0.9\linewidth]{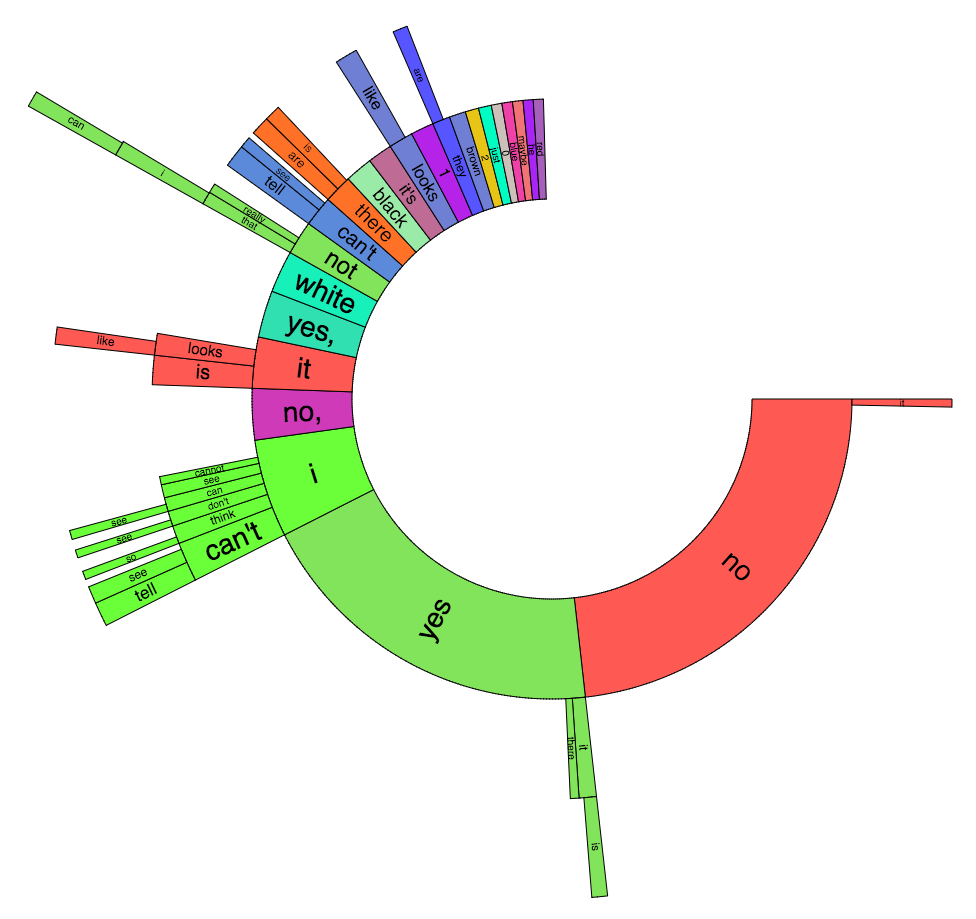}
\caption{Distribution of answers in \vd by their first four words.
The ordering of the words starts towards the center and radiates outwards.
The arc length is proportional to the number of questions containing the word.
White areas are words with contributions too small to show.}
\label{fig:ansCircle}
\end{figure}

\paragraph{Uncertainty in Answers in \vd.}
Since the answers in \vd are longer strings, we can visualize their distribution based
on the starting few words (\reffig{fig:ansCircle}). An interesting category of answers emerges
-- \myquote{I think so}, \myquote{I can't tell}, or \myquote{I can't see}  -- expressing doubt, uncertainty, or lack of information.
This is a consequence of the questioner not being able to see the image -- they are asking contextually relevant questions,
but not all questions may be answerable with certainty from that image. We believe this
is rich data for building more human-like AI that refuses to answer questions it doesn't have enough information to answer. 
See~\cite{ray_emnlp16} for a related, but complementary effort on question relevance in VQA.


\paragraph{Binary Questions $\ne$ Binary Answers in \vd.}
In VQA, binary questions are simply those with `yes', `no', `maybe' as answers~\cite{antol_iccv15}.
In \vd, we must distinguish between binary questions and binary answers.
Binary questions are those starting in `Do', `Did', `Have', `Has', `Is', `Are', `Was', `Were', `Can', `Could'.
Answers to such questions can
(1) contain only `yes' or `no',
(2) begin with `yes', `no', and contain additional information or clarification (Q: \myquote{Are there any animals in the image?},
A: \myquote{yes, 2 cats and a dog}),
(3) involve ambiguity (
\myquote{It's hard to see}, \myquote{Maybe}),
or
(4) answer the question without explicitly saying `yes' or `no'
(Q: \myquote{Is there any type of design or pattern on the cloth?},
A: \myquote{There are circles and lines on the cloth}).
We call answers that contain `yes' or `no' as binary answers~-- 149,367 and 76,346 answers in subsets (1) and (2) from above respectively.
%
Binary answers in \vqa are biased towards `yes'~\cite{antol_iccv15, zhang_cvpr16} --
61.40\% of yes/no answers are `yes'.
In \vd, the trend is reversed. Only 46.96\% are `yes' for all yes/no responses.
This is understandable since workers did not see the image, and were more likely to end up with negative responses.

\section{Qualitative Examples from \vd}
\label{sec:qual}

\begin{figure*}[ht]
\centering
\begin{subfigure}[t]{0.49\textwidth}
	\centering
	\includegraphics[trim={0cm 0cm 0cm 0cm},clip, width=\columnwidth]{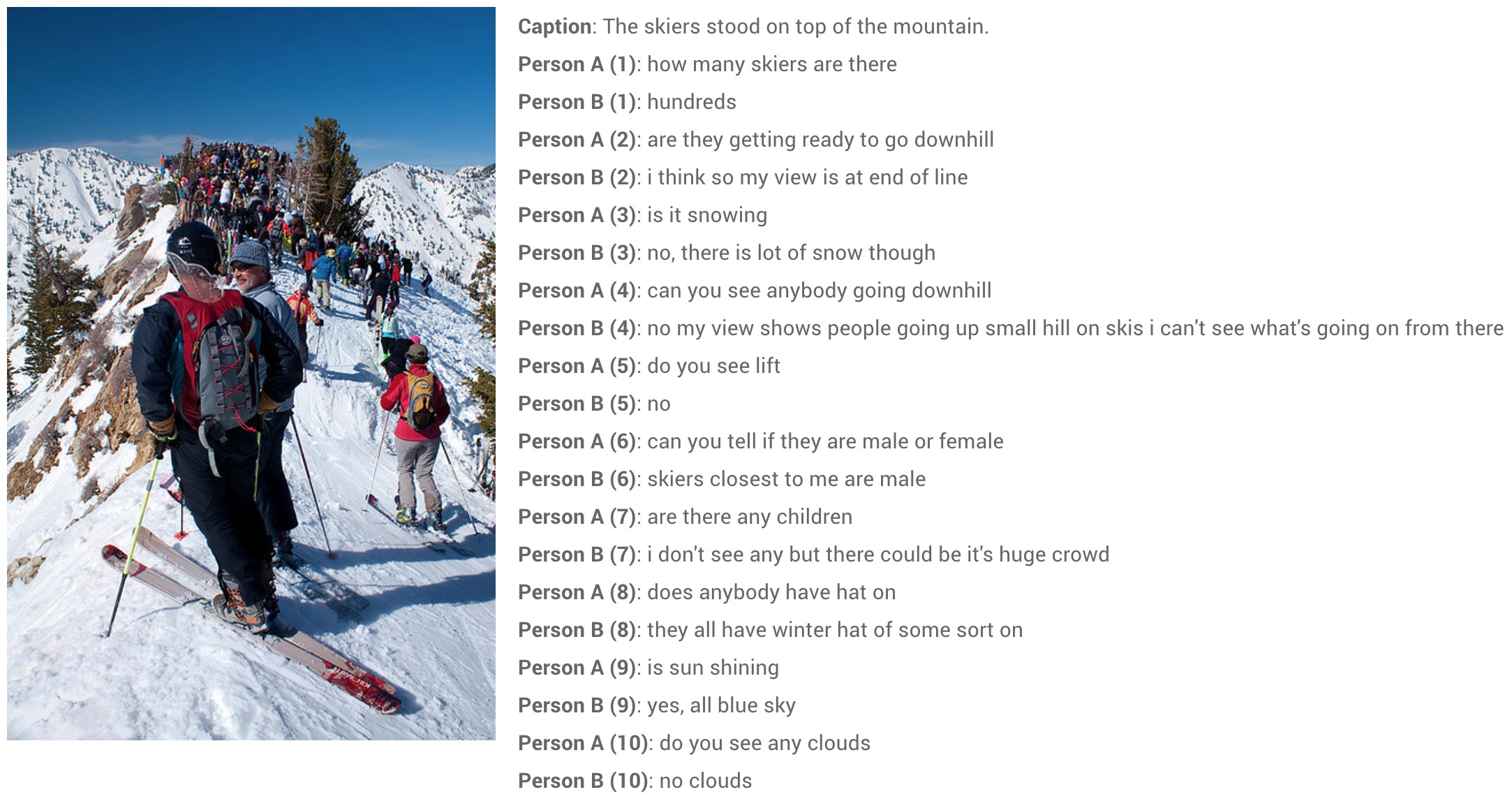}
	\caption{}
\end{subfigure}
\begin{subfigure}[t]{0.49\textwidth}
	\centering
	\includegraphics[trim={0cm 0cm 0cm 0cm},clip, width=\columnwidth]{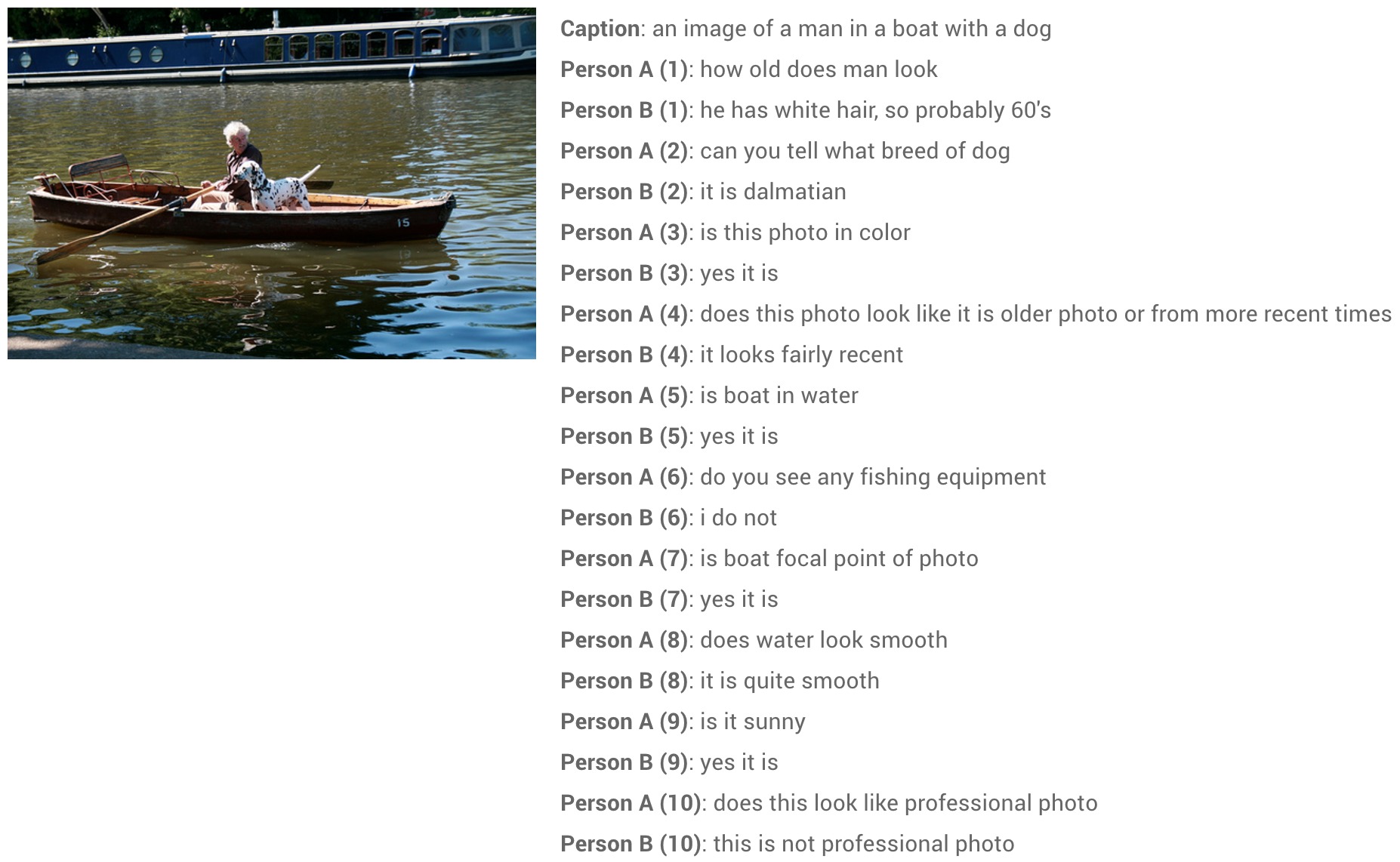}
	\caption{}
\end{subfigure}
\quad
\begin{subfigure}[t]{0.49\textwidth}
	\centering
	\includegraphics[trim={0cm 0cm 0cm 0cm},clip, width=\columnwidth]{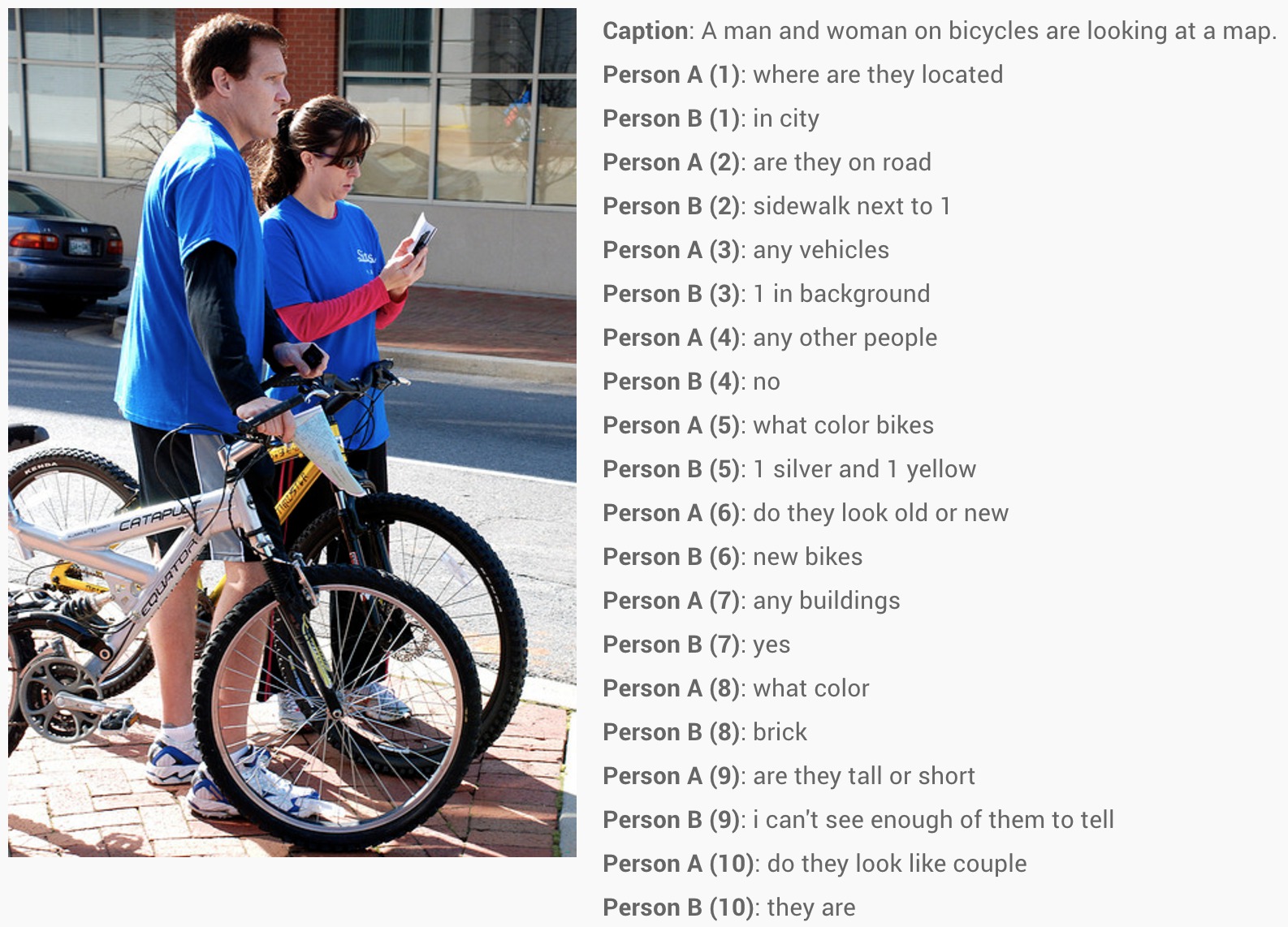}
	\caption{}
\end{subfigure}
\begin{subfigure}[t]{0.49\textwidth}
	\centering
	\includegraphics[trim={0cm 0cm 0cm 0cm},clip, width=\columnwidth]{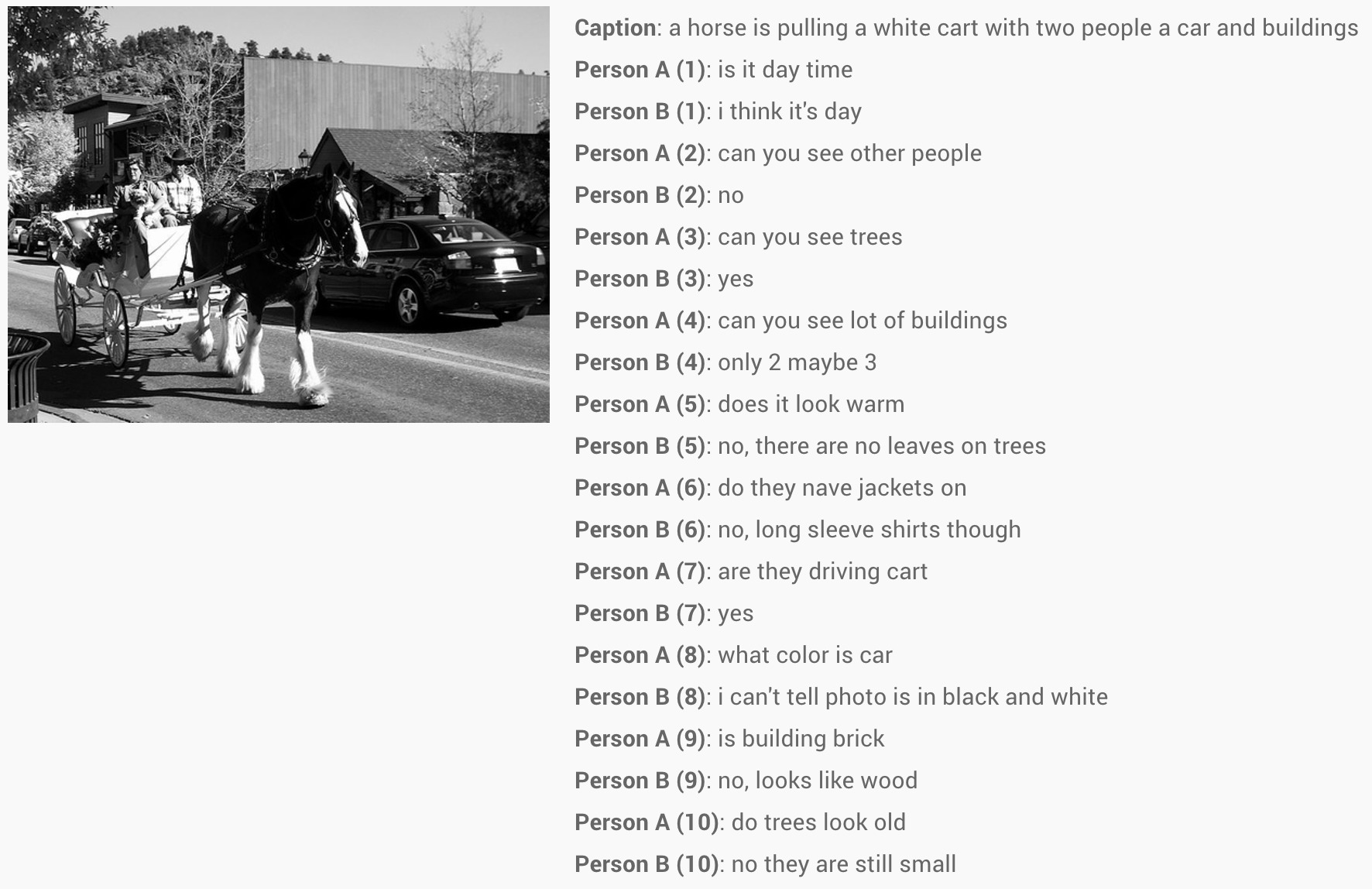}
	\caption{}
\end{subfigure}
\quad
\begin{subfigure}[t]{0.49\textwidth}
	\centering
	\includegraphics[trim={0cm 0cm 0cm 0cm},clip, width=\columnwidth]{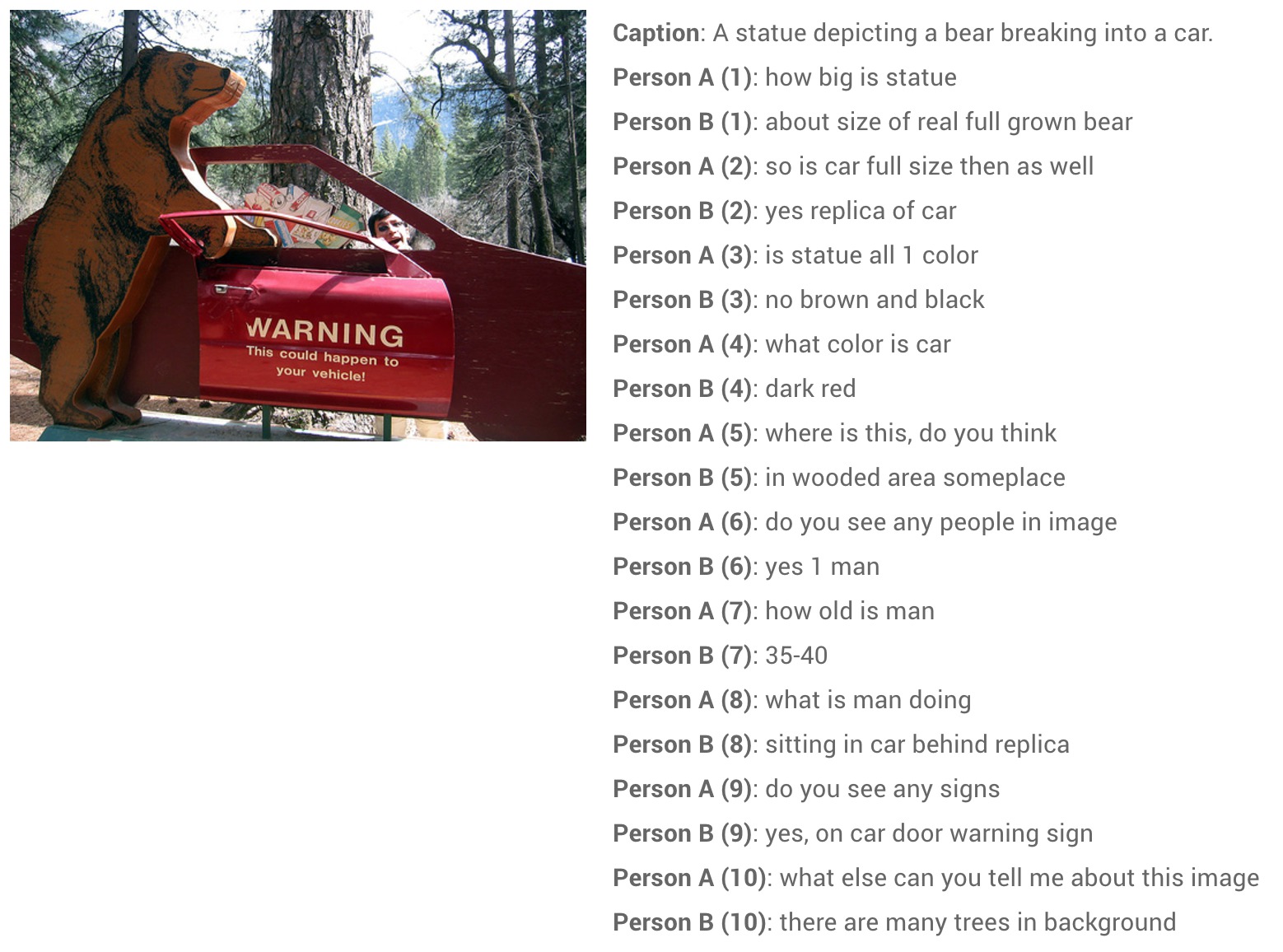}
	\caption{}
\end{subfigure}
\begin{subfigure}[t]{0.49\textwidth}
	\centering
	\includegraphics[trim={0cm 0cm 0cm 0cm},clip, width=\columnwidth]{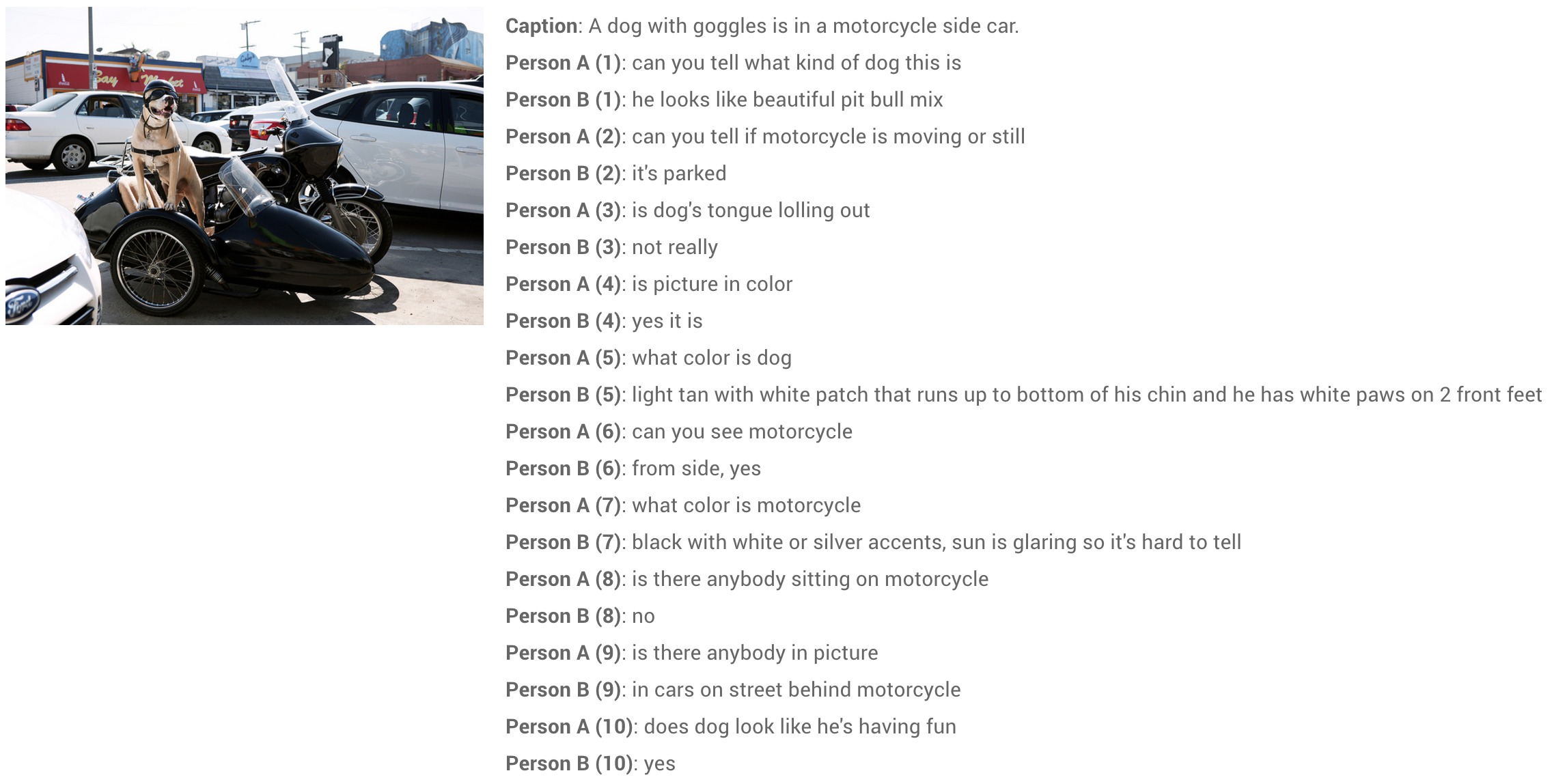}
	\caption{}
\end{subfigure}
\caption{Examples from \vd}
\label{fig:vd-examples-1}
\vspace{20pt}
\end{figure*}

\reffig{fig:vd-examples-1} shows random samples of \dialogs from the \vd dataset.

\section{Human-Machine Comparison}
\label{sec:human_supp}

\begin{table}[h]
{
    \small
    \setlength\tabcolsep{3.8pt}
    \centering
    \begin{tabular}{ccccccc}
    \toprule
    & \textbf{Model} & \textbf{MRR} & \textbf{R@1} & \textbf{R@5} & \textbf{Mean} \\
    \midrule
    \multirow{4}{*}{\rotatebox[origin=c]{90}{Human} $\begin{dcases} \\ \\ \\ \end{dcases}$} &
        Human-Q & 0.441 & 25.10 & 67.37 & 4.19 \\
    &Human-QH & 0.485 & 30.31 & 70.53 & 3.91 \\
    &Human-QI & 0.619 & 46.12 & 82.54 & 2.92 \\
    &Human-QIH & 0.635 & 48.03 & 83.76 & 2.83 \\
    \midrule
    \multirow{3}{*}{\rotatebox[origin=c]{90}{Machine} $\begin{dcases} \\ \\ \end{dcases}$} &
        HREA-QIH-G & 0.477 & 31.64 & 61.61 & 4.42 \\
        &MN-QIH-G & 0.481 & 32.16 & 61.94 & 4.47 \\
		&MN-QIH-D & 0.553 & 36.86 & 69.39 & 3.48 \\
    \bottomrule
    \end{tabular}
    \caption{Human-machine performance comparison on \vd v0.5, measured by mean reciprocal rank (MRR), recall@$k$ for $k=\{1, 5\}$ and mean rank. Note that higher is better for MRR and recall@k, while lower is better for mean rank.}
    \label{table:human_results}
}
\end{table}

We conducted studies on AMT to quantitatively evaluate human performance on this task for all combinations
of \{with image, without image\}$\times$\{with history, without history\} on 100 random images at each of the 10 rounds.
Specifically, in each setting, we show human subjects a jumbled list of 10 candidate answers for a question --
top-9 predicted responses from our `LF-QIH-D' model and the 1 ground truth answer --
and ask them to rank the responses.
Each task was done by 3 human subjects.

Results of this study are shown in the top-half of \reftab{table:human_results}.
We find that without access to the image, humans perform better when they have access to dialog history
-- compare the Human-QH row to Human-Q (R@1 of 30.31 \vs 25.10).
As perhaps expected, this gap narrows down when humans have access to the image --
compare Human-QIH to Human-QI (R@1 of 48.03 \vs 46.12).

Note that these numbers are not directly comparable to machine performance reported in the main paper because models are
tasked with ranking 100 responses, while humans are asked to rank 10 candidates.
This is because the task of ranking 100 candidate responses would be too cumbersome for humans.

To compute comparable human and machine performance, we evaluate our best discriminative (MN-QIH-D)
and generative (HREA-QIH-G, MN-QIH-G)\footnote{
We use both HREA-QIH-G, MN-QIH-G since they have similar accuracies.}
models on the same 10 options that were presented to humans.
Note that in this setting, both humans and machines have R$@10$ = $1.0$, since there are only 10 options.

\begin{figure*}[t]
	\centering
	\begin{subfigure}[b]{0.95\textwidth}
		\centering
		\includegraphics[width=\textwidth, trim={0cm 0.5cm 0cm 0.09cm}, clip]{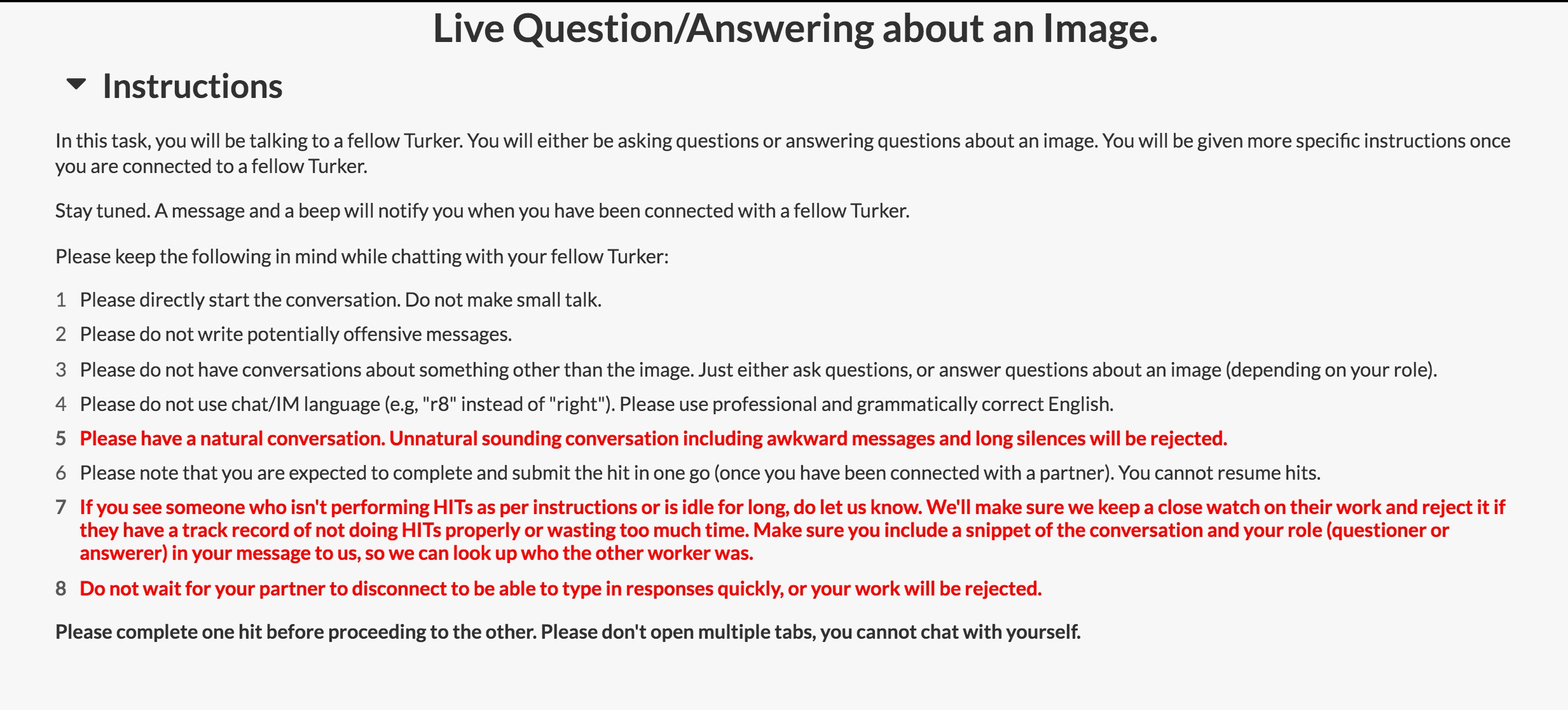}
		\caption{Detailed instructions for Amazon Mechanical Turkers on our interface}
	\label{fig:amt-instructions}
	\end{subfigure}

	\begin{subfigure}[b]{0.95\textwidth}
		\centering
		\includegraphics[width=\textwidth]{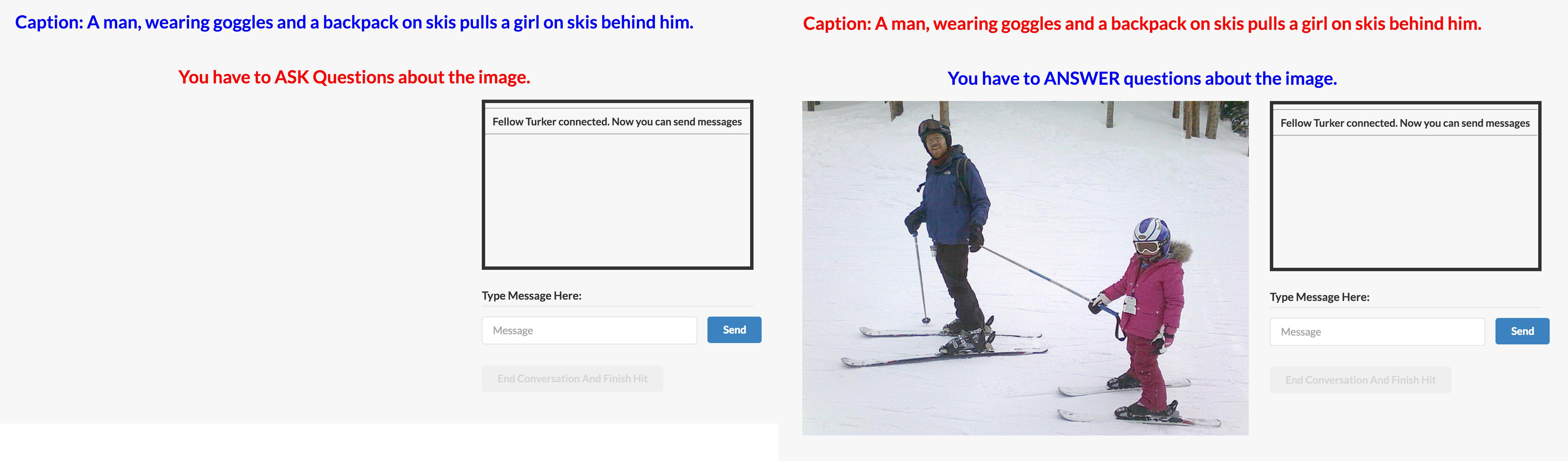}
		\caption{Left: What questioner sees; Right: What answerer sees.}
		\label{fig:amt-interface-q}
	\end{subfigure}
\end{figure*}

\reftab{table:human_results} bottom-half shows the results of this comparison.
We can see that, as expected, humans with full information (\ie Human-QIH) perform the best with a large gap
in human and machine performance (compare R@5: Human-QIH $83.76\%$ \vs MN-QIH-D $69.39\%$).
This gap is even larger when compared to generative models, which unlike the discriminative models are
not actively trying to exploit the biases in the answer candidates
(compare R@5: Human-QIH $83.76\%$ \vs HREA-QIH-G $61.61\%$).

Furthermore, we see that humans outperform the best machine \emph{even when not looking at the image},
simply on the basis of the context provided by the history (compare
R@5: Human-QH $70.53\%$ \vs MN-QIH-D $69.39\%$).

Perhaps as expected, with access to the image but not the history, humans are significantly better than the best machines
(R@5: Human-QI $82.54\%$ \vs MN-QIH-D $69.39\%$). With access to history humans perform even better.

From in-house human studies and worker feedback on AMT, we find that \dialog history plays the following roles for humans:
(1) provides a context for the question and paints a picture of the scene, which
helps eliminate certain answer choices (especially when the image is not available),
(2) gives cues about the answerer's response style, which helps identify the right answer among similar answer choices, and
(3) disambiguates amongst likely interpretations of the image (\ie, when objects are small or occluded),
again, helping identify the right answer among multiple plausible options.
\section{Interface}
\label{sec:interface}

In this section, we show our interface to connect two Amazon Mechanical Turk workers live, which we used to collect our data.

\textbf{Instructions.}
To ensure quality of data, we provide detailed instructions on our interface as shown in \reffig{fig:amt-instructions}.
Since the workers do not know their roles before starting the study, we provide instructions for both questioner and answerer roles.

\textbf{After pairing:}
Immediately after pairing two workers, we assign them roles of a questioner and a answerer and display role-specific instructions as shown in
\reffig{fig:amt-interface-q}. 
Observe that the questioner does not see the image while the answerer does have access to it.
Both questioner and answerer see the caption for the image.


\section{Additional Analysis of \vd}
\label{sec:analysis}

In this section, we present additional analyses characterizing our \vd dataset.

\subsection{Question and Answer Lengths}

\begin{figure}[h]
    \centering
    \includegraphics[width=\linewidth]{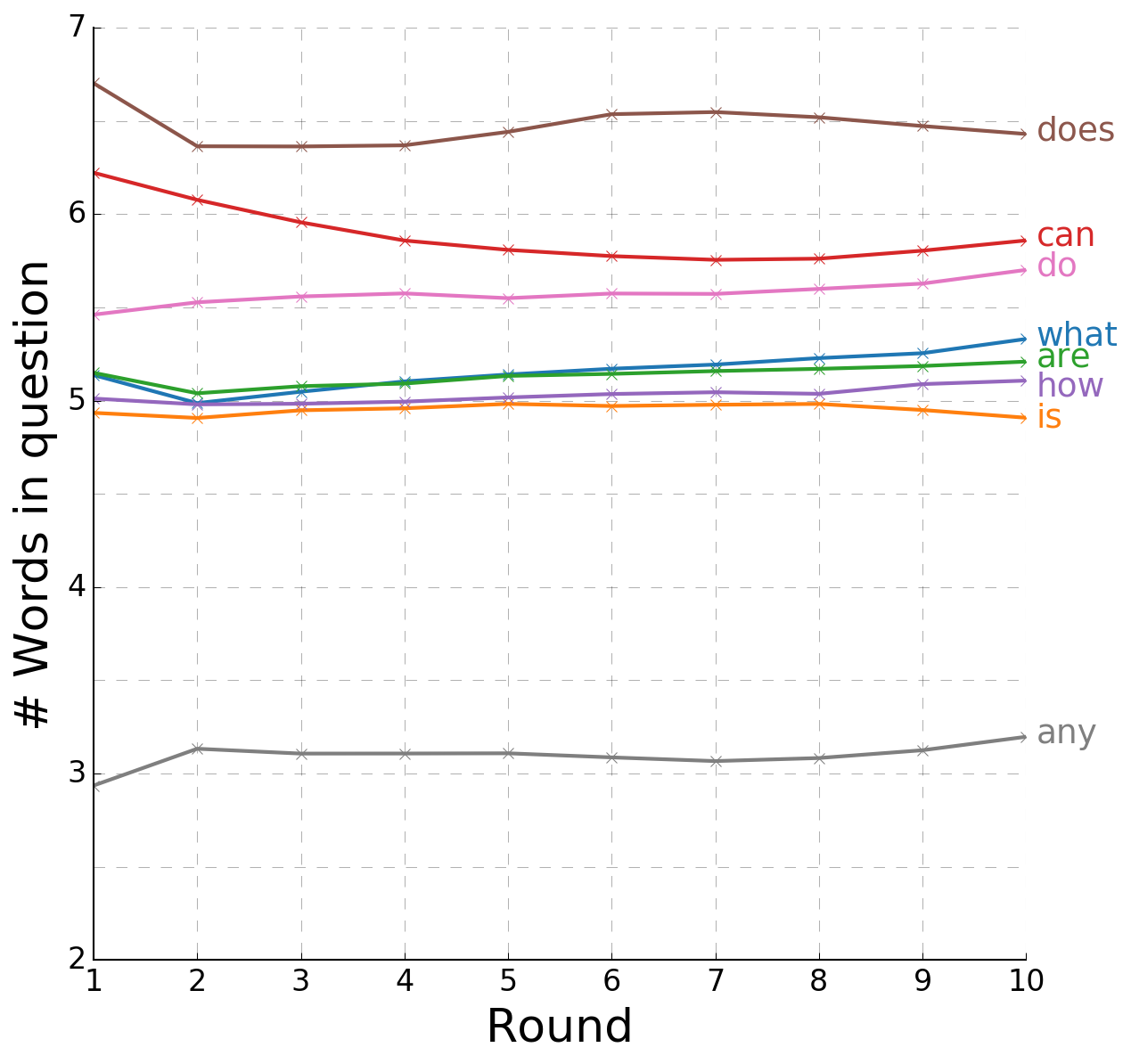}
    \caption{Question lengths by type and round. Average length of question by type is fairly consistent across rounds. Questions starting with `any' (`any people?', `any other fruits?', etc.) tend to be the shortest.}
    \label{fig:qlength_type_round}
\end{figure}

\begin{figure}[h]
    \centering
    \includegraphics[width=\linewidth]{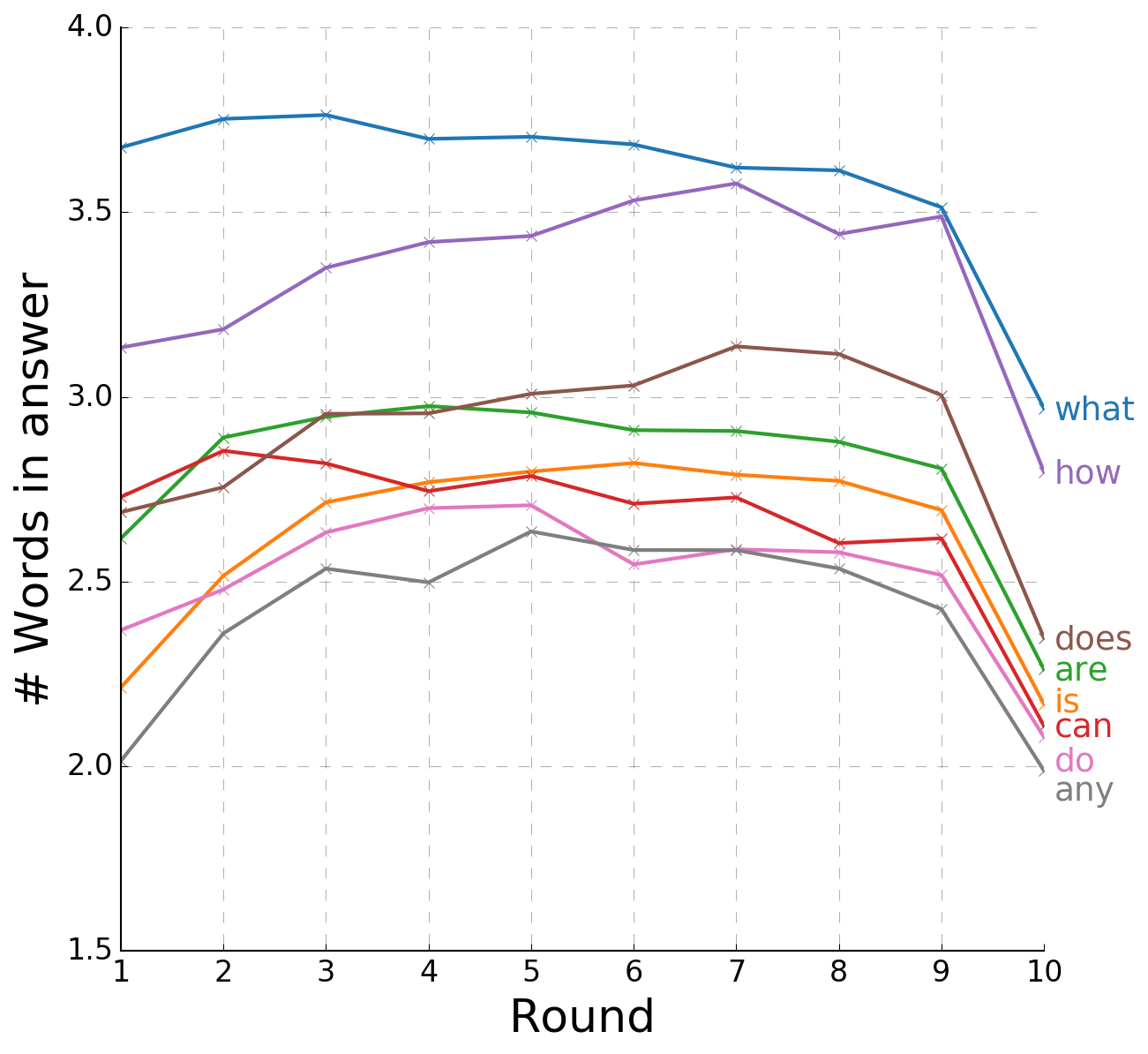}
    \caption{Answer lengths by question type and round. Across question types, average response length tends to be longest in the middle of the conversation.}
    \label{fig:alength_type_round}
\end{figure}

    \reffig{fig:qlength_type_round} shows question lengths by type and round.
    Average length of question by type is consistent across rounds.
    Questions starting with `any' (\myquote{any people?}, \myquote{any other fruits?}, \etc) tend to be the shortest.
    \reffig{fig:alength_type_round} shows answer lengths by type of question they were said in response to and round.
    In contrast to questions, there is significant variance in answer lengths.
    Answers to binary questions (\myquote{Any people?}, \myquote{Can you see the dog?}, \etc) tend to be
    short while answers to `how' and `what' questions tend to be more explanatory and long.
    Across question types, answers tend to be the longest in the middle of conversations.

\subsection{Question Types}

\begin{figure}[h]
    \centering
    \includegraphics[width=\linewidth]{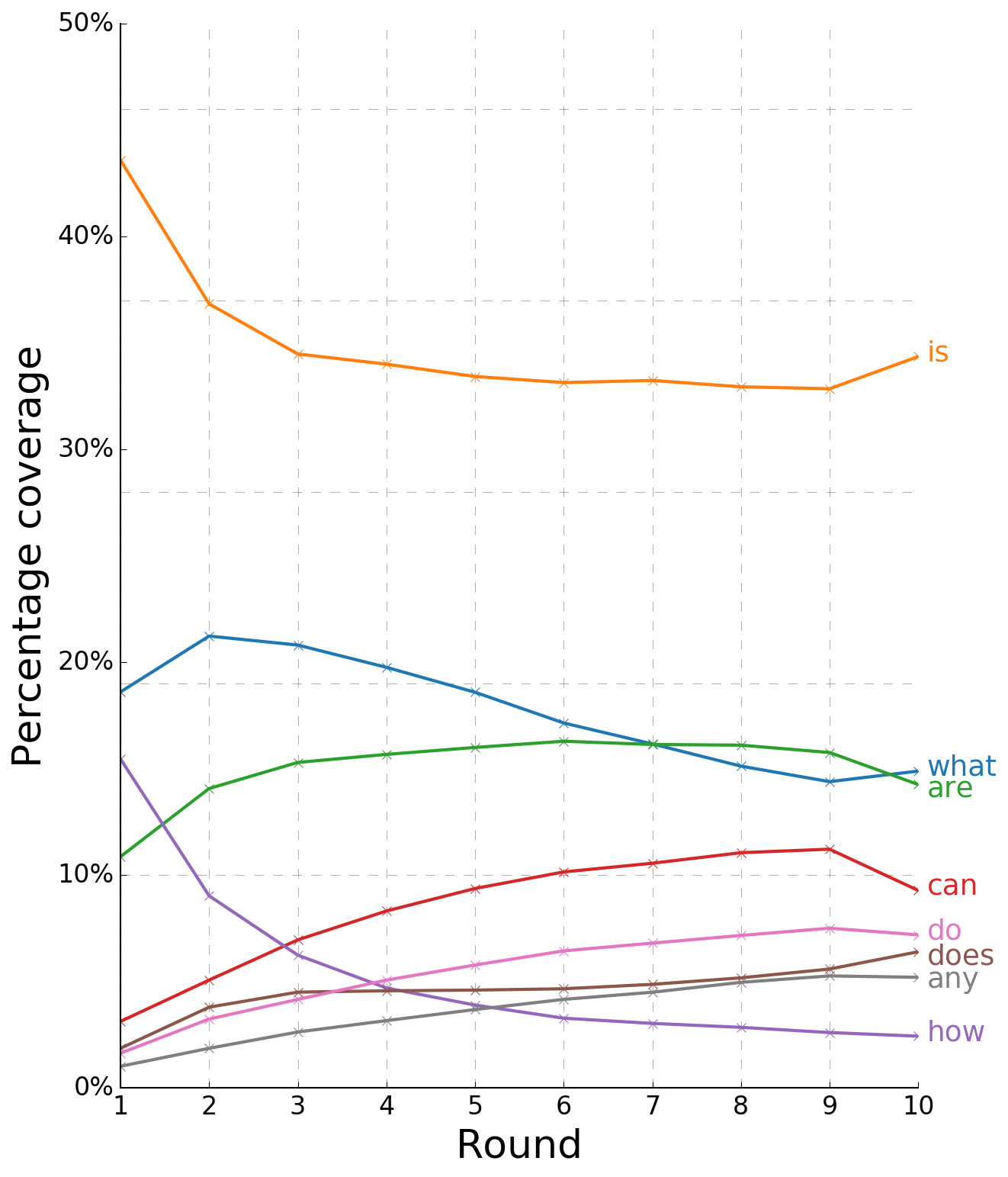}
    \caption{Percentage coverage of question types per round. As conversations progress, `Is', `What' and `How' questions reduce while `Can', `Do', `Does', `Any' questions occur more often. Questions starting with `Is' are the most popular in the dataset.}
    \label{fig:qtype_round}
\end{figure}

    \reffig{fig:qtype_round} shows round-wise coverage by question type.
    We see that as conversations progress, `is', `what' and `how' questions reduce while `can', `do', `does', `any' questions occur more often.
    Questions starting with `Is' are the most popular in the dataset.

\begin{figure*}[ht]
        \centering
        \includegraphics[width=\linewidth]{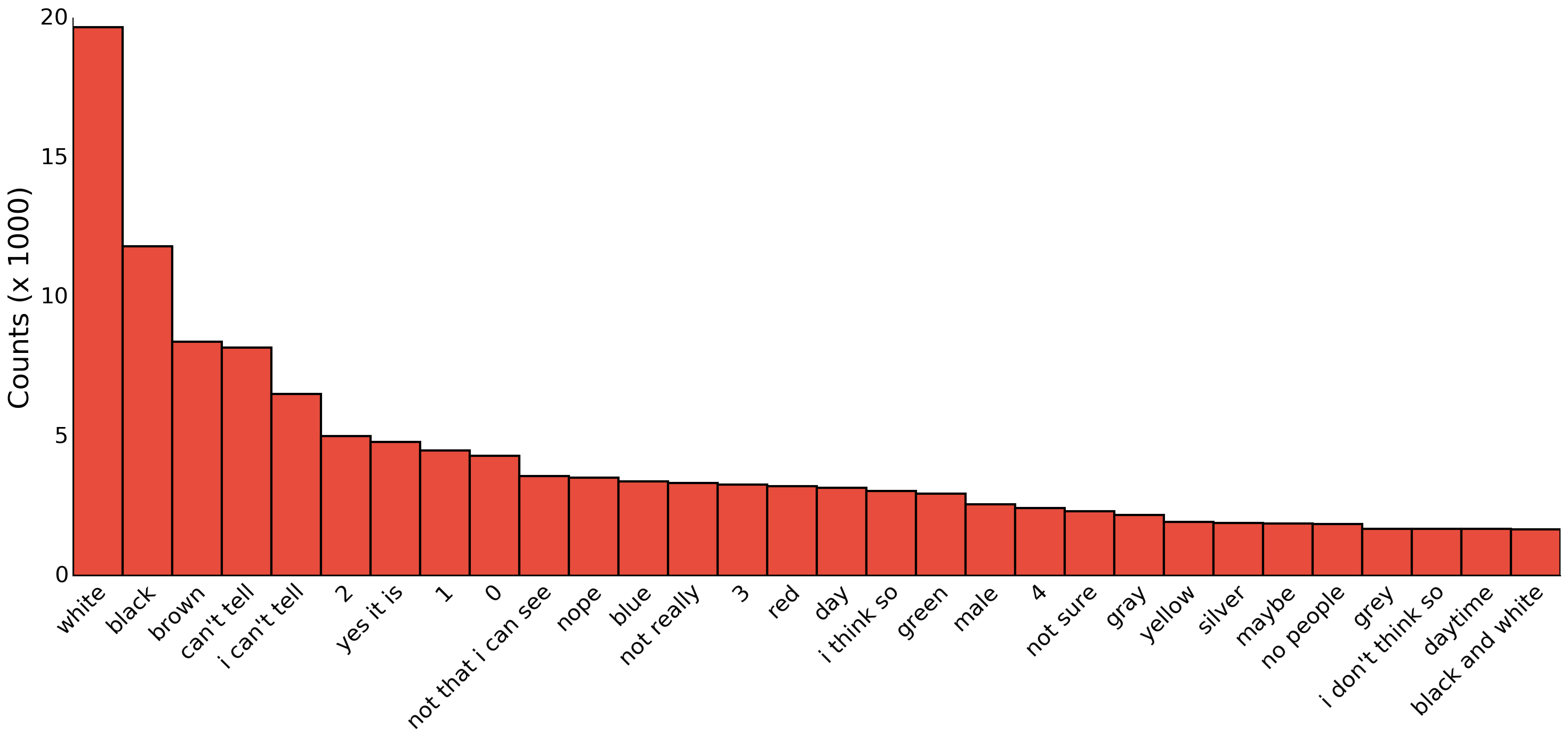}
        \caption{Most frequent answer responses except for `yes'/`no'}
        \label{fig:top_answers}
\end{figure*}


\section{Performance on \vd v0.5}
\label{sec:vdot5-results}

\begin{table}[t]
{
    \small
    \setlength\tabcolsep{3.8pt}
    \centering
    \begin{tabular}{ccccccc}
    \toprule
    & \textbf{Model} & \textbf{MRR} & \textbf{R@1} & \textbf{R@5} & \textbf{R@10} & \textbf{Mean} \\
    \midrule
    \multirow{3}{*}{\rotatebox[origin=c]{90}{Baseline} $\begin{dcases} \\ \\ \\ \end{dcases}$}
            &Answer prior & 0.311 & 19.85 & 39.14 & 44.28 & 31.56 \\[1.5pt]
    &NN-Q & 0.392 & 30.54 & 46.99 & 49.98 & 30.88 \\[1.5pt]   
    &NN-QI & 0.385 & 29.71 & 46.57 & 49.86 & 30.90 \\[1.5pt]
    \midrule
    \multirow{9}{*}{\rotatebox[origin=c]{90}{Generative} $\begin{dcases}\\  \\ \\ \\  \\ \\ \\ \end{dcases}$}
        & \lf-Q-G & 0.403 & 29.74 & 50.10 & 56.32 & 24.06 \\
    &\lf-QH-G & 0.425 & 32.49 & 51.56 & 57.80 & 23.11 \\
    &\lf-QI-G & 0.437 & 34.06 & 52.50 & 58.89 & 22.31 \\
    &\lf-QIH-G & 0.430 & 33.27 & 51.96 & 58.09 & 23.04 \\
    \cdashline{2-7}
    & \hre-QH-G & 0.430 & 32.84 & 52.36 & 58.64 & 22.59 \\
    &\hre-QIH-G & 0.442 & 34.37 & 53.40 & 59.74 & 21.75 \\
    &\hre{}A-QIH-G & 0.442 & 34.47 & 53.43 & 59.73 & 21.83 \\
    \cdashline{2-7}
    &\mn-QH-G & 0.434 & 33.12 & 53.14 & 59.61 & 22.14 \\
    &\mn-QIH-G & \textbf{0.443} & \textbf{34.62} & \textbf{53.74} & \textbf{60.18} & \textbf{21.69} \\
    \midrule
    \multirow{9}{*}{\rotatebox[origin=c]{90}{Discriminative} $\begin{dcases} \\ \\ \\  \\ \\ \\  \\ \end{dcases}$}
            &\lf-Q-D & 0.482 & 34.29 & 63.42 & 74.31 & 8.87 \\
    &\lf-QH-D & 0.505 & 36.21 & 66.56 & 77.31 & 7.89\\
    &\lf-QI-D & 0.502 & 35.76 & 66.59 & 77.61 & 7.72  \\
    &\lf-QIH-D & 0.511 & 36.72 & 67.46 & 78.30 & 7.63\\
    \cdashline{2-7}
    &\hre-QH-D & 0.489 & 34.74 & 64.25 & 75.40 & 8.32\\
    &\hre-QIH-D & 0.502 & 36.26 & 65.67 & 77.05 & 7.79 \\
    &\hre{}A-QIH-D & 0.508 & 36.76 & 66.54 & 77.75 & 7.59 \\
    \cdashline{2-7}
    &\mn-QH-D & 0.524 & 36.84 & 67.78 & 78.92 & 7.25 \\
    &\mn-QIH-D & \textbf{0.529} & \textbf{37.33} & \textbf{68.47} & \textbf{79.54} & \textbf{7.03} \\
    \midrule
    \multirow{2}{*}{\rotatebox[origin=c]{90}{VQA} $\begin{dcases} \\ \end{dcases}$} &
    SAN1-QI-D & 0.506 & 36.21 & 67.08 & 78.16 & 7.74 \\
    &HieCoAtt-QI-D & 0.509 & 35.54 & 66.79 & 77.94 & 7.68 \\
    \midrule
    \multicolumn{7}{c}{Human Accuracies}\\
    \midrule
    \multirow{4}{*}{\rotatebox[origin=c]{90}{Human} $\begin{dcases} \\ \\ \\ \end{dcases}$} &
        Human-Q & 0.441 & 25.10 & 67.37 & - & 4.19 \\
    &Human-QH & 0.485 & 30.31 & 70.53 & - & 3.91 \\
    &Human-QI & 0.619 & 46.12 & 82.54 & - & 2.92 \\
    &Human-QIH & 0.635 & 48.03 & 83.76 & - & 2.83 \\
    \bottomrule
    \end{tabular}
    \caption{Performance of methods on \vd v0.5, measured by mean reciprocal rank (MRR), recall@$k$ for $k=\{1, 5, 10\}$ and mean rank. Note that higher is better for MRR and recall@k, while lower is better for mean rank.
    \mnfull has the best performance in both discriminative and generative settings.}
    \label{table:model_results}
}
\end{table}

\reftab{table:model_results} shows the results for our proposed models and baselines on \vd v0.5.
A few key takeaways --
First, as expected, all learning based models significantly outperform non-learning baselines.
Second, all discriminative models significantly outperform generative models,
which as we discussed is expected since discriminative models can tune to the biases in the answer options.
This improvement comes with the significant limitation of not being able to actually generate responses,
and we recommend the two decoders be viewed as separate use cases.
Third, our best generative and discriminative models are
\mn-QIH-G with 0.44 MRR,
and \mn-QIH-D with 0.53 MRR
that outperform a suite of models and sophisticated baselines.
Fourth, we observe that models with $H$ perform better than $Q$-only models, highlighting the importance of history in \vd.
Fifth, models looking at $I$ outperform both the blind models ($Q$, $QH$) by at least $2\%$ on recall@$1$
in both decoders.
Finally, models that use both $H$ and $I$ have best performance.

\begin{figure}[h]
    \begin{subfigure}[b]{0.45\columnwidth}
        \centering
        \includegraphics[width=\textwidth]{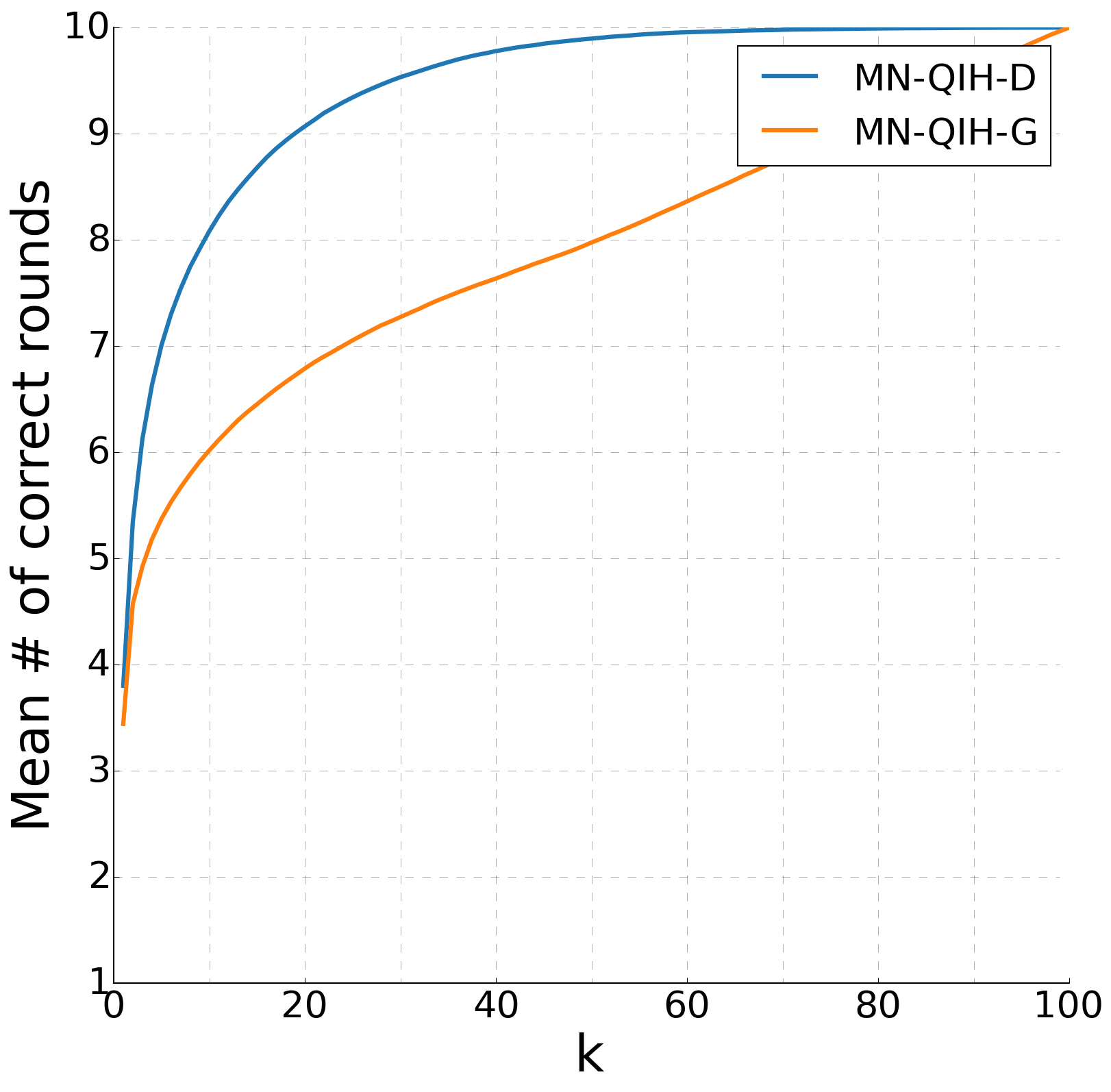}
        \caption{}
        \label{fig:cr_k}
    \end{subfigure}
    \begin{subfigure}[b]{0.45\columnwidth}
        \centering
        \includegraphics[width=\textwidth]{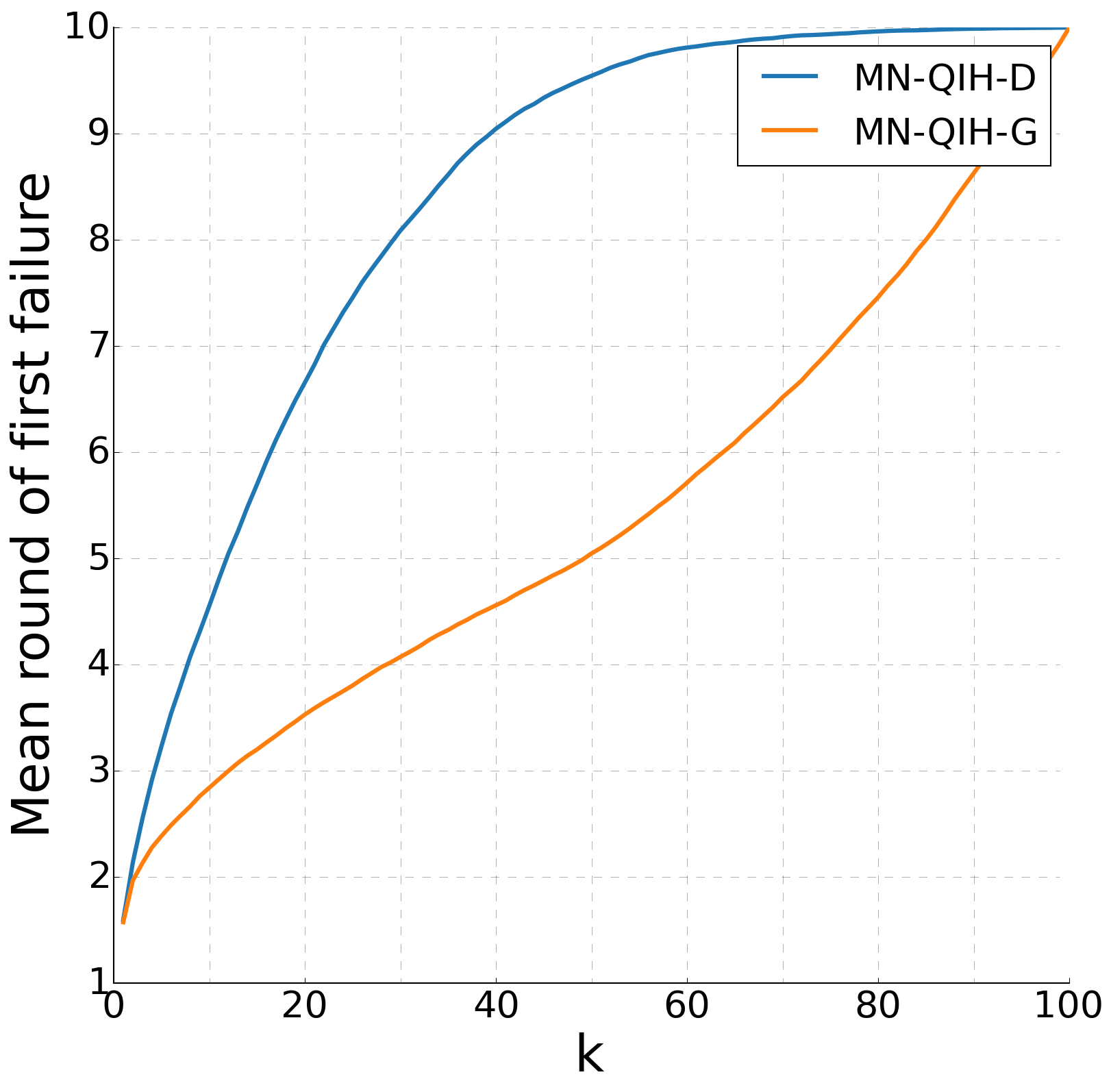}
        \caption{}
        \label{fig:fpof_k}
    \end{subfigure}
    \caption{Dialog-level evaluation}
    \label{fig:failure}
\end{figure}

\pagebreak
\textbf{Dialog-level evaluation.}
Using R$@5$ to define round-level `success',
our best discriminative model MN-QIH-D gets $7.01$ rounds out of 10 correct, while generative MN-QIH-G gets $5.37$.
Further, the mean first-failure-round (under $R@5$)
for MN-QIH-D is $3.23$, and $2.39$ for MN-QIH-G.
\reffig{fig:cr_k} and \reffig{fig:fpof_k} show plots for all values of $k$ in $R@k$.

\section{Experimental Details}
\label{sec:exp-details}

In this section, we describe details about our models, data preprocessing, training procedure and hyperparameter selection.

\subsection{Models}
\begin{figure*}[ht]
    \centering
    \begin{subfigure}[t]{0.92\textwidth}
    \includegraphics[width=\linewidth]{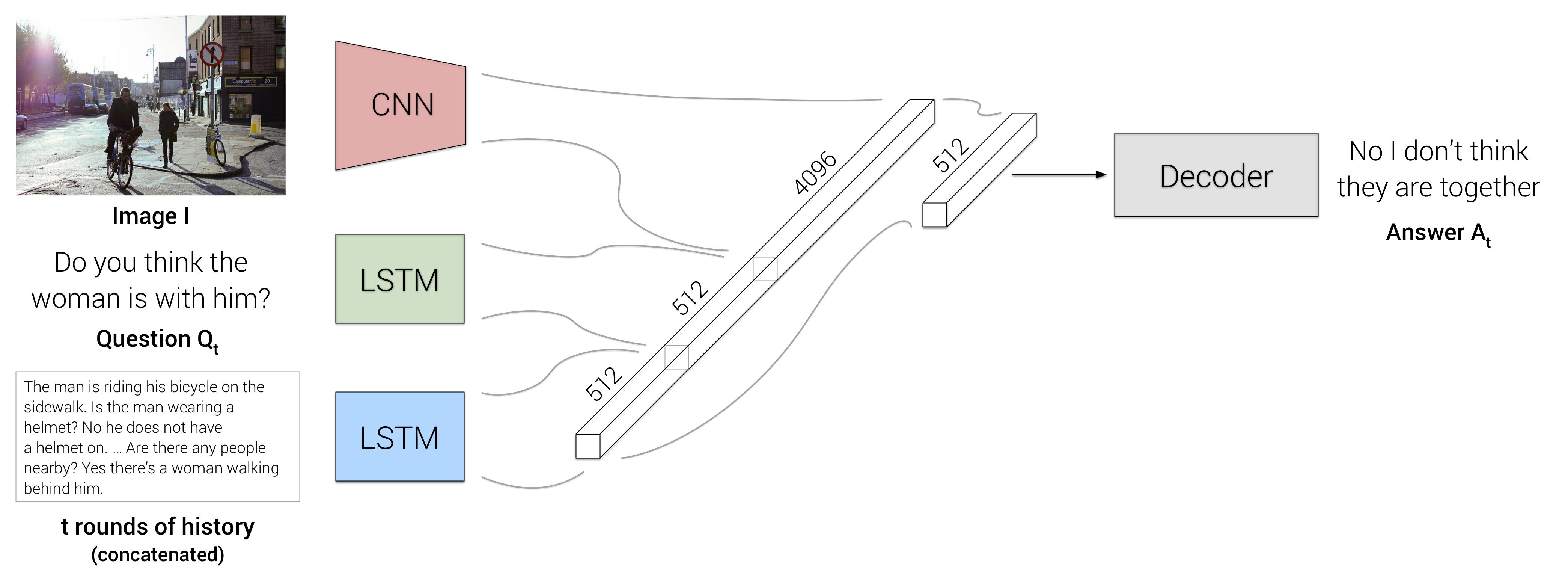}
    \caption{Late Fusion Encoder}
    \label{fig:lf_enc}
    \end{subfigure}
    \begin{subfigure}[t]{0.92\textwidth}
    \includegraphics[width=\linewidth]{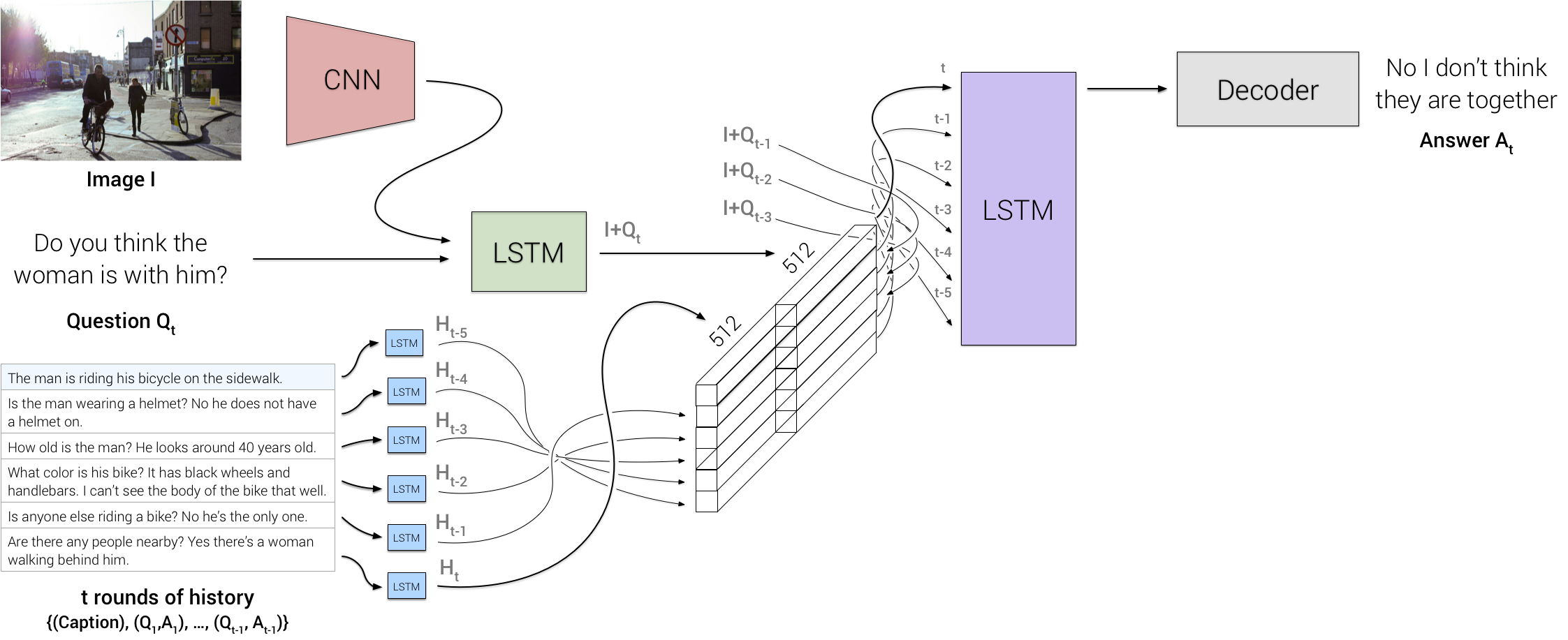}
    \caption{Hierarchical Recurrent Encoder}
    \label{fig:hre_enc}
	\end{subfigure}
	\begin{subfigure}[t]{0.92\textwidth}
    \includegraphics[width=\linewidth]{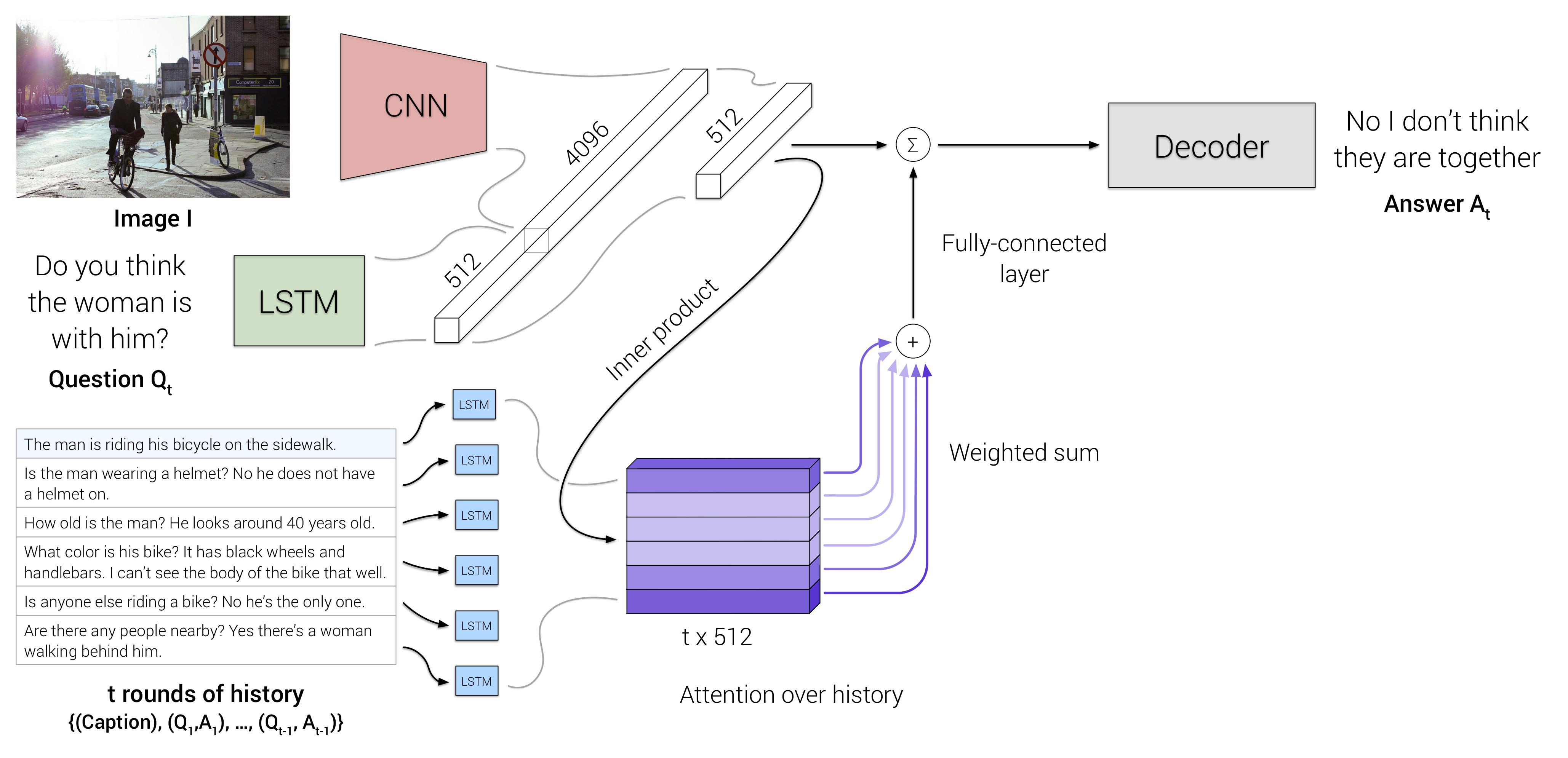}
    \caption{Memory Network Encoder}
    \label{fig:mn_enc}
    \end{subfigure}
    \caption{}
\end{figure*}

\paragraph{\lffull (\lf) Encoder.}
    We encode the image with a VGG-16 CNN, question and concatenated history with separate LSTMs and concatenate the three representations.
    This is followed by a fully-connected layer and tanh non-linearity to a $512$-d vector, which is used to decode the response.
    \reffig{fig:lf_enc} shows the model architecture for our \lf encoder.

\paragraph{\hrefull (\hre).}
    In this encoder, the image representation from VGG-16 CNN is early fused with the question. Specifically, the image representation is concatenated with every question word as it is fed to an LSTM.
    Each QA-pair in \dialog history is independently encoded by another LSTM with shared weights.
    The image-question representation, computed for every round from $1$ through $t$, is concatenated with history representation from the previous round and constitutes a sequence of question-history vectors.
    These vectors are fed as input to a \dialog-level LSTM, whose output state at $t$ is used to decode the response to $Q_t$.
    \reffig{fig:hre_enc} shows the model architecture for our \hre.

\begin{figure*}[ht]
    \centering
    \includegraphics[width=0.90\textwidth]{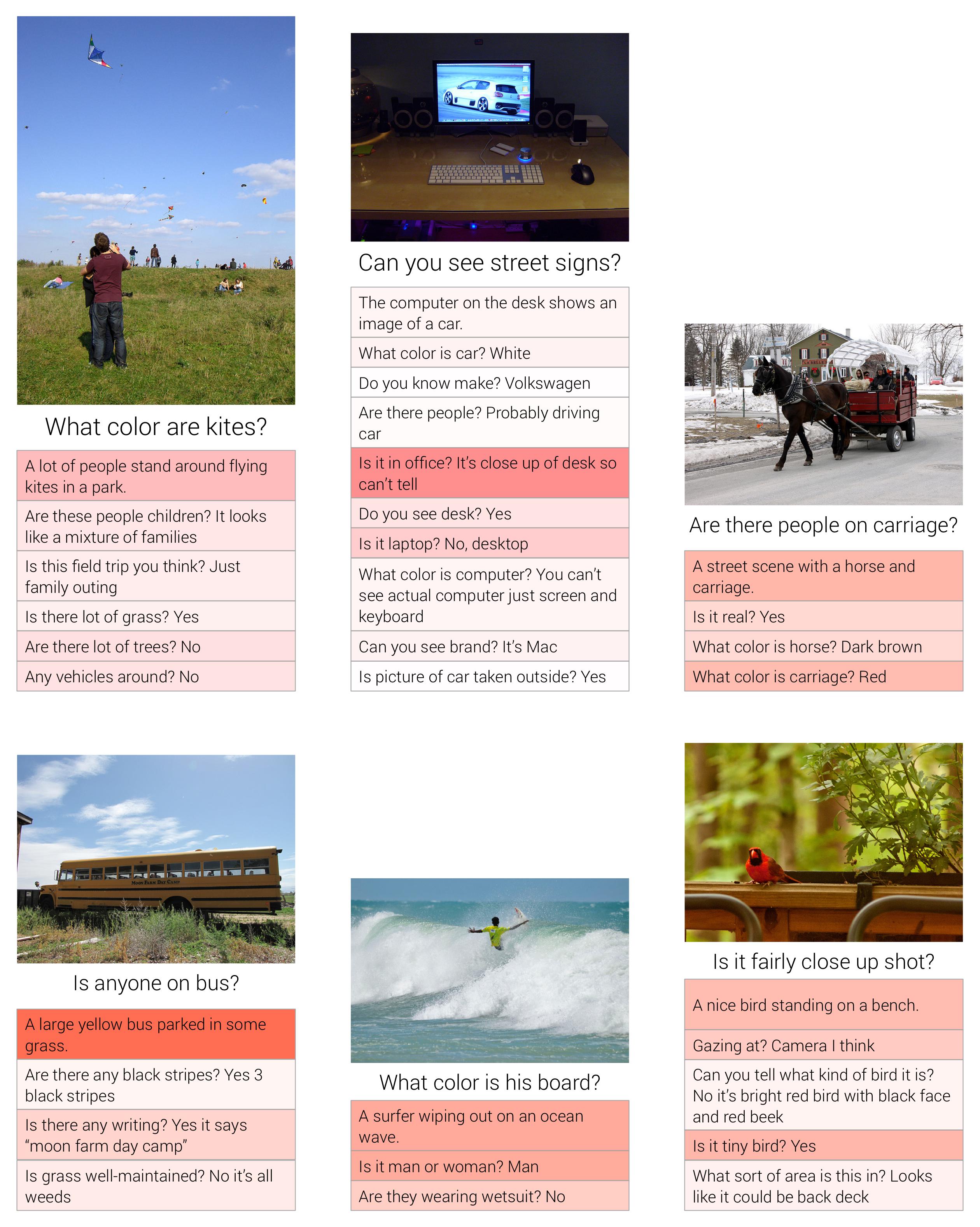} \\
    \caption{Selected examples of attention over history facts from our \mnfull encoder. The intensity of color in each row indicates the strength of attention placed on that round by the model.}
    \label{fig:mn_att}
\end{figure*}

\paragraph{\mnfull.}
    The image is encoded with a VGG-16 CNN and question with an LSTM.
    We concatenate the representations and follow it by a fully-connected layer and tanh non-linearity to get a `query vector'.
    Each caption/QA-pair (or `fact') in \dialog history is encoded independently by an LSTM with shared weights.
    The query vector is then used to compute attention over the $t$ facts by inner product.
    Convex combination of attended history vectors is passed through a fully-connected layer and tanh non-linearity, and added back to the query vector.
    This combined representation is then passed through another fully-connected layer and tanh non-linearity and then used to decode the response.
    The model architecture is shown in \reffig{fig:mn_enc}.
    \reffig{fig:mn_att} shows some examples of attention over history facts from our \mn encoder.
    We see that the model learns to attend to facts relevant to the question being asked.
    For example, when asked `What color are kites?', the model attends to `A lot of people stand around flying kites in a park.'
    For `Is anyone on bus?', it attends to `A large yellow bus parked in some grass.'
    Note that these are selected examples, and not always are these attention weights interpretable.
\subsection{Training}
\paragraph{Splits.}
Recall that \vd v0.9 contained 83k \dialogs on COCO-\train and 40k on COCO-\val images.
We split the 83k into 80k for training, 3k for validation, and use the 40k as test.

 \paragraph{Preprocessing.}
 We spell-correct \vd data using the Bing API~\cite{bing_spellapi}.
 Following VQA, we lowercase all questions and answers,
 convert digits to words, and remove contractions, before tokenizing using the Python NLTK~\cite{nltk}.
 We then construct a dictionary of words that appear at least five times in the train set, giving us a vocabulary of around $7.5$k.

 \paragraph{Hyperparameters.}
 All our models are implemented in Torch~\cite{torch}. 
 Model hyperparameters are chosen by early stopping on \val based on the Mean Reciprocal Rank (MRR) metric.
 All LSTMs are 2-layered with $512$-dim hidden states.
 We learn $300$-dim embeddings for words and images.
 These word embeddings are shared across question, history, and decoder LSTMs.
 We use Adam~\cite{kingma_iclr15} with a learning rate of $10^{-3}$ for all models.
 Gradients at each iterations are clamped to $[-5, 5]$ to avoid explosion.
 Our code, architectures, and trained models are available at {\small\url{https://visualdialog.org}}.

{
\small
\bibliographystyle{ieee}
\bibliography{strings,main}

\begin{thebibliography}{10}\itemsep=-1pt

\bibitem{nltk}
{NLTK}.
\newblock \url{http://www.nltk.org/}.

\bibitem{torch}
{Torch}.
\newblock \url{http://torch.ch/}.

\bibitem{aagrawal_emnlp16}
A.~Agrawal, D.~Batra, and D.~Parikh.
\newblock {Analyzing the Behavior of Visual Question Answering Models}.
\newblock In {\em EMNLP}, 2016.

\bibitem{agrawal_emnlp16}
H.~Agrawal, A.~Chandrasekaran, D.~Batra, D.~Parikh, and M.~Bansal.
\newblock Sort story: Sorting jumbled images and captions into stories.
\newblock In {\em EMNLP}, 2016.

\bibitem{alexa}
Amazon.
\newblock {Alexa}.
\newblock \url{http://alexa.amazon.com/}.

\bibitem{antol_iccv15}
S.~Antol, A.~Agrawal, J.~Lu, M.~Mitchell, D.~Batra, C.~L. Zitnick, and
  D.~Parikh.
\newblock {VQA: Visual Question Answering}.
\newblock In {\em ICCV}, 2015.

\bibitem{bigham_uist10}
J.~P. Bigham, C.~Jayant, H.~Ji, G.~Little, A.~Miller, R.~C. Miller, R.~Miller,
  A.~Tatarowicz, B.~White, S.~White, and T.~Yeh.
\newblock {VizWiz: Nearly Real-time Answers to Visual Questions}.
\newblock In {\em UIST}, 2010.

\bibitem{bordes_arxiv15}
A.~Bordes, N.~Usunier, S.~Chopra, and J.~Weston.
\newblock {Large-scale Simple Question Answering with Memory Networks}.
\newblock {\em arXiv preprint arXiv:1506.02075}, 2015.

\bibitem{bordes_arxiv16}
A.~Bordes and J.~Weston.
\newblock {Learning End-to-End Goal-Oriented Dialog}.
\newblock {\em arXiv preprint arXiv:1605.07683}, 2016.

\bibitem{christie_emnlp16}
G.~Christie, A.~Laddha, A.~Agrawal, S.~Antol, Y.~Goyal, K.~Kochersberger, and
  D.~Batra.
\newblock Resolving language and vision ambiguities together: Joint
  segmentation and prepositional attachment resolution in captioned scenes.
\newblock In {\em EMNLP}, 2016.

\bibitem{mizil_cmcl11}
C.~Danescu-Niculescu-Mizil and L.~Lee.
\newblock {Chameleons in imagined conversations: A new approach to
  understanding coordination of linguistic style in dialogs.}
\newblock In {\em Proceedings of the Workshop on Cognitive Modeling and
  Computational Linguistics, ACL 2011}, 2011.

\bibitem{das_emnlp16}
A.~Das, H.~Agrawal, C.~L. Zitnick, D.~Parikh, and D.~Batra.
\newblock {Human Attention in Visual Question Answering: Do Humans and Deep
  Networks Look at the Same Regions?}
\newblock In {\em EMNLP}, 2016.

\bibitem{vries_cvpr17}
H.~de~Vries, F.~Strub, S.~Chandar, O.~Pietquin, H.~Larochelle, and A.~C.
  Courville.
\newblock {GuessWhat?! Visual object discovery through multi-modal dialogue}.
\newblock In {\em CVPR}, 2017.

\bibitem{dodge_iclr16}
J.~Dodge, A.~Gane, X.~Zhang, A.~Bordes, S.~Chopra, A.~Miller, A.~Szlam, and
  J.~Weston.
\newblock {Evaluating Prerequisite Qualities for Learning End-to-End Dialog
  Systems}.
\newblock In {\em ICLR}, 2016.

\bibitem{donahue_cvpr15}
J.~Donahue, L.~A. Hendricks, S.~Guadarrama, M.~Rohrbach, S.~Venugopalan,
  K.~Saenko, and T.~Darrell.
\newblock {Long-term Recurrent Convolutional Networks for Visual Recognition
  and Description}.
\newblock In {\em CVPR}, 2015.

\bibitem{fang_cvpr15}
H.~Fang, S.~Gupta, F.~N. Iandola, R.~K. Srivastava, L.~Deng, P.~Doll{\'{a}}r,
  J.~Gao, X.~He, M.~Mitchell, J.~C. Platt, C.~L. Zitnick, and G.~Zweig.
\newblock {From Captions to Visual Concepts and Back}.
\newblock In {\em CVPR}, 2015.

\bibitem{gao_nips15}
H.~Gao, J.~Mao, J.~Zhou, Z.~Huang, L.~Wang, and W.~Xu.
\newblock {Are You Talking to a Machine? Dataset and Methods for Multilingual
  Image Question Answering}.
\newblock In {\em NIPS}, 2015.

\bibitem{geman_pnas14}
D.~Geman, S.~Geman, N.~Hallonquist, and L.~Younes.
\newblock {A Visual Turing Test for Computer Vision Systems}.
\newblock In {\em PNAS}, 2014.

\bibitem{goyal_cvpr17}
Y.~Goyal, T.~Khot, D.~Summers-Stay, D.~Batra, and D.~Parikh.
\newblock Making the v in vqa matter: Elevating the role of image understanding
  in visual question answering.
\newblock In {\em CVPR}, 2017.

\bibitem{he_cvpr16}
K.~He, X.~Zhang, S.~Ren, and J.~Sun.
\newblock {Deep Residual Learning for Image Recognition}.
\newblock In {\em CVPR}, 2016.

\bibitem{hermann_nips15}
K.~M. Hermann, T.~Kocisky, E.~Grefenstette, L.~Espeholt, W.~Kay, M.~Suleyman,
  and P.~Blunsom.
\newblock Teaching machines to read and comprehend.
\newblock In {\em NIPS}, 2015.

\bibitem{hu_eccv16}
R.~Hu, M.~Rohrbach, and T.~Darrell.
\newblock Segmentation from natural language expressions.
\newblock In {\em ECCV}, 2016.

\bibitem{huang_naacl16}
T.-H. Huang, F.~Ferraro, N.~Mostafazadeh, I.~Misra, A.~Agrawal, J.~Devlin,
  R.~Girshick, X.~He, P.~Kohli, D.~Batra, L.~Zitnick, D.~Parikh,
  L.~Vanderwende, M.~Galley, and M.~Mitchell.
\newblock Visual storytelling.
\newblock In {\em NAACL HLT}, 2016.

\bibitem{sutskever_nips14}
Q.~V.~L. Ilya~Sutskever, Oriol~Vinyals.
\newblock {Sequence to Sequence Learning with Neural Networks}.
\newblock In {\em NIPS}, 2014.

\bibitem{jabri_eccv16}
A.~Jabri, A.~Joulin, and L.~van~der Maaten.
\newblock Revisiting visual question answering baselines.
\newblock In {\em ECCV}, 2016.

\bibitem{kannan_kdd16}
A.~Kannan, K.~Kurach, S.~Ravi, T.~Kaufmann, A.~Tomkins, B.~Miklos, G.~Corrado,
  L.~Luk{\'a}cs, M.~Ganea, P.~Young, et~al.
\newblock {Smart Reply: Automated Response Suggestion for Email}.
\newblock In {\em KDD}, 2016.

\bibitem{karpathy_cvpr15}
A.~Karpathy and L.~Fei-Fei.
\newblock {Deep visual-semantic alignments for generating image descriptions}.
\newblock In {\em CVPR}, 2015.

\bibitem{kingma_iclr15}
D.~Kingma and J.~Ba.
\newblock {Adam: A Method for Stochastic Optimization}.
\newblock In {\em ICLR}, 2015.

\bibitem{kong_cvpr14}
C.~Kong, D.~Lin, M.~Bansal, R.~Urtasun, and S.~Fidler.
\newblock {What are you talking about? text-to-image coreference}.
\newblock In {\em CVPR}, 2014.

\bibitem{lemon_eacl06}
O.~Lemon, K.~Georgila, J.~Henderson, and M.~Stuttle.
\newblock {An ISU dialogue system exhibiting reinforcement learning of dialogue
  policies: generic slot-filling in the TALK in-car system}.
\newblock In {\em EACL}, 2006.

\bibitem{li_emnlp16}
J.~Li, W.~Monroe, A.~Ritter, M.~Galley, J.~Gao, and D.~Jurafsky.
\newblock {Deep Reinforcement Learning for Dialogue Generation}.
\newblock In {\em EMNLP}, 2016.

\bibitem{mscoco}
T.-Y. Lin, M.~Maire, S.~Belongie, J.~Hays, P.~Perona, D.~Ramanan, P.~Dollár,
  and C.~L. Zitnick.
\newblock {Microsoft COCO: Common Objects in Context}.
\newblock In {\em ECCV}, 2014.

\bibitem{liu_emnlp16}
C.-W. Liu, R.~Lowe, I.~V. Serban, M.~Noseworthy, L.~Charlin, and J.~Pineau.
\newblock {How NOT To Evaluate Your Dialogue System: An Empirical Study of
  Unsupervised Evaluation Metrics for Dialogue Response Generation}.
\newblock In {\em EMNLP}, 2016.

\bibitem{liu_eccv16}
W.~Liu, D.~Anguelov, D.~Erhan, C.~Szegedy, S.~Reed, C.-Y. Fu, and A.~C. Berg.
\newblock {SSD: Single Shot MultiBox Detector}.
\newblock In {\em ECCV}, 2016.

\bibitem{lowe_sigdial15}
R.~Lowe, N.~Pow, I.~Serban, and J.~Pineau.
\newblock {The Ubuntu Dialogue Corpus: A Large Dataset for Research in
  Unstructured Multi-Turn Dialogue Systems}.
\newblock In {\em SIGDIAL}, 2015.

\bibitem{lu_github15}
J.~Lu, X.~Lin, D.~Batra, and D.~Parikh.
\newblock {Deeper LSTM and Normalized CNN Visual Question Answering model}.
\newblock \url{https://github.com/VT-vision-lab/VQA_LSTM_CNN}, 2015.

\bibitem{lu_nips16}
J.~Lu, J.~Yang, D.~Batra, and D.~Parikh.
\newblock {Hierarchical Question-Image Co-Attention for Visual Question
  Answering}.
\newblock In {\em NIPS}, 2016.

\bibitem{malinowski_nips14}
M.~Malinowski and M.~Fritz.
\newblock {A Multi-World Approach to Question Answering about Real-World Scenes
  based on Uncertain Input}.
\newblock In {\em NIPS}, 2014.

\bibitem{malinowski_iccv15}
M.~Malinowski, M.~Rohrbach, and M.~Fritz.
\newblock Ask your neurons: {A} neural-based approach to answering questions
  about images.
\newblock In {\em ICCV}, 2015.

\bibitem{mei_aaai16}
H.~Mei, M.~Bansal, and M.~R. Walter.
\newblock Listen, attend, and walk: Neural mapping of navigational instructions
  to action sequences.
\newblock In {\em AAAI}, 2016.

\bibitem{bing_spellapi}
Microsoft.
\newblock {Bing Spell Check API}.
\newblock
  \url{https://www.microsoft.com/cognitive-services/en-us/bing-spell-check-api/documentation}.

\bibitem{mnih_nature15}
V.~Mnih, K.~Kavukcuoglu, D.~Silver, A.~A. Rusu, J.~Veness, M.~G. Bellemare,
  A.~Graves, M.~Riedmiller, A.~K. Fidjeland, G.~Ostrovski, S.~Petersen,
  C.~Beattie, A.~Sadik, I.~Antonoglou, H.~King, D.~Kumaran, D.~Wierstra,
  S.~Legg, and D.~Hassabis.
\newblock Human-level control through deep reinforcement learning.
\newblock {\em Nature}, 518(7540):529--533, 02 2015.

\bibitem{mostafazadeh_arxiv17}
N.~Mostafazadeh, C.~Brockett, B.~Dolan, M.~Galley, J.~Gao, G.~P. Spithourakis,
  and L.~Vanderwende.
\newblock {Image-Grounded Conversations: Multimodal Context for Natural
  Question and Response Generation}.
\newblock {\em arXiv preprint arXiv:1701.08251}, 2017.

\bibitem{paek_elds01}
T.~Paek.
\newblock {Empirical methods for evaluating dialog systems}.
\newblock In {\em Proceedings of the workshop on Evaluation for Language and
  Dialogue Systems-Volume 9}, 2001.

\bibitem{plummer_iccv15}
B.~A. Plummer, L.~Wang, C.~M. Cervantes, J.~C. Caicedo, J.~Hockenmaier, and
  S.~Lazebnik.
\newblock Flickr30k entities: Collecting region-to-phrase correspondences for
  richer image-to-sentence models.
\newblock In {\em ICCV}, 2015.

\bibitem{rajpurkar_emnlp16}
P.~Rajpurkar, J.~Zhang, K.~Lopyrev, and P.~Liang.
\newblock {SQuAD: 100,000+ Questions for Machine Comprehension of Text}.
\newblock In {\em EMNLP}, 2016.

\bibitem{ramanathan_eccv14}
V.~Ramanathan, A.~Joulin, P.~Liang, and L.~Fei-Fei.
\newblock {Linking people with "their" names using coreference resolution}.
\newblock In {\em ECCV}, 2014.

\bibitem{ray_emnlp16}
A.~Ray, G.~Christie, M.~Bansal, D.~Batra, and D.~Parikh.
\newblock {Question Relevance in VQA: Identifying Non-Visual And False-Premise
  Questions}.
\newblock In {\em EMNLP}, 2016.

\bibitem{ren_nips15}
M.~Ren, R.~Kiros, and R.~Zemel.
\newblock {Exploring Models and Data for Image Question Answering}.
\newblock In {\em NIPS}, 2015.

\bibitem{rohrbach_eccv16}
A.~Rohrbach, M.~Rohrbach, R.~Hu, T.~Darrell, and B.~Schiele.
\newblock Grounding of textual phrases in images by reconstruction.
\newblock In {\em ECCV}, 2016.

\bibitem{rohrbach_cvpr15}
A.~Rohrbach, M.~Rohrbach, N.~Tandon, and B.~Schiele.
\newblock A dataset for movie description.
\newblock In {\em CVPR}, 2015.

\bibitem{serban_acl16}
I.~V. Serban, A.~Garc{\'{\i}}a{-}Dur{\'{a}}n, {\c{C}}.~G{\"{u}}l{\c{c}}ehre,
  S.~Ahn, S.~Chandar, A.~C. Courville, and Y.~Bengio.
\newblock {Generating Factoid Questions With Recurrent Neural Networks: The 30M
  Factoid Question-Answer Corpus}.
\newblock In {\em ACL}, 2016.

\bibitem{serban_aaai16}
I.~V. Serban, A.~Sordoni, Y.~Bengio, A.~Courville, and J.~Pineau.
\newblock {Building End-To-End Dialogue Systems Using Generative Hierarchical
  Neural Network Models}.
\newblock In {\em AAAI}, 2016.

\bibitem{serban_arxiv16}
I.~V. Serban, A.~Sordoni, R.~Lowe, L.~Charlin, J.~Pineau, A.~Courville, and
  Y.~Bengio.
\newblock {A Hierarchical Latent Variable Encoder-Decoder Model for Generating
  Dialogues}.
\newblock {\em arXiv preprint arXiv:1605.06069}, 2016.

\bibitem{silver_nature16}
D.~Silver, A.~Huang, C.~J. Maddison, A.~Guez, L.~Sifre, G.~Van Den~Driessche,
  J.~Schrittwieser, I.~Antonoglou, V.~Panneershelvam, M.~Lanctot, et~al.
\newblock {Mastering the game of Go with deep neural networks and tree search}.
\newblock {\em Nature}, 529(7587):484--489, 2016.

\bibitem{simonyan_iclr15}
K.~Simonyan and A.~Zisserman.
\newblock {Very deep convolutional networks for large-scale image recognition}.
\newblock In {\em ICLR}, 2015.

\bibitem{tapaswi_cvpr16}
M.~Tapaswi, Y.~Zhu, R.~Stiefelhagen, A.~Torralba, R.~Urtasun, and S.~Fidler.
\newblock {MovieQA: Understanding Stories in Movies through
  Question-Answering}.
\newblock In {\em CVPR}, 2016.

\bibitem{tu_mm16}
K.~Tu, M.~Meng, M.~W. Lee, T.~E. Choe, and S.~C. Zhu.
\newblock {Joint Video and Text Parsing for Understanding Events and Answering
  Queries}.
\newblock {\em IEEE MultiMedia}, 2014.

\bibitem{venugopalan_iccv15}
S.~Venugopalan, M.~Rohrbach, J.~Donahue, R.~J. Mooney, T.~Darrell, and
  K.~Saenko.
\newblock {Sequence to Sequence - Video to Text}.
\newblock In {\em ICCV}, 2015.

\bibitem{venugopalan_naacl15}
S.~Venugopalan, H.~Xu, J.~Donahue, M.~Rohrbach, R.~J. Mooney, and K.~Saenko.
\newblock {Translating Videos to Natural Language Using Deep Recurrent Neural
  Networks}.
\newblock In {\em NAACL HLT}, 2015.

\bibitem{vinyals_arxiv15}
O.~Vinyals and Q.~Le.
\newblock {A Neural Conversational Model}.
\newblock {\em arXiv preprint arXiv:1506.05869}, 2015.

\bibitem{vinyals_cvpr15}
O.~Vinyals, A.~Toshev, S.~Bengio, and D.~Erhan.
\newblock {Show and tell: {A} neural image caption generator}.
\newblock In {\em CVPR}, 2015.

\bibitem{wang_arxiv16}
L.~Wang, S.~Guo, W.~Huang, Y.~Xiong, and Y.~Qiao.
\newblock {Knowledge Guided Disambiguation for Large-Scale Scene Classification
  with Multi-Resolution CNNs}.
\newblock {\em arXiv preprint arXiv:1610.01119}, 2016.

\bibitem{eliza}
J.~Weizenbaum.
\newblock {ELIZA}.
\newblock \url{http://psych.fullerton.edu/mbirnbaum/psych101/Eliza.htm}.

\bibitem{weston_iclr16}
J.~Weston, A.~Bordes, S.~Chopra, and T.~Mikolov.
\newblock {Towards AI-Complete Question Answering: A Set of Prerequisite Toy
  Tasks}.
\newblock In {\em ICLR}, 2016.

\bibitem{fb_blog16}
S.~Wu, H.~Pique, and J.~Wieland.
\newblock {Using Artificial Intelligence to Help Blind People `See' Facebook}.
\newblock
  http://newsroom.fb.com/news/2016/04/using-artificial-intelligence-to-help-blind-people-see-facebook/,
  2016.

\bibitem{yang_cvpr16}
Z.~Yang, X.~He, J.~Gao, L.~Deng, and A.~J. Smola.
\newblock {Stacked Attention Networks for Image Question Answering}.
\newblock In {\em CVPR}, 2016.

\bibitem{yu_iccv15}
L.~Yu, E.~Park, A.~C. Berg, and T.~L. Berg.
\newblock {Visual Madlibs: Fill in the blank Image Generation and Question
  Answering}.
\newblock In {\em ICCV}, 2015.

\bibitem{zhang_cvpr16}
P.~Zhang, Y.~Goyal, D.~Summers-Stay, D.~Batra, and D.~Parikh.
\newblock {Yin and Yang: Balancing and Answering Binary Visual Questions}.
\newblock In {\em CVPR}, 2016.

\bibitem{zhu_cvpr16}
Y.~Zhu, O.~Groth, M.~Bernstein, and L.~Fei-Fei.
\newblock {Visual7W: Grounded Question Answering in Images}.
\newblock In {\em CVPR}, 2016.

\bibitem{zitnick_ai16}
C.~L. Zitnick, A.~Agrawal, S.~Antol, M.~Mitchell, D.~Batra, and D.~Parikh.
\newblock Measuring machine intelligence through visual question answering.
\newblock {\em AI Magazine}, 2016.

\end{thebibliography}
}

\end{document}